\documentclass[conference]{IEEEtran}
\IEEEoverridecommandlockouts
\usepackage{cite}
\usepackage{amsmath,amssymb,amsfonts}
\usepackage{algorithmic}
\usepackage{graphicx}
\usepackage{textcomp}
\usepackage{xcolor}
\def\BibTeX{{\rm B\kern-.05em{\sc i\kern-.025em b}\kern-.08em
    T\kern-.1667em\lower.7ex\hbox{E}\kern-.125emX}}
\begin{document}

\title{Asymmetric Bilateral Phase Correlation for Optical Flow Estimation in the Frequency Domain}

\author{\IEEEauthorblockN{Vasileios Argyriou}
\textit{Kingston University London}\\
London, UK \\
Vasileios.Argyriou@kingston.ac.uk}

\maketitle

\begin{abstract}
We address the problem of motion estimation in images operating in the frequency domain. A method is presented which extends phase correlation to handle multiple motions present in an area. Our scheme is based on a novel Bilateral-Phase Correlation (BLPC) technique that incorporates the concept and principles of Bilateral Filters retaining the motion boundaries by taking into account the difference both in value and distance in a manner very similar to Gaussian convolution. The optical flow is obtained by applying the proposed method at certain locations selected based on the present motion differences and then performing non-uniform interpolation in a multi-scale iterative framework. Experiments with several well-known datasets with and without ground-truth show that our scheme outperforms recently proposed state-of-the-art phase correlation based optical flow methods.
\end{abstract}


\section{Introduction}

Optical flow estimation is one of the fundamental problems in computer vision \cite{Fortun2015,Jia2011} with significant applications such as structure from motion, SLAM, 3D reconstruction \cite{ArgyriouChapter} object and pedestrian tracking, face, object or building detection and behaviour recognition \cite{7563843,6977392}, robots or vehicle navigation, image super-resolution, medical image registration, restoration, and compression. Dense motion estimation or optical flow is an ill-posed problem and it is defined as the process of approximating the 3D movement in a scene on a 2D image sequence based on the illumination changes due to camera or object motion. As an outcome of this process a dense vector field is obtained providing motion information for each pixel in the image sequences.

Several aspects make this problem particularly challenging including occlusions (i.e. points can appear or disappear between two frames), and the aperture problem (i.e. regions of uniform appearance with the local cues to not be able to provide any information about the motion). In the literature there are many approaches to overcome these challenges and compute the optical flow; and most of them belong in one of the following categories. Block matching techniques \cite{Liu1993} are based on the assumption that all pixels in a block undergo the same motion and are mainly applied for standards conversion, and in international standards for video communications such as MPEGx and H.26x; Gradient-based techniques \cite{Horn1981,Lucas1981} are utilising the spatio-temporal image gradients introducing also some robust or smoothness constraints \cite{Weickert2001}; Bayesian techniques \cite{Glocker2010} that utilize probability smoothness constraints or Markov Random Fields (MRFs) over the entire image, focusing on finding the Maximum a Posteriori (MAP) solution. These methods are also combined with multi-scale schemes offering more accurate and robust motion vectors; GPU based approaches using local operations have been proposed \cite{Gwosdek2010} providing accurate estimates; and Phase correlation techniques \cite{Tekalp1995} that operate in the frequency domain computing the optical flow by applying locally phase correlation to the frames.

\begin{figure}[htb]
\centering
\includegraphics[width=0.24\columnwidth]{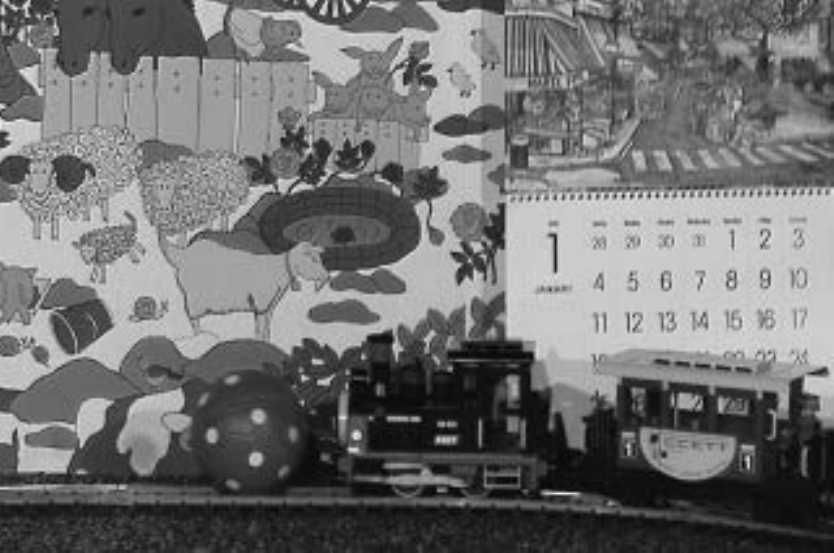}
\includegraphics[width=0.24\columnwidth]{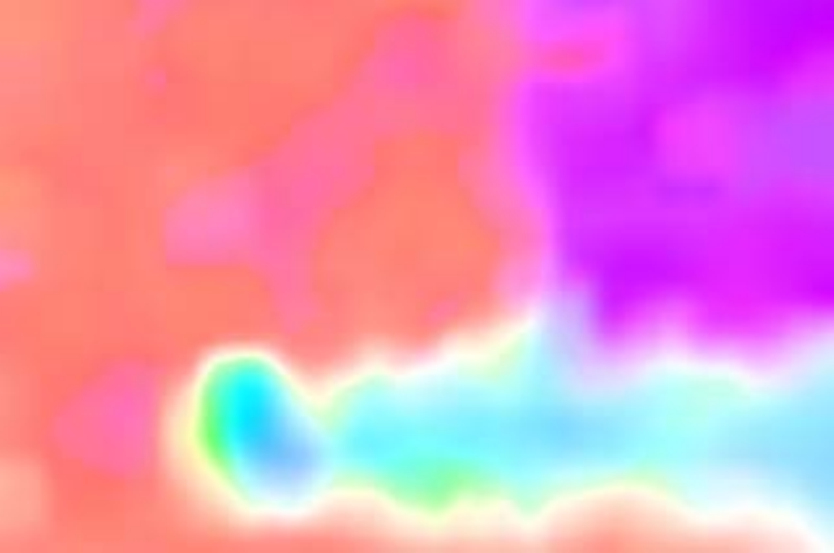}
\includegraphics[width=0.24\columnwidth]{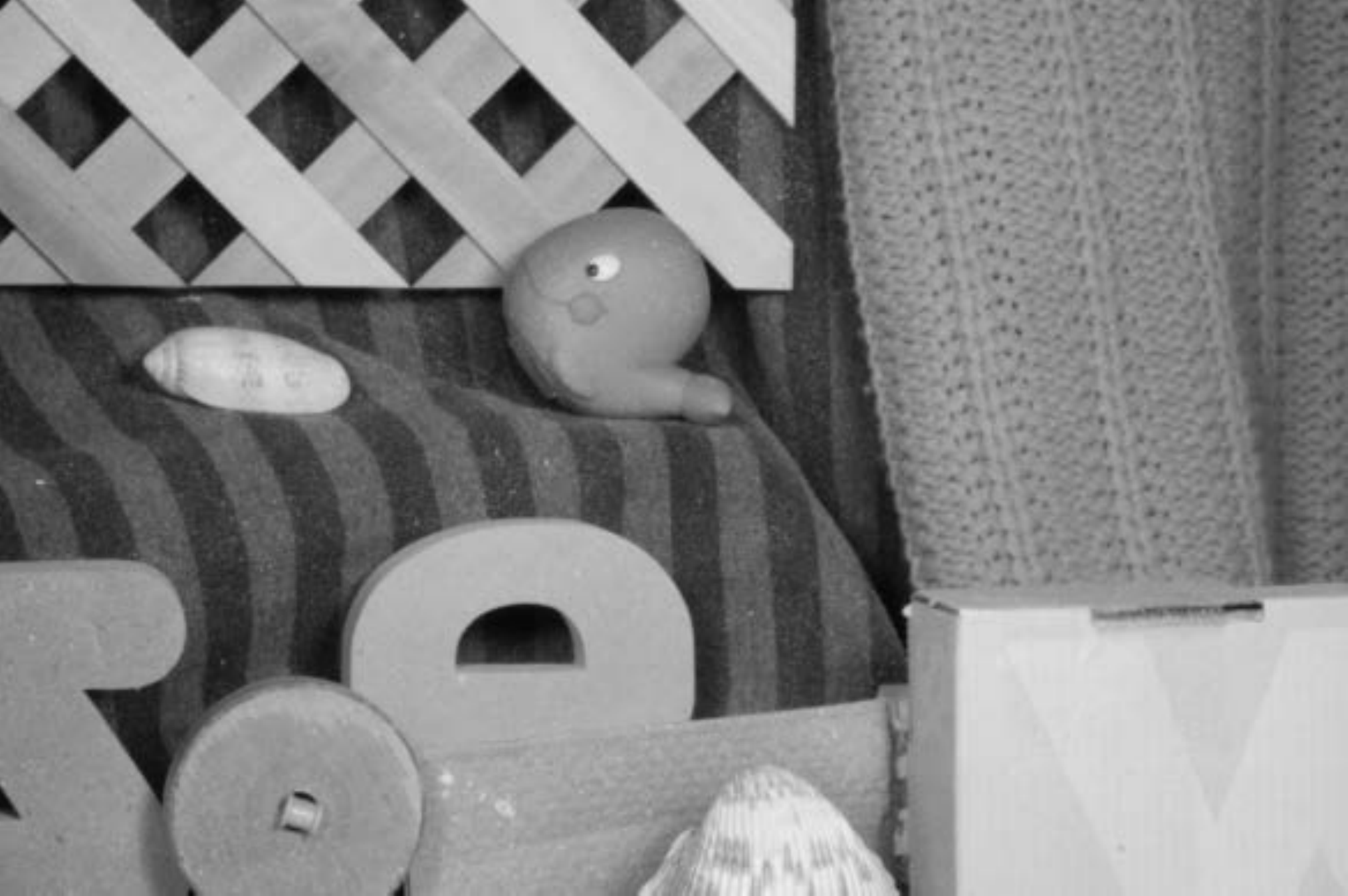}
\includegraphics[width=0.24\columnwidth]{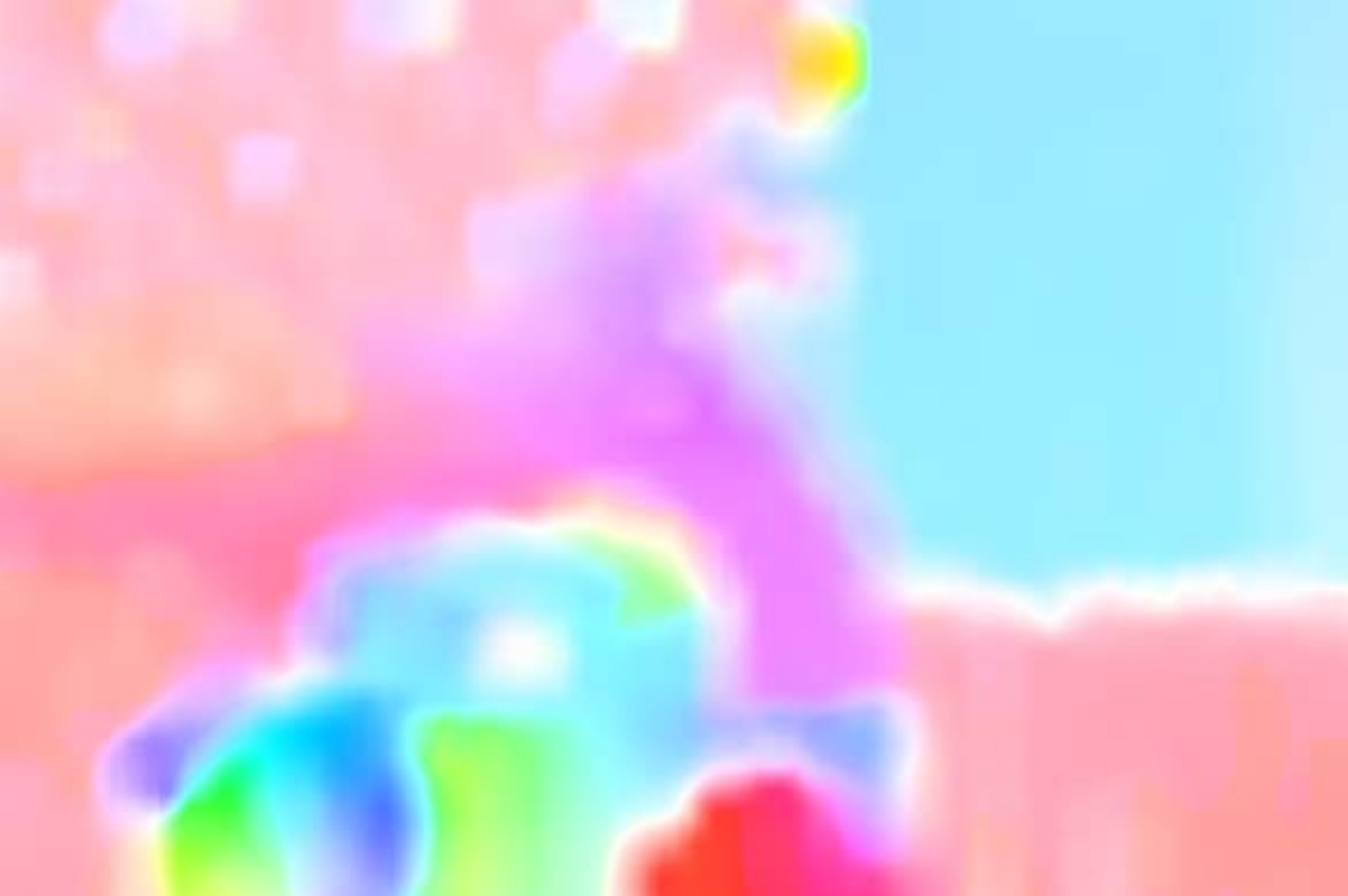}
\caption{An example inaccurate or blurred motion vectors around motion boundaries or depth discontinuities.}\label{fig:FigBPC01}
\end{figure}

In the core of these methods we usually have phase correlation \cite{Pearson1977,Thomas1987}, which has become one of the motion estimation methods of choice for a wide range of professional studio and broadcasting applications. Phase Correlation (PC) and other frequency domain approaches (that are based on the shift property of the Fourier Transform (FT)) offer speed through the use of FFT routines and enjoy a high degree of accuracy featuring several significant properties: immunity to uniform variations of illumination, insensitivity to changes in spectral energy and excellent peak localization accuracy. Furthermore, it provides sub-pixel accuracy that has a significant impact on motion compensated error performance and image registration for tracking, recognition, super-resolution and other applications, as theoretical and experimental analyses have suggested. Sub-pixel accuracy mainly can be achieved through the use of bilinear interpolation, which is also applicable to frequency domain motion estimation methods.

One of the main issues of frequency domain registration methods is that it is hard to identify the correct motion parameters in the case of multiple motions present in a region. This requirement is more pronounced especially around motion boundaries or depth discontinuities resulting to inaccurate or blurred motion vectors (see figure \ref{fig:FigBPC01}). One of the most common approaches to overcome this issue is to incorporate block matching, computing the motion compensated error for a set of candidate motions \cite{Alba2010,Reyes2013,Reyes2014}. Another solution could be to reduce the block/window size used to apply Phase Correlation, but this affects further the accuracy of the estimates, since in order to obtain reliable motion estimates large blocks of image data are required. Also, smaller windows will not support large motion vectors. The problem becomes more visible in cases that overlapped windows are used of high density and when accurate motion border estimation is essential. Furthermore, another issue that effects the optical flow techniques is the overall complexity especially for high resolution images or videos (4K-UltraHD) and the available computational power especially from mobile devices. Additionally, since the number of real-time computer vision applications constantly increases, the overall performance and complexity should allow real-time or near real-time optical flow estimation for such high resolutions \cite{Tao2012}.

In this paper we introduce a novel high-performance optical flow algorithm operating in the frequency domain based on the principles of Phase Correlation. The overall complexity is very low allowing near real-time flow estimation for very high resolution video sequences. In order to overcome the problem of blurred or erroneous motion estimates at the motion borders, providing accurate and sharp motion vectors, a novel Asymmetric  Bilateral-Phase Correlation technique is introduced, incorporating the concept and principles of Bilateral Filters. We propose to use the idea of taking into account the difference in value with the neighbours to preserve motion edges in combination with a weighted average of nearby pixels, in a manner very similar to Gaussian convolution, integrated into Phase Correlation process. The key idea of the bilateral filter is that for a pixel to influence another pixel, it should not only occupy a nearby location but also have a similar value, \cite{Tomasi1998,Durand2002}. Finally, subpixel accuracy is obtained through the use of fast and simple interpolation schemes. Experiments with well-known datasets of video sequences with and without ground truth have shown that our scheme performs significantly better than other state-of-art frequency domain optical flow methods, while overall in comparison with state of the art methods provides very accurate results with low complexity especially for high resolution 4K video sequences. In summary, our contributions are: 1) A hierarchical framework for optical flow estimation, applying motion estimation techniques only to a small amount of regions of interest based on the motion compensated prediction error allowing us to control the overall complexity and required computational power. 2) A novel asymmetric bilateral filter operating simultaneously on two separate frames and using unequal size blocks, allowing us to carry the local spatial properties and intensity constraints of one frame to the other (i.e. using the properties of a small block/window to filter the whole image). 3) Integration of the proposed asymmetric bilateral filter in the Phase Correlation process that can be applied either in the pixel domain as a pre-processing step or directly in the frequency domain as a multiplication in a higher dimensional space.

This paper is organised as follows. In Section 2, we review the state-of-the-art in optical flow estimation using phase correlation and other pixel or gradient based approaches. In Section 3, we discuss the principles of the proposed Bilateral Phase Correlation (BLPC) and the key features of this optical flow framework are analysed. In Section 4 we present experimental results while in Section 5 we draw conclusions arising from this paper.

\section{Related work}

In the following section, we review the major optical flow methodologies including frequency domain approaches, and energy minimization methods developed for computer vision applications.

\begin{figure}[htb]
\centering
\includegraphics[width=0.5\columnwidth]{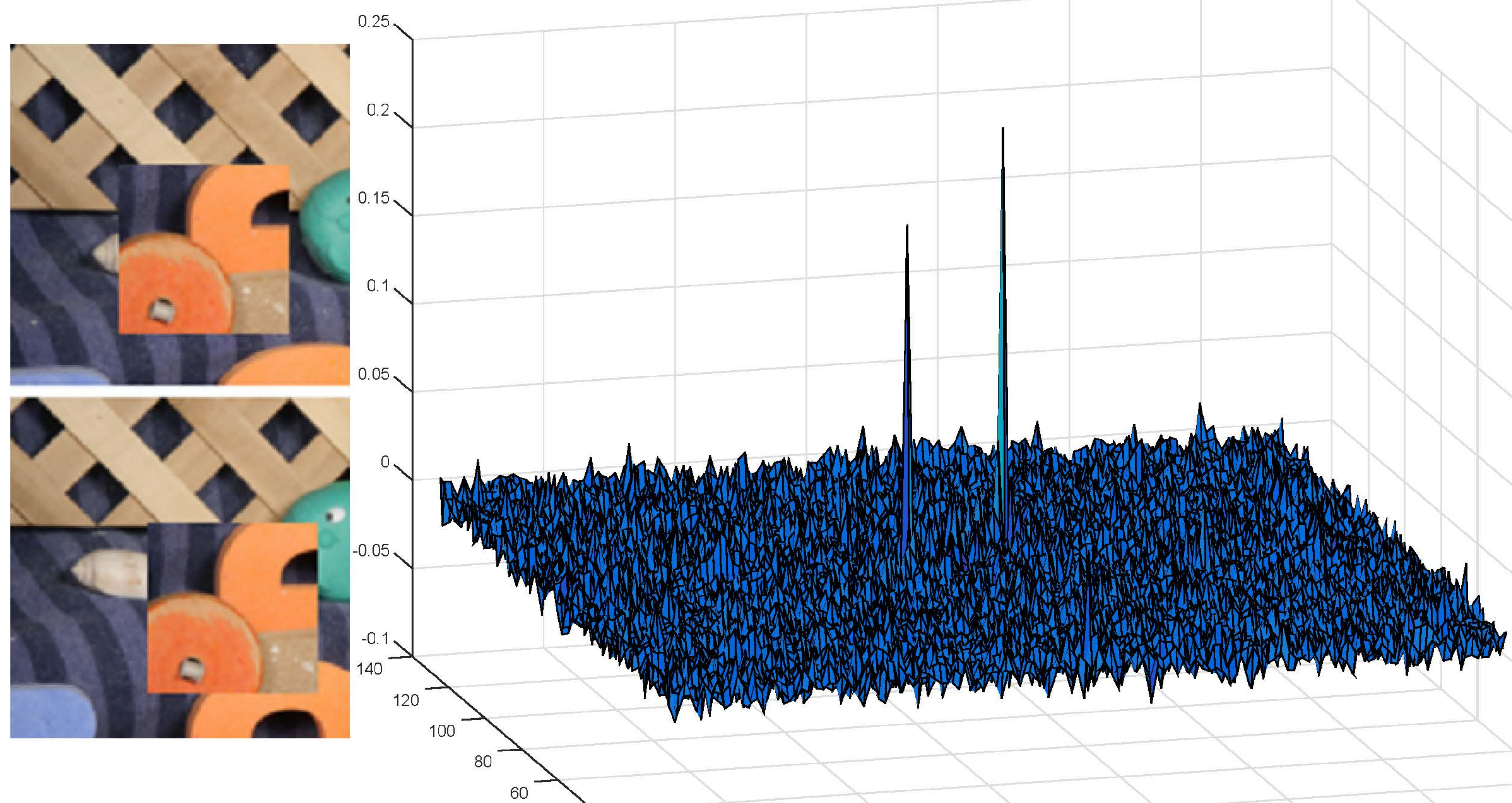}
\caption{Example of correlation surface characterised by the presence of two peaks corresponding to the present motions.}\label{fig:FigBPC02}
\end{figure}

A brief review of current state-of-the-art Fourier-based methods for optical flow estimation is presented \cite{Kruger1996}. Several subpixel motion estimation and image registration algorithms operating in the frequency domain have been introduced \cite{Tong2015,Zhongke2003,Ghoul2016,Maik2014,Uss2014,Cheng2013,Tong2015b,Douini2017,Mozerov2013}.  In \cite{Hoge2003}, Hoge proposes to apply a rank-1 approximation to the phase difference matrix and then performs unwrapping estimating the motion vectors. The work in \cite{Keller2007} is a noise-robust extension to \cite{Hoge2003}, where noise is assumed to be Additive White Gaussian Noise (AWGN). The authors in \cite{Balci2006} derive the exact parametric model of the phase difference matrix and solve an optimization problem for fitting the analytic model to the noisy data. To estimate the subpixel shifts, Stone et al. \cite{Stone2001} fit the phase values to a 2D linear function using linear regression, after masking out frequency components corrupted by aliasing. An extension to this method for the additional estimation of planar rotation has been proposed in \cite{Vandewalle2006}. Foroosh et al. \cite{Foroosh2002} showed that the phase correlation function is the Dirichlet kernel and provided analytic results for the estimation of the subpixel shifts using a $sinc$ approximation. Finally, a fast method for subpixel estimation based on FFTs has been proposed in \cite{Ren2007}. Notice that the above methods either assume aliasing-free images \cite{Foroosh2002,Ren2010,Ren2007,Balci2006,Argyriou2007,Argyriou2011}, or cope with aliasing by frequency masking \cite{Stone2001,Keller2007,Hoge2003,Vandewalle2006}, which requires fine tuning.

Over the last decades different optical flow methods based on the above Phase Correlation techniques have been proposed. In \cite{Thomas1987} Thomas proposes a vector measurement method that has the accuracy of the Fourier techniques, combined with the ability to measure multiple objects' motion in a scene. The method is also able to assign vectors to individual pixels if required. The authors in \cite{Yan2008} proposed raster scanning of the image pair method using a moving window to estimate the motion of each small window with the introduced compound phase correlation algorithm. In order to improve the fitting accuracy of the phase difference plane a phase fringe filter is utilised in Fourier domain following an improved extension of the work in \cite{Hoge2003}. The main issue with this approach is that it is not clear how it performs in the presence of multiple motions and furthermore it requires the motion boundaries to be quite distinct. In \cite{Ho2008} a regular grid of patches is generated and the optical flow is estimated by calculating the phase correlation of each pair of co-sited patches using the Fourier Mellin Transform (FMT). This approach allows the estimation not only of translation but also scale and rotation motion of image patches. The main limitation of this algorithm is that it cannot handle multiple motions and preserve the motion edges. In the work presented in \cite{Argyriou2009} the authors use an adaptive grid to extract image patches and then apply gradient correlation in an iterative process using 2D median filters to overcome the issues related to multiple motions. Despite the filtering stage the motion blurring is not avoided in many cases due to local characteristics that may effect the filter. The authors in \cite{Reyes2013,Alba2010,Reyes2014,Gautama2002} proposed a solution, based on the concept suggested by Thomas in \cite{Thomas1987}, defining a large enough set of candidate motion vectors, and using a combinatorial optimization algorithm (such as block-matching) to find, for each point of interest, the candidate which best represents the motion at that location. This is the main difference with the proposed method since in their work they try to minimize directly the motion compensated error. Also they are using large windows and the obtained set contains the most representative maxima of phase-correlation between the two input images, computed for different overlapping regions. This approach provides better accuracy and contains less spurious candidates but the problem with the motion boundaries or depth discontinuities remains since it depends on the matching algorithm resulting inaccurate estimates at the borders.

\begin{figure}[htb]
\centering
\includegraphics[width=0.7\columnwidth]{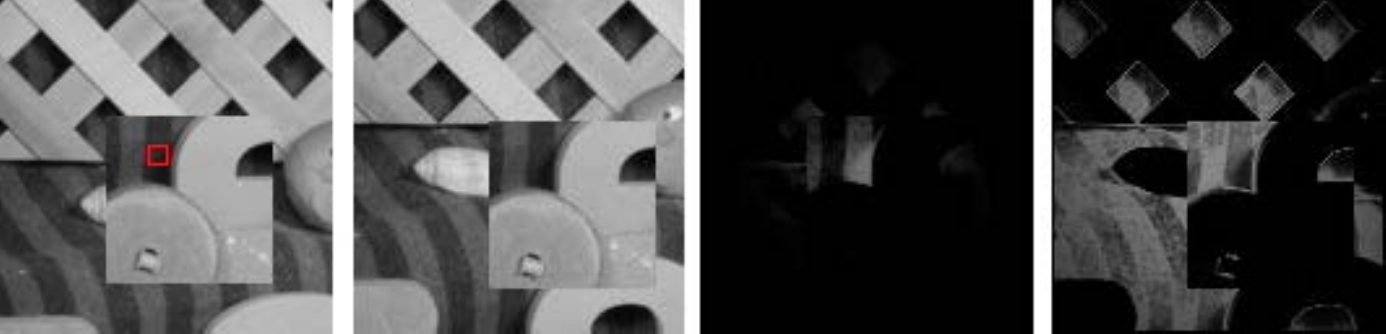}
\caption{An example of obtained images showing that the proposed approach retains pixels that are close to the centre of the window and have their intensities similar to the values of the pixels at the neighboring area of the centre. In these images we observe visually the effect of the proposed novel asymmetric Bilateral filter integrated in the PC method. In more details at the first image we have the small red window that we use to extract the filter restrictions (spatial and intensity constraints) that are then applied to the whole second frame. This allows us to correlate the two frames using only the Bilateral filter information inside the red square.
}\label{fig:FigBPC03}
\end{figure}


\section{Bilateral-PC for optical flow estimation}

In this work, we propose a new optical flow estimation technique operating in the frequency domain based on a novel extension of Phase Correlation that incorporates the principles of bilateral filters. Initially an overview of Phase Correlation is presented, followed by an analysis of the proposed Bilateral-PC technique. Finally, a novel framework for efficient optical flow estimation in the Fourier domain is discussed supporting high resolution images or videos (4K).

Let $I_i(\textbf{m})\textrm{, }\textbf{m}=[x,y]^T\in\mathcal{R}^2\textrm{, }i=1,2$ be two image functions, related by an unknown translation $\textbf{t}=[t_{x},t_{y}]^T\in\mathcal{R}^2$
\begin{equation}
I_2(\textbf{m})=I_1(\textbf{m}-\textbf{t}) \label{equ:AA1}
\end{equation}
Phase correlation schemes are utilised to estimate the translational displacement. Considering each image $I_i(\textbf{m})$ as a continuous periodic image function, then the Fourier transform of $I_{1}$ is given by
\begin{align}
\hat{I}_1(\textbf{k}) = \sum_{\textbf{m}}I_1(\textbf{m})e^{-j(2\pi/N)\textbf{k}^{T}\textbf{m}}
\label{equ:AA2}
\end{align}
where $-N/2\leq\textbf{k}< N/2$, $ \textbf{k}=[k,l]^{T}\in Z^{2}$, $ \textbf{m}=[m,n]^{T}\in \mathcal{R}^{2}$ and $-N/2\leq\textbf{m}< N/2$.

Moving to the shifted version of the image, $I_{2}$, its DFT is given based on the Fourier shift property and assuming no aliasing by
\begin{equation}
\hat{I}_2(\textbf{k}) = \hat{I}_1(\textbf{k}) e^{-j(2\pi/N)\textbf{k}^{T}(N\textbf{t})} \label{equ:AA4}
\end{equation}

\begin{figure}[htb]
\centering
\includegraphics[width=0.6\columnwidth]{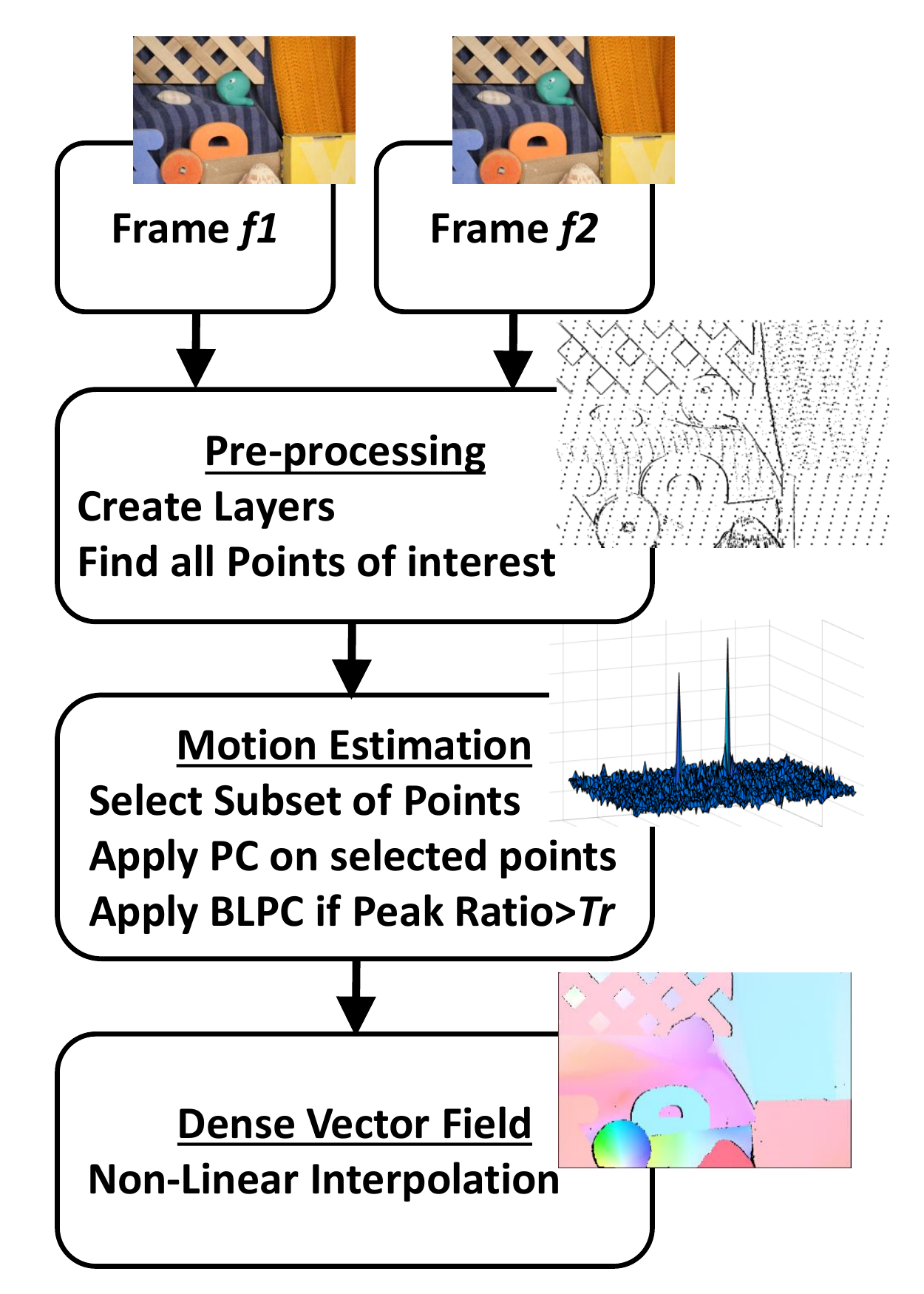}
\caption{The overall methodology and the related stages for optical flow estimation applied in each layer of the image pyramid.}\label{fig:FigBPC05}
\end{figure}

\begin{figure}[htb]
\centering
\includegraphics[width=0.95\columnwidth]{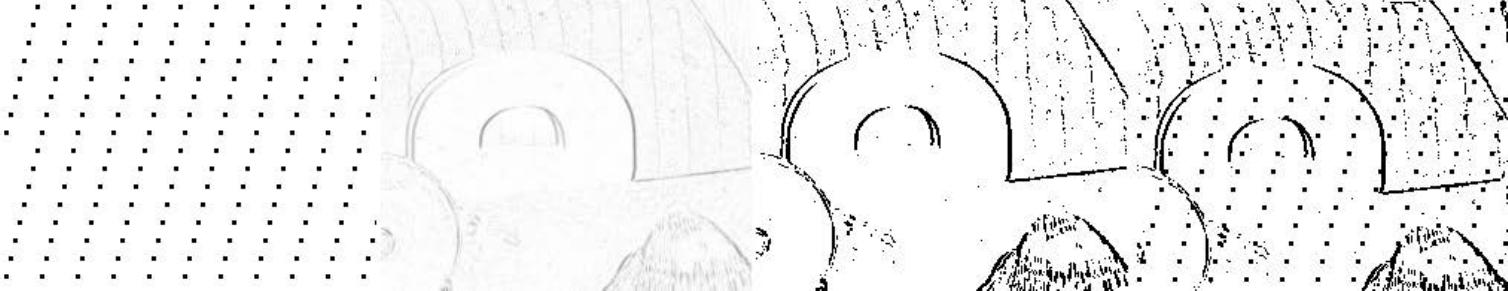}
\caption{An example of the selected pixel locations by applying first uniform sampling (first from left), and then calculating the frame difference (second), applying a binary threshold (third) and finally combining both (fourth).}\label{fig:FigBPC06}
\end{figure}

\begin{figure}[htb]
\centering
\includegraphics[width=0.7\columnwidth]{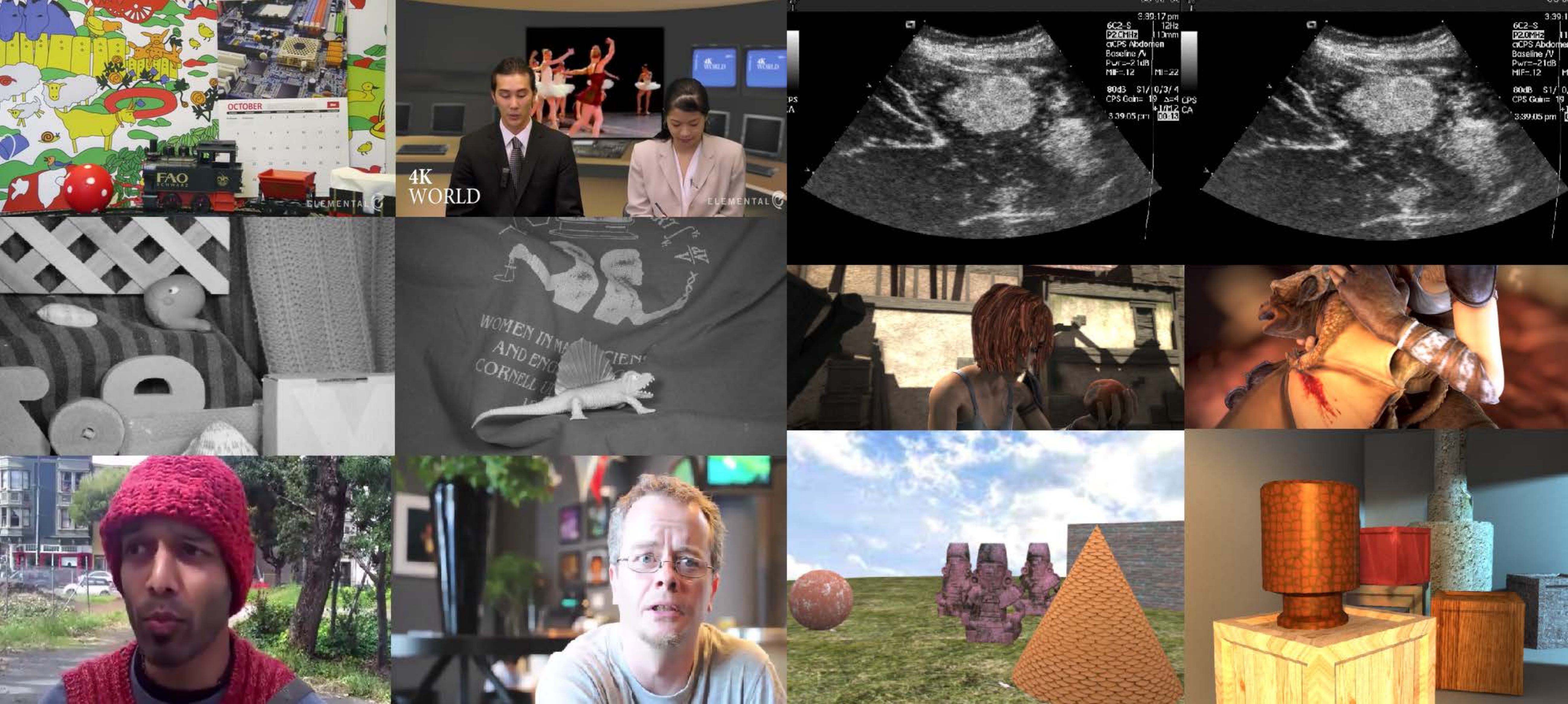}
\caption{Selected frames from the datasets used in our evaluation.}\label{fig:BPCVideoImages}
\end{figure}

\begin{figure}[htb]
\centering
\includegraphics[width=0.7\columnwidth]{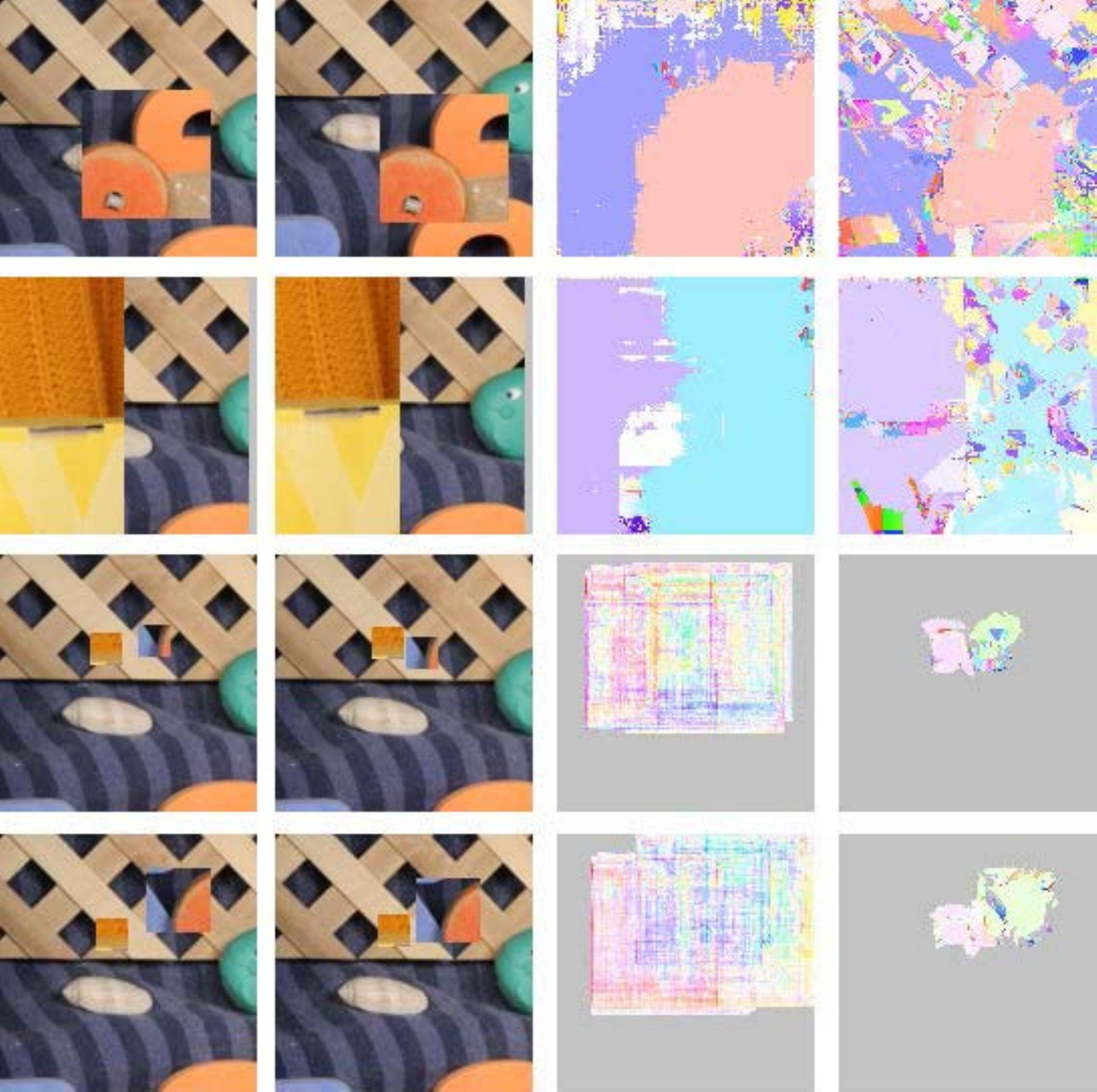}
\caption{Selected frames from the artificial examples used in our evaluation. In the last two columns we can see the obtained colour coded optical flows using PC and BLPC, respectively. It is clear how well the proposed BLPC methods distinguishes the multiple motions, without being affected from the background movement in each pixel location in comparison to PC.} \label{fig:BPCArtificialImages}
\end{figure}

Since the shift property of the Fourier Transform refers to integer shifts and no aliasing-free signals are assumed, we regard that our sampling device eliminates aliasing.
Traditionally phase correlation (PC) is used to estimate the translational displacement, which is perhaps the most widely used correlation-based method in image registration. It finds the maximum of the phase difference function which is defined as the inverse FT of the normalized cross-power spectrum
\begin{equation}
\textrm{PC}(\textbf{u}) \triangleq
F^{-1}\left\{\frac{\hat{I}_2(\textbf{k})\hat{I}_1^*(\textbf{k})}{|\hat{I}_2(\textbf{k})||\hat{I}_1^*(\textbf{k})|}
\right\} = F^{-1}\{e^{j\textbf{k}^T\textbf{t}} \} =
\delta{(\textbf{u}-\textbf{t})} \label{Pcorr}
\end{equation}
where $*$ denotes complex conjugate and $F^{-1}$ the inverse Fourier transform.


\begin{figure}[htb]
\centering
\includegraphics[width=0.2\columnwidth]{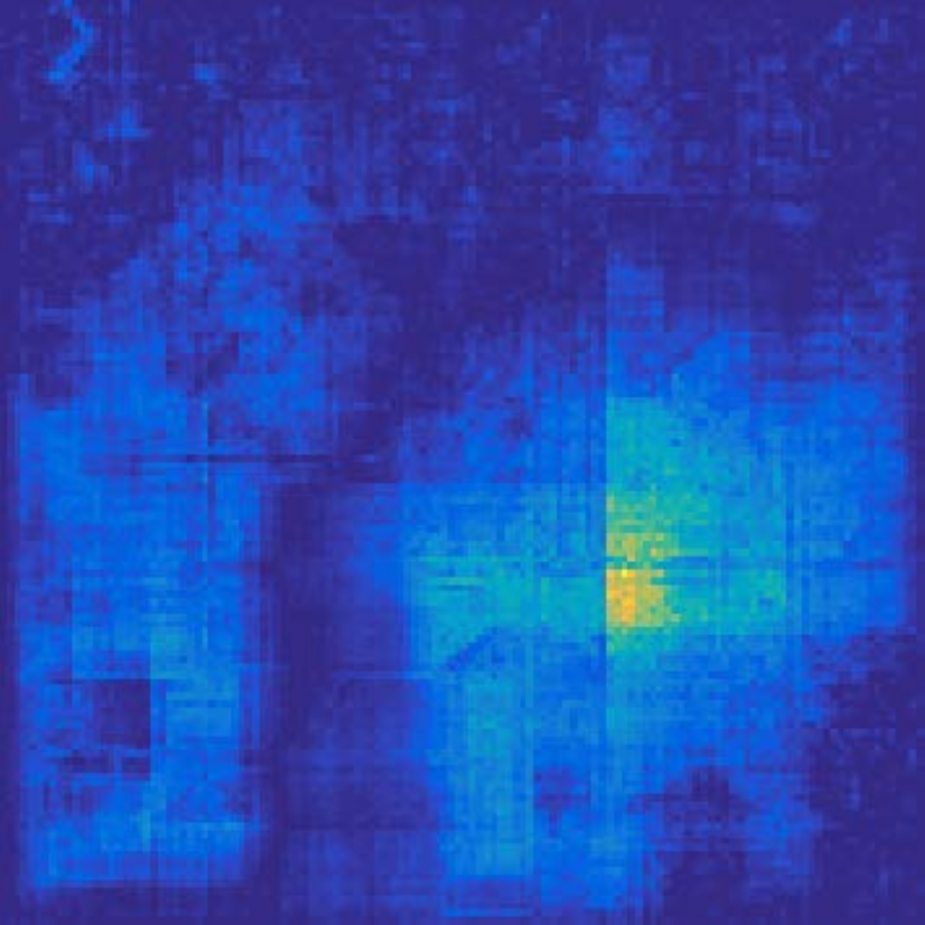}
\includegraphics[width=0.2\columnwidth]{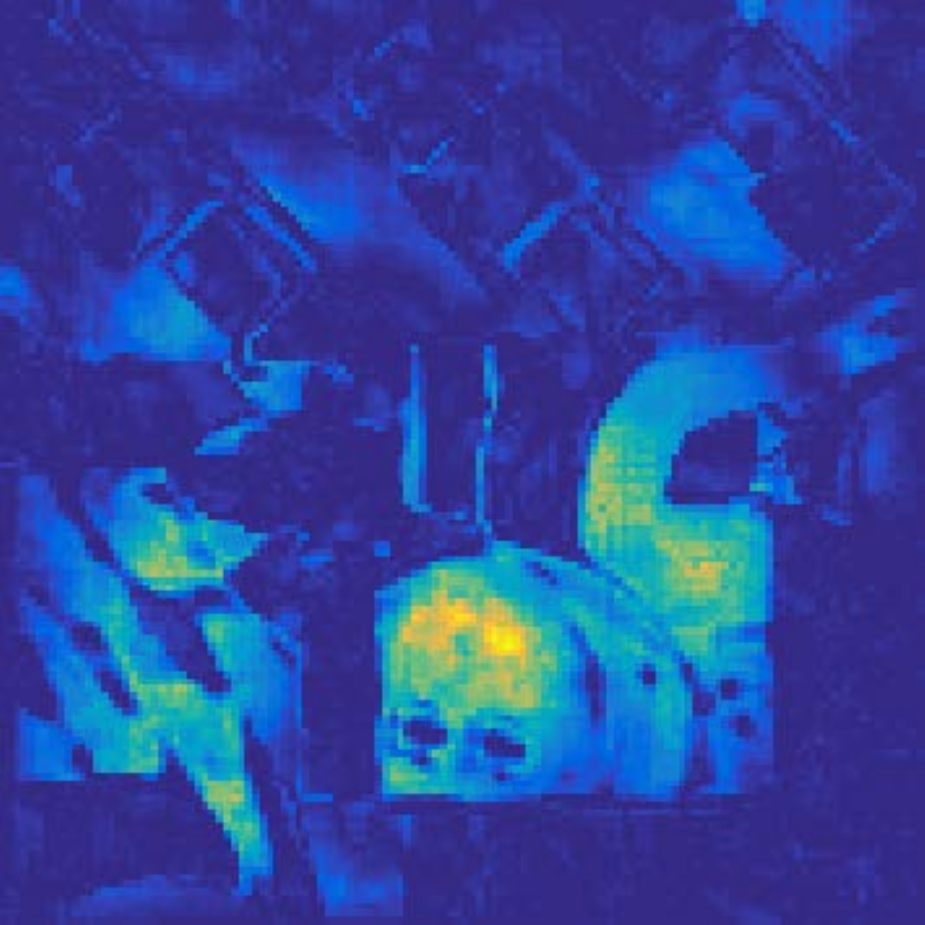} \hspace{2 mm}
\includegraphics[width=0.2\columnwidth]{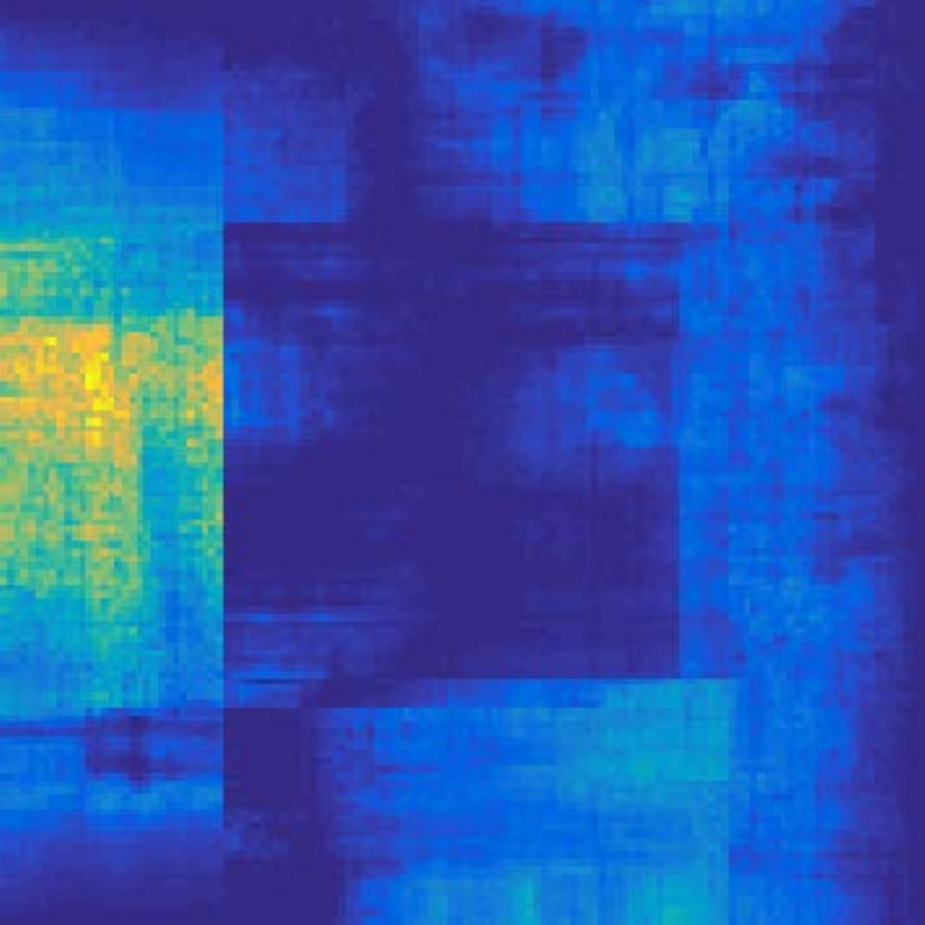}
\includegraphics[width=0.2\columnwidth]{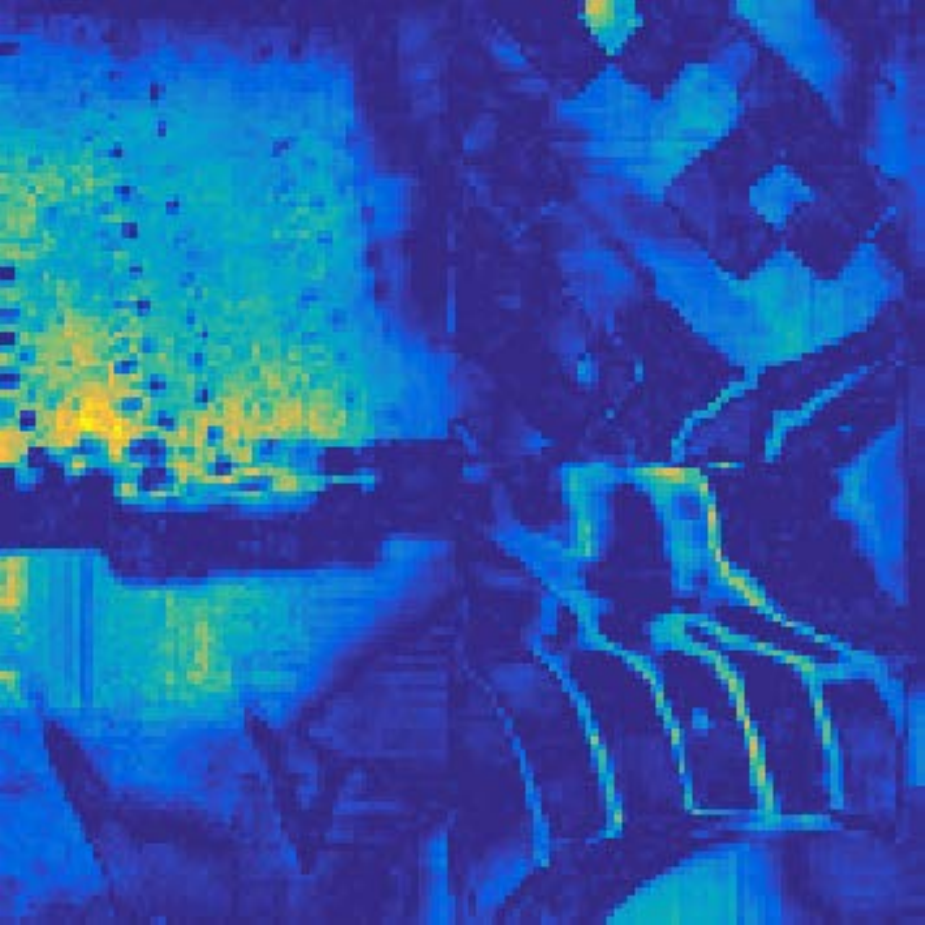}\\ \vspace{1 mm}
\includegraphics[width=0.2\columnwidth]{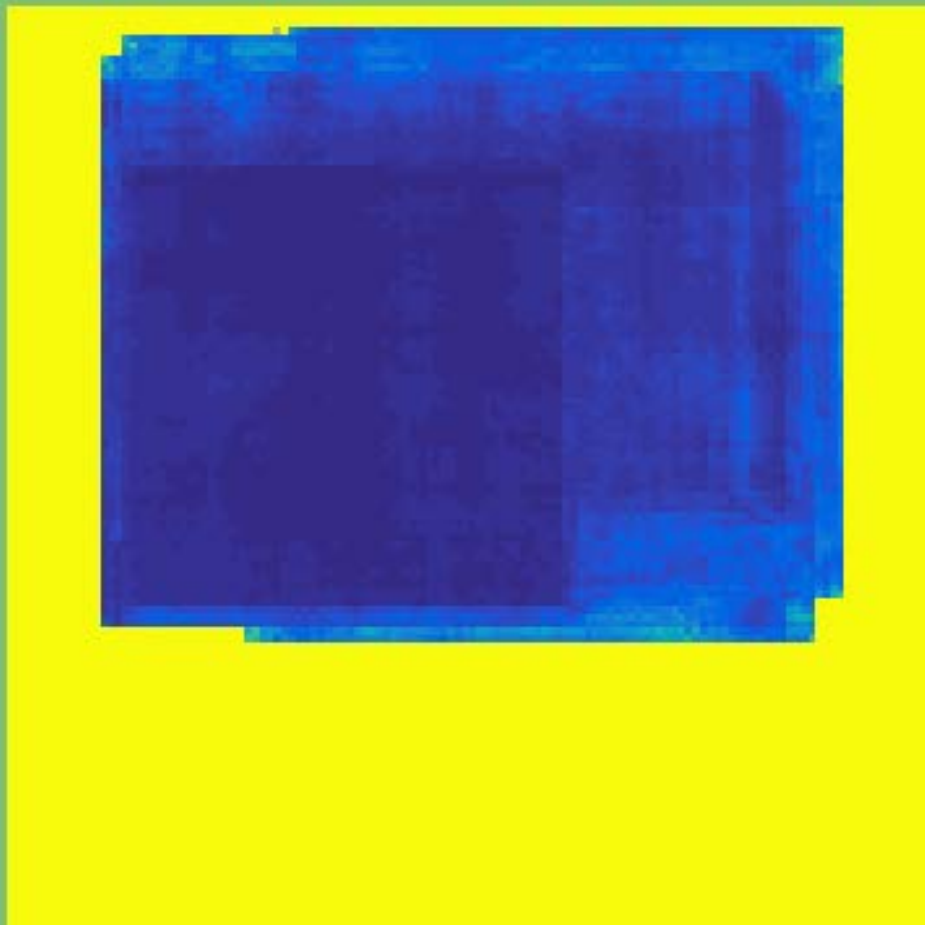}
\includegraphics[width=0.2\columnwidth]{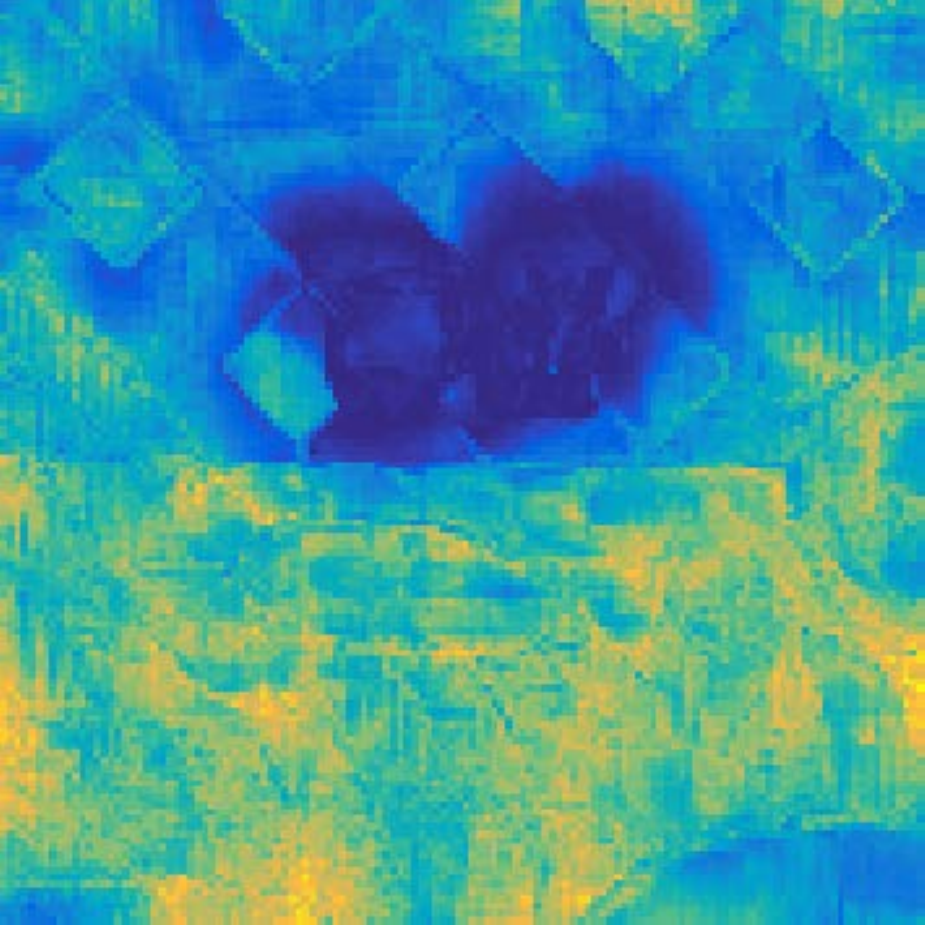} \hspace{2 mm}
\includegraphics[width=0.2\columnwidth]{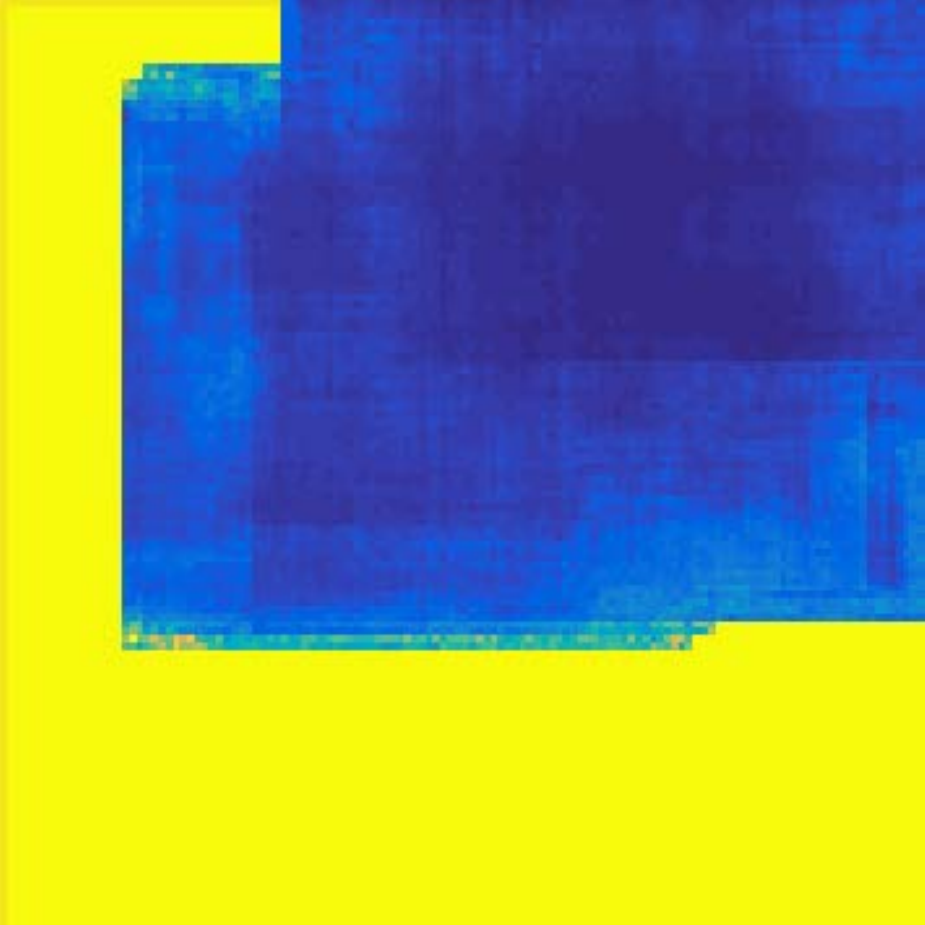}
\includegraphics[width=0.2\columnwidth]{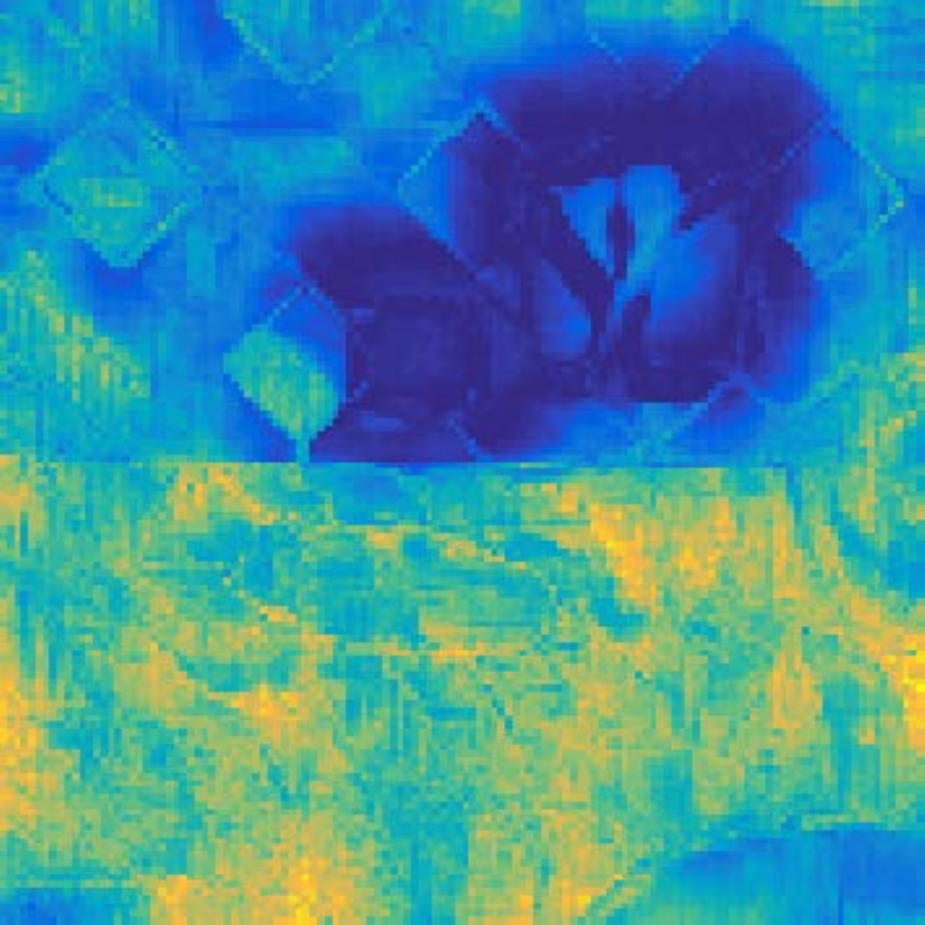}\\ \vspace{1 mm}
\includegraphics[width=0.2\columnwidth]{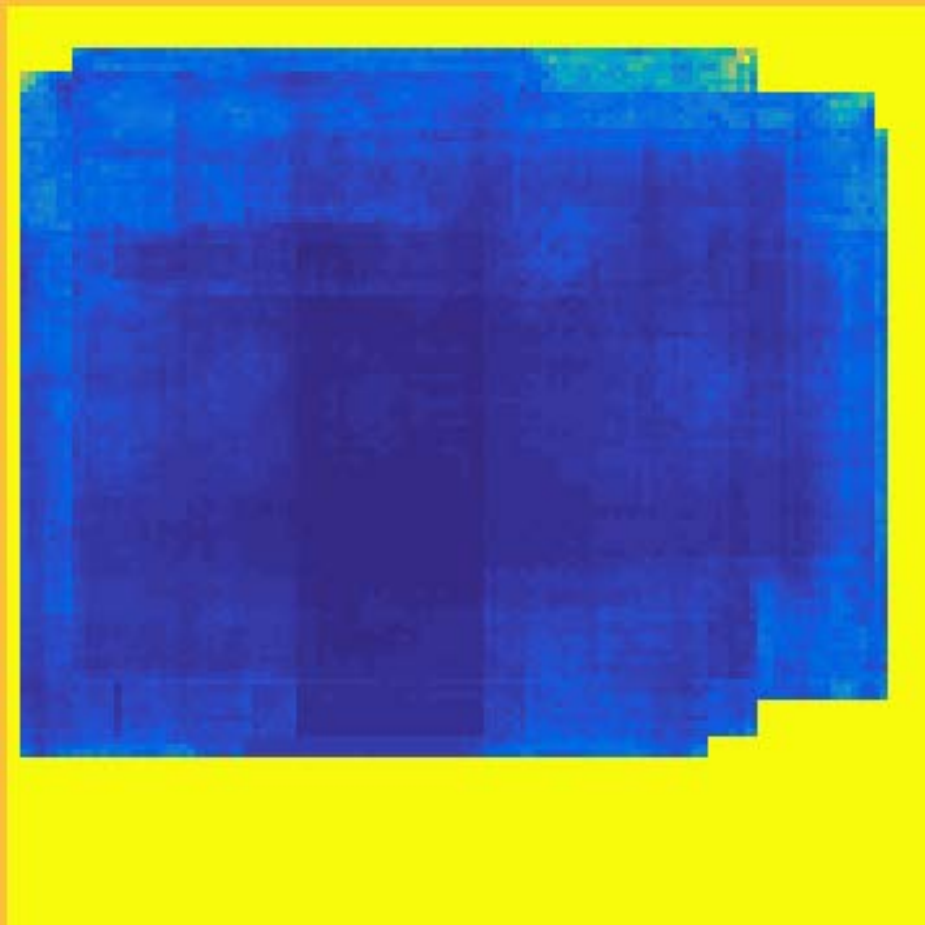}
\includegraphics[width=0.2\columnwidth]{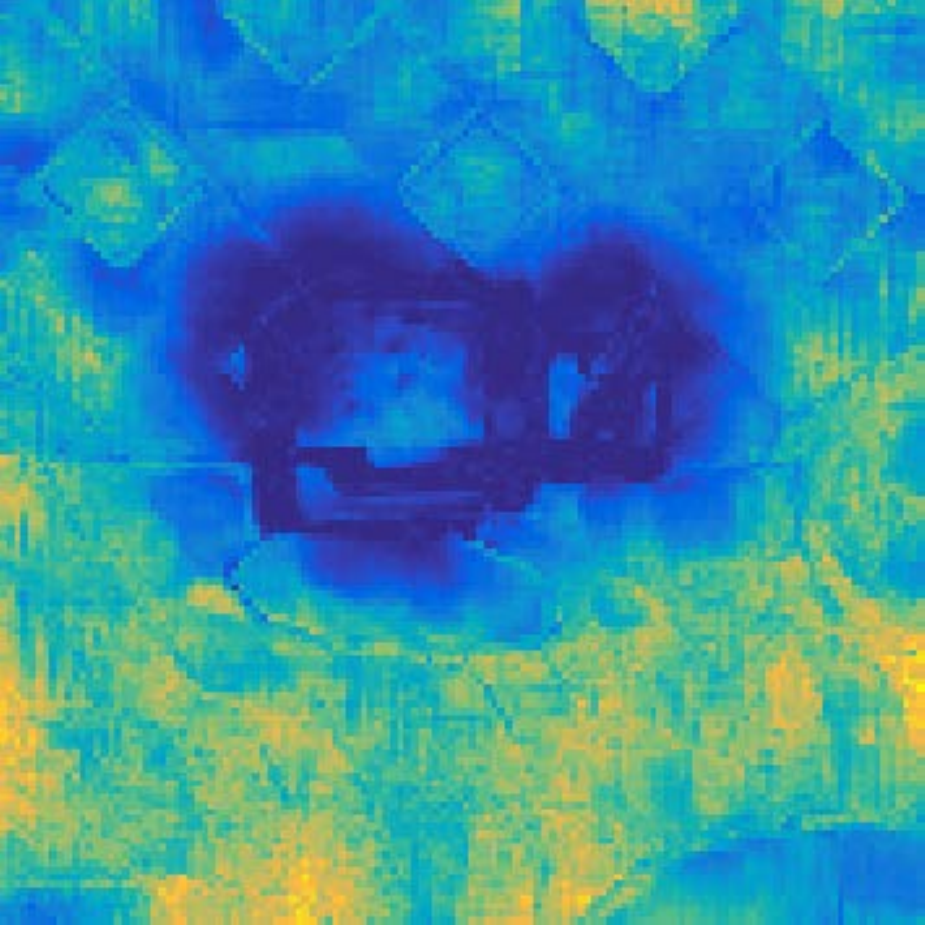}\hspace{2 mm}
\includegraphics[width=0.2\columnwidth]{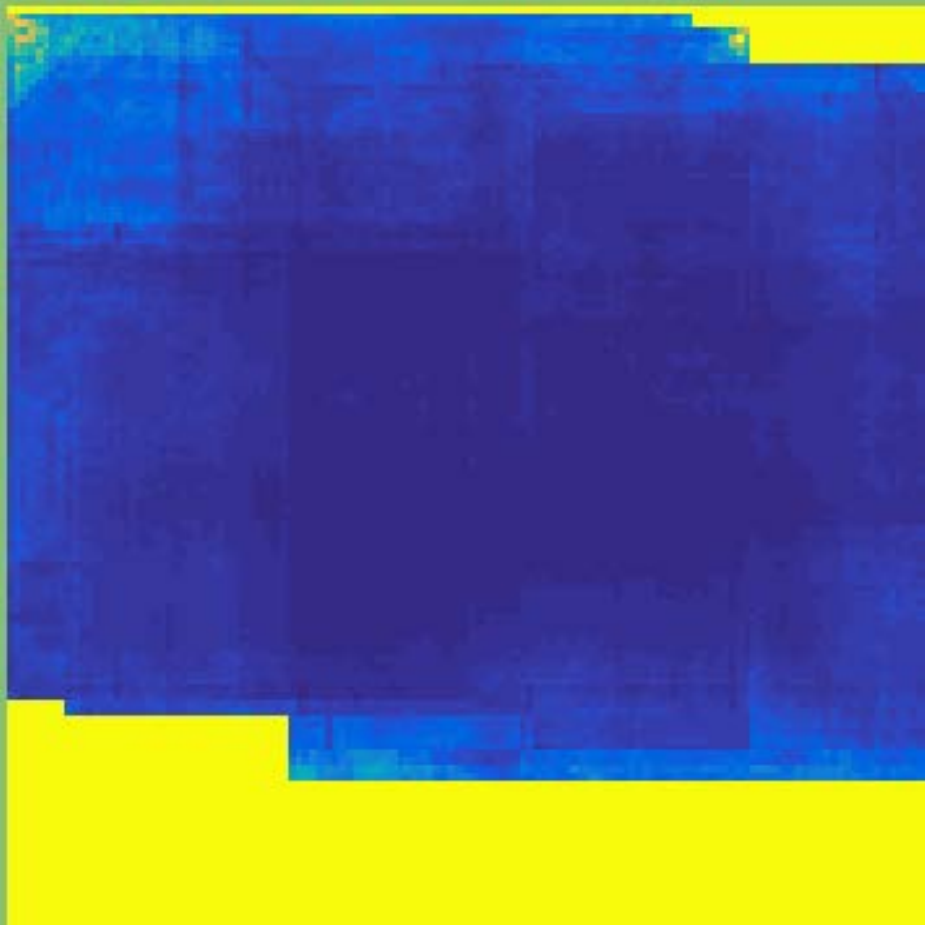}
\includegraphics[width=0.2\columnwidth]{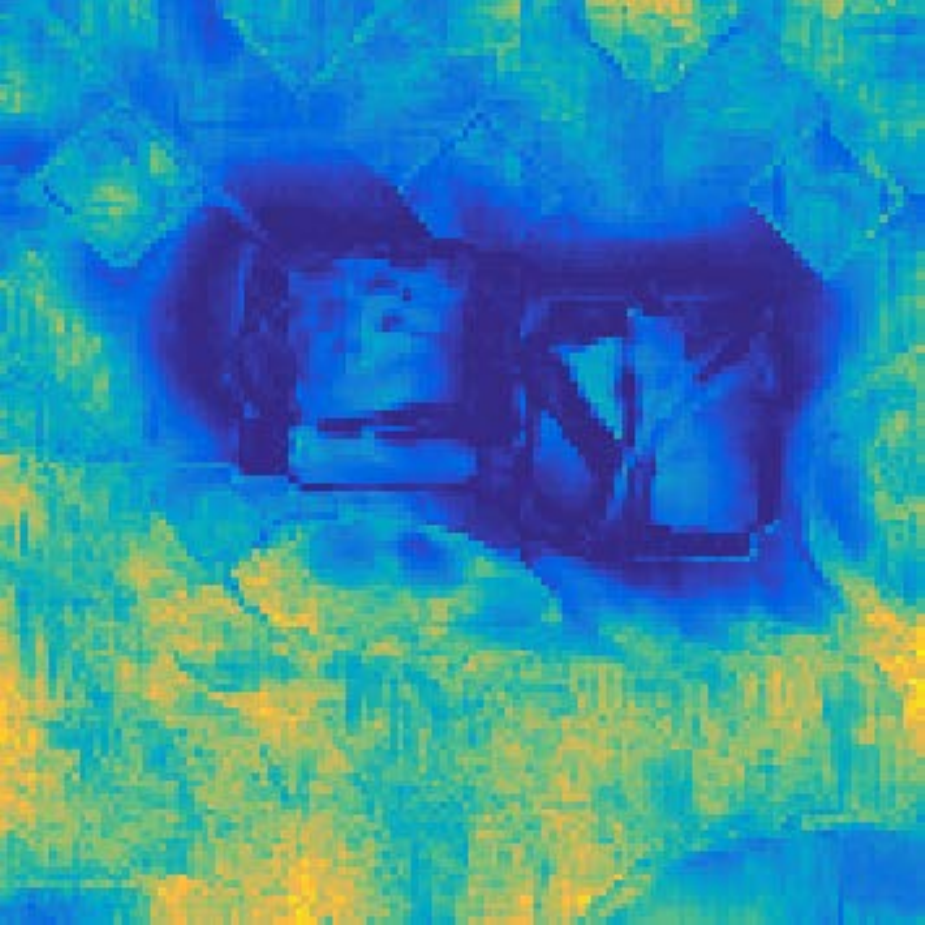}
\caption{The ratio of the highest over the second highest peak for each correlation surface over each pixel of all the artificial examples. We have three rows of images and two pairs of columns. In each column pair, the left is obtained using the classical PC methods, while the other corresponds to the BLPC. The above images demonstrate the improved accuracy of the BLPC method since we have higher peaks in the correlation surfaces and especially and the borders of the moving objects (clear shapes nd borders of the moving objects in relation to figure \ref{fig:BPCArtificialImages}). }
\label{fig:BPCArtificial02}
\end{figure}

\subsection{Proposed methodology for Bilateral-PC}
One of the main problems that we encounter with Phase Correlation is the unreliable estimates in the presence of multiple motions (i.e. motion boundaries or depth discontinuities). Considering the example shown in figure \ref{fig:FigBPC02}, where two motions are present (foreground object and the background), the obtained correlation surface is characterised by the presence of two peaks corresponding to the two motions. As a result, it is not feasible to identify the correspondence between pixels and the estimated motion vectors. This problem is more pronounced when a dense vector field (optical flow) is estimated using overlapped windows. In such cases the estimated motion vector is assigned only to the pixel located at the centre of each window. To overcome this problem a novel Bilateral Phase Correlation technique is proposed, based on the principles of Bilateral filters. The selection of this filter family is due to the higher accuracy on preserving the object or motion edges in our case. Bilateral-PC (BLPC) takes into account the difference in value with the neighbours to preserve motion edges in combination with a weighted average of nearby pixels, which allows for a pixel to influence another one not only if it occupies a nearby location but also have a similar value.

The bilateral filter is defined as a weighted average of nearby pixels, in a manner very similar to Gaussian convolution.
\begin{align}
GC[I^{i}]_{\mathbf{p}}=\sum_{\mathbf{q}\in S}G_{\sigma}(\|\mathbf{p}- \mathbf{q}\|)I^{i}_{\mathbf{q}}
\label{equ:BPC01}
\end{align}
where $i=1,2$ indicates the frame/window (see equation (\ref{equ:AA1})), $\sum_{\mathbf{q}\in S}$ denotes a sum over all image pixels indexed by $\mathbf{p}$ and $\| . \|$ represents the $L_2$ norm, e.g., $\|\mathbf{p} -− \mathbf{q} \|$ is the Euclidean distance between pixel locations $\mathbf{p}$ and $\mathbf{q}$. Also, $G_{\sigma}(x)$ denotes the 2D Gaussian kernel
\begin{align}
G_{\sigma}(x)=\frac{1}{2\pi \sigma^{2}}\exp(-\frac{x^2}{2\sigma^2})
\label{equ:BPC02}
\end{align}

\begin{table}
\centering
\caption{Results for the artificial dataset using several performance measures applying the methods on all the pixel locations.}
\begin{tabular}{c| c c c c c c c } 
\hline\hline 
 & MSE & PSNR & NRMS & AE & AEF & Time \\ 
\hline 
PC & 0.041 &   16.23  &  27.0 &   92.4 &   21.02 &   $\underline{26.9}$\\
BLPC &   $\underline{0.037}$ &   $\underline{16.47}$ &   $\underline{25.8}$  &  $\underline{70.8}$  &  $\underline{14.3}$ & 30.3\\
\hline 
\end{tabular}
\label{table:BPCArtificial01}
\end{table}

Additionally, the bilateral filter takes into account the difference in value between the neighbors to preserve edges while smoothing. Therefore, in order a pixel to influence another one, it should also have a similar value not only to be in close distance. Thus, the bilateral filter is defined as
\begin{align}
BF[I^{i}]_{\mathbf{p}}=\frac{1}{W_{\mathbf{p}}}\sum_{\mathbf{q}\in S}G_{\sigma^{i}_{s}}(\|\mathbf{p}- \mathbf{q}\|)G_{\sigma^{i}_{r}}(|I^{i}_{\mathbf{p}}- I^{i}_{\mathbf{q}}|)I^{i}_{\mathbf{q}}
\label{equ:BPC03}
\end{align}
where $| . |$ denotes absolute value and $W_{\mathbf{p}}=\sum_{\mathbf{q}\in S}G_{\sigma^{i}_{s}}(\|\mathbf{p}- \mathbf{q}\|)G_{\sigma^{i}_{r}}(|I^{i}_{\mathbf{p}}- I^{i}_{\mathbf{q}}|)$ is the normalization factor. The parameters $\sigma^{i}_{s}$ and $\sigma^{i}_{r}$ specify the amount of filtering for the image $I^{i}$. The bilateral filter is not a convolution of the spatial weight $G_{\sigma^{i}_{s}}(\|\mathbf{p}- \mathbf{q}\|)$ with the product $G_{\sigma^{i}_{r}}(|I^{i}_{\mathbf{p}}- I^{i}_{\mathbf{q}}|)I^{i}_{\mathbf{q}}$, because the range weight $G_{\sigma^{i}_{r}}(|I^{i}_{\mathbf{p}}- I^{i}_{\mathbf{q}}|)$ depends on the pixel value $I^{i}_{\mathbf{p}}$. Considering that, and selecting a fixed intensity value $I_{f}=I^{1}(w/2,h/2)$ equal to the intensity of the pixel at the centre of the selected window of size $(w,h)$, we can overcome this problem. Then, the product $G_{\sigma^{i}_{r}}(|I_{f}- I^{i}_{\mathbf{q}}|)I^{i}_{\mathbf{q}}$ is computed and convolved with the the Gaussian kernel $G_{\sigma^{i}_{r}}$, resulting the same outcome of bilateral filer at all pixels $\mathbf{p}$ with $I^{i}_{\mathbf{p}}=I_{f}$ after normalisation. Instead of using a single intensity $I_{f}$, a set of values $\{I_{f1},...,I_{fn}\}$ obtained from the neighborhood of $I_{f}$ using a $m \times m$ window (e.g. $m=3$), are selected. As a result we obtain
\begin{align}
\tilde{Q}^{i}_{k}(\mathbf{q})=G_{\sigma^{i}_{r}}(|I_{fk}- I^{i}_{\mathbf{q}}|)I^{i}_{\mathbf{q}}
\label{equ:BPC04}
\end{align}
After the convolution with the special kernel and the normalisation we have
\begin{align}
Q^{i}_{k}=(G_{\sigma^{i}_{s}}\otimes \tilde{Q}^{i}_{k}) \div (G_{\sigma^{i}_{s}} \otimes G_{\sigma^{i}_{r}})
\label{equ:BPC05}
\end{align}
where $\div$ indicates per-pixel division and denominator corresponds to the total sum of the weights. The final images $I^{B}_{1}=BF[I^{1}]_{\mathbf{p}}$ and $I^{B}_{2}=BF[I^{2}]_{\mathbf{p}}$ are estimated by upsampling and then interpolating equation (\ref{equ:BPC05}). Also $\sigma^{1}$ is selected to be much smaller in comparison to $\sigma^{2}$. An example of the obtained images is shown in figure \ref{fig:FigBPC03} and observing the outcomes, it can be seen that this approach retains pixels that are close to the centre of the window and have their intensities similar to the values of the pixels at the neighboring area of the centre.

Since the filtered images $\hat{I}^{B}_{1}$ and $\hat{I}^{B}_{2}$ are obtained and transferred to the Fourier domain, equation (\ref{Pcorr}) is used to estimate the correlation surface. Since the filtering process retains only the pixels related to the one at the centre, the correlation surface contains a single dominant peak, which yields an estimate of the shift parameters and can be recovered as
\begin{align}
(\Delta \hat{x}, \Delta \hat{y})=\arg \max\limits_{x, y} |PC(x, y)|
\label{equ:BPC06}
\end{align}
Finally, subpixel accuracy is obtained through the use of fast interpolation schemes \cite{Ren2010}. In more details, the estimate in the $x$ and $y$ directions is given by
\begin{align}
\Delta \hat{x}=\frac{D_{x}}{C(0,0)+|D_{x}|}
\Delta \hat{y}=\frac{D_{y}}{C(0,0)+|D_{y}|}
\label{equ:BPC07}
\end{align}
where $D_{x}=C(1,0)-C(-1,0)$, $D_{y}=C(0,1)-C(0,-1)$ and $C(k,l)=PC(x_{0}+k,y_{0}+l)$ with $k,l\in [-1,0,1]$.

\begin{table}[t]
\centering
\caption{The evaluation results for the 4K Test Sequences Reviving the Classics dataset \cite{Elemental2014} without ground truth. using several performance measures.}
\begin{tabular}{c| c c c c } 
\hline\hline 
Measures & MSE & PSNR & NRMS &  Time(s)  \\ 
\hline 
Thomas \cite{Thomas1987}    &  5.492 &  24.682 & 199.55  & \underline{44.822} \\
Argyr. \cite{Argyriou2009}  &  5.453 &  24.747 & 198.83  & 45.591 \\
Gaut. \cite{Gautama2002}  &  21.89 &  17.847 & 408.68  & 236.68 \\
Yan \cite{Yan2008}    &  5.498 &  24.679 & 199.67  & 2594.7 \\
Ho \cite{Ho2008}  &  3.263 &  26.753 & 159.56  & 291.64 \\
Reyes \cite{Reyes2013}    &  1.934 &  29.240 & 121.72 & 54.569 \\
Alba \cite{Reyes2014}    &  2.043 &  28.775 & 126.81 & 66.483 \\
BLPC   &  \underline{0.908} &  \underline{32.680} &  \underline{81.64} & 48.669  \\
\hline 
\end{tabular}
\label{table:BPCReal01}
\end{table}

\begin{table}[t]
\centering
\caption{The evaluation results for the dataset with ground truth provided by Baker \cite{Baker2007} using several performance measures}
\begin{tabular}{c| c c c c c c } 
\hline\hline 
 & MSE & PSNR & NRMS & AE & AEF & Time \\ 
\hline 
Tho \cite{Thomas1987} &   2.30 &   21.24  &  40.31  &  62.0  &   4.63  &   9.71 \\
Arg \cite{Argyriou2009}  &    2.34  &  21.31  &  40.25 &   60.6  &   4.62 &   11.47 \\
Gau \cite{Gautama2002}  &    2.40 &   17.45  &  63.76 &   25.1  &   8.22 &    8.40 \\
Yan \cite{Yan2008} &   2.31 &   21.22 &   40.41 &   62.4 &    4.64 &    9.78 \\
Ho \cite{Ho2008}  &    3.14  &  24.43 &   31.31  &  23.4 &    3.88 &   74.3 \\
Rey \cite{Reyes2013} &     1.97 &   26.42 &   20.88 &   11.2 &  3.38 & \underline{2.72} \\
Alb \cite{Reyes2014} &     2.06 &   26.25 &   21.64 &   10.2 &    3.38 &  7.59 \\
BLPC  &    \underline{1.97} &   \underline{26.68} &   \underline{20.57} & \underline{9.9} & \underline{3.37}  &   2.96 \\
\hline 
\end{tabular}
\label{table:BPCReal02}
\end{table}

\begin{table}[t]
\centering
\caption{The evaluation results for samples of videos with faces from Dhall's dataset \cite{Dhall2012} without ground truth using several performance measures.}
\begin{tabular}{c| c c c c } 
\hline\hline 
Measures & MSE & PSNR & NRMS & Time(s)  \\ 
\hline 
Thomas \cite{Thomas1987}   &     1.4471   &     33.082    &    21.108  &      12.784 \\
Argyr. \cite{Argyriou2009}  &     1.4352   &     33.093    &    21.046  &      16.009 \\
Gaut. \cite{Gautama2002}  &    23.143   &     18.382    &    78.725   &     10.452 \\
Yan \cite{Yan2008} &     1.4488   &      33.08    &    21.117 &        13.63 \\
Ho \cite{Ho2008}    &     1.081   &      33.81    &    18.621  &      132.56 \\
Reyes \cite{Reyes2013}   &     0.3299   &     36.999    &    10.218  &      1.3671 \\
Alba \cite{Reyes2014}  &    0.29531   &     36.891    &   10.343   &     8.6594 \\
BLPC & \underline{ 0.2595} &  \underline{37.419}  &  \underline{9.0248} & \underline{0.9956} \\
\hline 
\end{tabular}
\label{table:BPCReal03}
\end{table}

\begin{table}[t]
\centering
\caption{The evaluation results for the the CT dataset from Liang \cite{Liang2014,Liang2015} without ground truth using several performance measures.}
\begin{tabular}{c| c c c c } 
\hline\hline 
Measures & MSE & PSNR & NRMS & Time(s)  \\ 
\hline 
Thomas \cite{Thomas1987}  &     1.7381  &      28.755     &   25.455   &     13.526 \\
Argyr. \cite{Argyriou2009} &    1.7381   &     28.756 &      25.455   &     16.661 \\
Gaut. \cite{Gautama2002}  &     7.2814   &     21.807  &      54.722  &      10.535 \\
Yan \cite{Yan2008} &      1.7382   &     28.755 &       25.455  &      17.148 \\
Ho \cite{Ho2008} &    1.8028   &     28.648  &      25.882  &      131.35 \\
Reyes \cite{Reyes2013} &   1.6309   &     28.947 &       24.724  &     0.9989 \\
Alba \cite{Reyes2014} &     1.639     &    28.93  &      24.784  &      6.2942 \\
BLPC  &    \underline{1.6283}  &       \underline{28.95}   &     \underline{24.705 } &      \underline{0.4863} \\
\hline 
\end{tabular}
\label{table:BPCReal04}
\end{table}

\begin{table}[t]
\centering
\caption{The evaluation results for the UCLgtOFv1.1 dataset with ground truth \cite{Aodha2010} using several performance measures.}
\begin{tabular}{c| c c c c c c c } 
\hline\hline 
 & MSE & PSNR & NRMS & AE & AEF & Tim  \\ 
\hline 
Tho \cite{Thomas1987} & 19.2  &  17.51  &    74.68   &   80.5  &  19.57   &   11 \\
Arg \cite{Argyriou2009} & 19.1  &  17.53  & 74.63   & 79.9  &   19.56   &   10.9 \\
Gau \cite{Gautama2002}  &  30.4 & 15.62  &  96.31  & 66.7  &  76.03  &  7.9 \\
Yan \cite{Yan2008} &   19.2  &  17.53  &  74.75  &  80.8   &   19.6  &   9.6 \\
Ho \cite{Ho2008}  &    16.2  & 18.58  &  68.89 &   69.4  &  18.7   &    91 \\
Rey \cite{Reyes2013} &  7.5  & 22.04  &  47.11 & 30.8 & 11.33 & \underline{1.5} \\
Alb \cite{Reyes2014}  &   8.5  &  21.45  & 50.51  &   34.3  &  12.29   &   9.4 \\
BLPC &  \underline{7.2}   &   \underline{22.82} &  \underline{46.19} & \underline{23.3}  &  \underline{7.44}  &   1.6\\
\hline 
\end{tabular}
\label{table:BPCReal05}
\end{table}

\subsection{Optical flow estimation framework}

In this section, we present the proposed optical flow estimation framework using the introduced Bilateral-PC. The overall methodology is separated into the following stages as it is shown in figure \ref{fig:FigBPC05}. This approach is based on a multi-resolution coarse-to-fine algorithm that constructs an image pyramid by down-sampling the original image into a maximum of three layers by a factor of power of two. The optical flow can then be estimated initially for the smallest image pair in the pyramid, and is used to unwarp the next one by up-sampling and scaling the previous layer.

The first stage of this framework includes all the pre-processing tasks, such as pyramid formation, estimation of key points and removal of the less significant ones. Initially, the image $I_{i}$ is down-sampled by a factor of power of two until the obtained resolution is below a threshold $M_{min}\times N_{min}$ that depends on the available computational power or the desired complexity of the system. If the down-sampling process was applied more than once before we reach the minimum resolution threshold, one more layer is extracted in-between the smallest one and the original size image. Since, the three layers are obtained (small $I^{s}_{i}$, medium $I^{m}_{i}$, and the original $I^{o}_{i}$), key points are estimated on the smallest one.
Let $\textbf{p}_{u}=[x,y]^T\in\mathcal{R}^2$ be a set of point locations in $I^{s}_{i}$, selected by applying a uniform sampling using a step $s_{u}$ (see figure \ref{fig:FigBPC06}). Also, the absolute frame difference $I^{s}_{d}=|I^{s}_{i}-I^{s}_{i+1}|$ is calculated and then a second set of points $\textbf{p}_{d}=[x,y]^T\in\mathcal{R}^2$ is obtained as
\begin{align}
(x, y)=\arg\limits_{x, y} \{I^{s}_{d}(x,y)\} \textrm{, if }I^{s}_{d}(x,y)\geq T_{d}
\label{equ:BPC08}
\end{align}
where $T_{d}=\alpha\sigma_{I^{s}_{d}}$ and $\alpha$ is a scaling factor (see figure \ref{fig:FigBPC06}). The impact of this threshold is proportional to the overall complexity, and adjusting that value can improve or reduce the overall quality with an impact on the computational complexity. The selected values experimentally proved that provide a good balance between the overall accuracy and complexity for all video sequences.

Since the key points $\textbf{p}_{u}$ and $\textbf{p}_{d}$ are estimated, if their total number is above a limit $T_{p}$, that depends on the desired complexity of the proposed optical flow algorithm, the $\textbf{p}_{d}$ are uniformly down-sampled. A sampling step is selected to result a new set of points $\textbf{p}^{'}_{d}$, aiming to have in total less points than $T_{p}$ by combining both $\textbf{p}_{u}$ and $\textbf{p}^{'}_{d}$.

During the second stage of the proposed methodology, motion estimation is applied at the selected key locations. Initially, Phase Correlation is applied using a square window of size $(m_{w}\times m_{w})$ centered at each key point location. From the obtained correlation surface $\textrm{PC}(\textbf{u})$ in equation (\ref{Pcorr}), the ratio of the highest $P^{1}_{\textrm{PC}}$ over the second highest $P^{2}_{\textrm{PC}}$ peak is estimated $r_{\textrm{PC}}=P^{1}_{\textrm{PC}}/P^{2}_{\textrm{PC}}$. If $r_{\textrm{PC}}\geq T_{r}$, then this is an indication that more than one motions are present in this window. Therefore, Bilateral-PC is now applied to obtain a correlation surface with a single peak and a consequently more accurate motion estimate. The threshold $T_{r}$ is specified based on the logarithm of the current window size $m_{w}$. Additionally, for the points that were removed after down-sampling
\begin{align}
\hat{\textbf{p}}_{d} = \textbf{p}_{d} \bigcap \textbf{p}^{'}_{d}
\label{equ:BPC09}
\end{align}
motion vectors are estimated using \cite{Lucas1981}. In order to obtain a dense vector field non-uniform interpolation with bilateral filtering is applied on the motion components of the sparse key point locations.


\begin{figure}[htb]
\centering
\includegraphics[width=0.15\columnwidth]{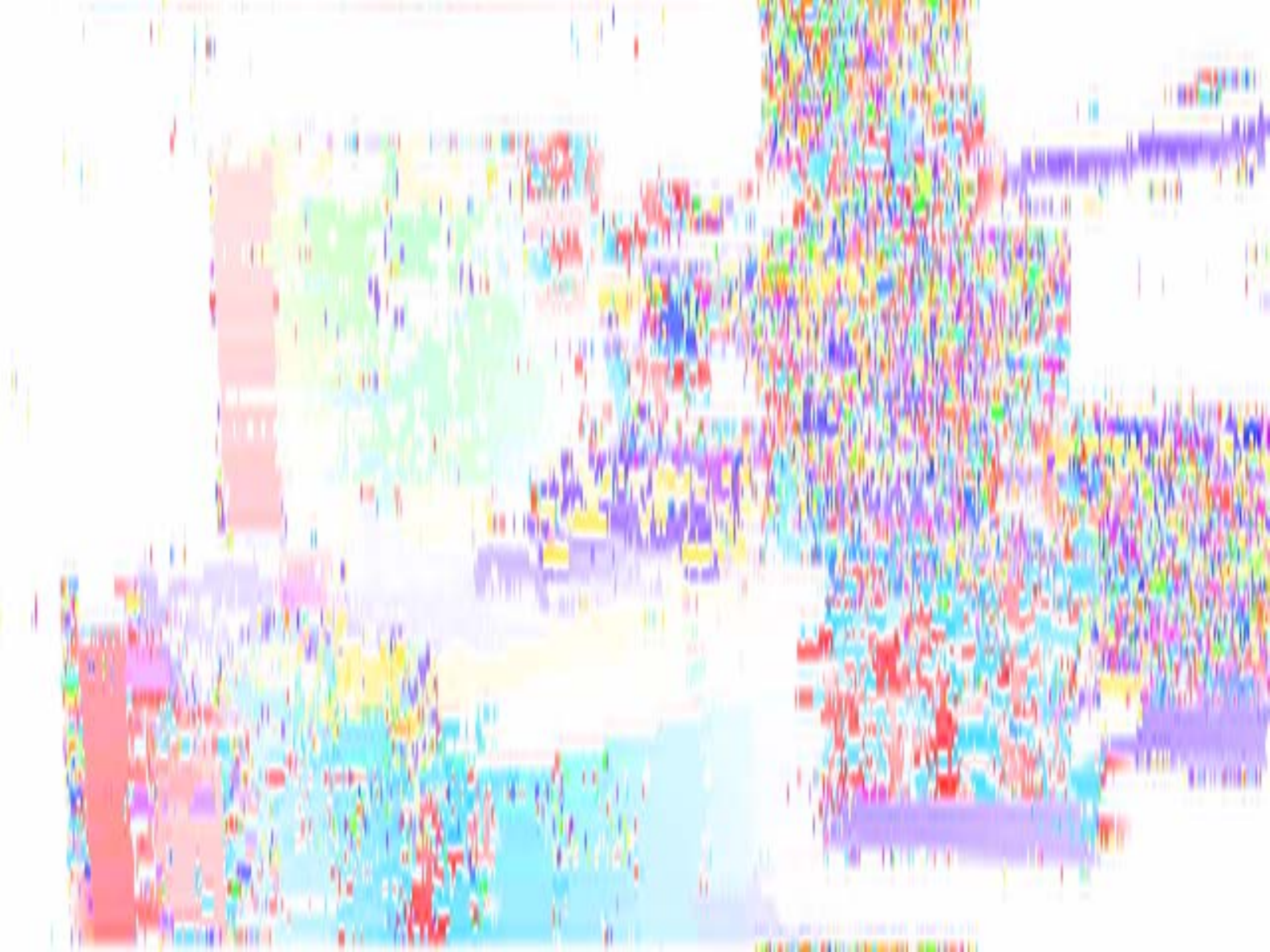}
\includegraphics[width=0.15\columnwidth]{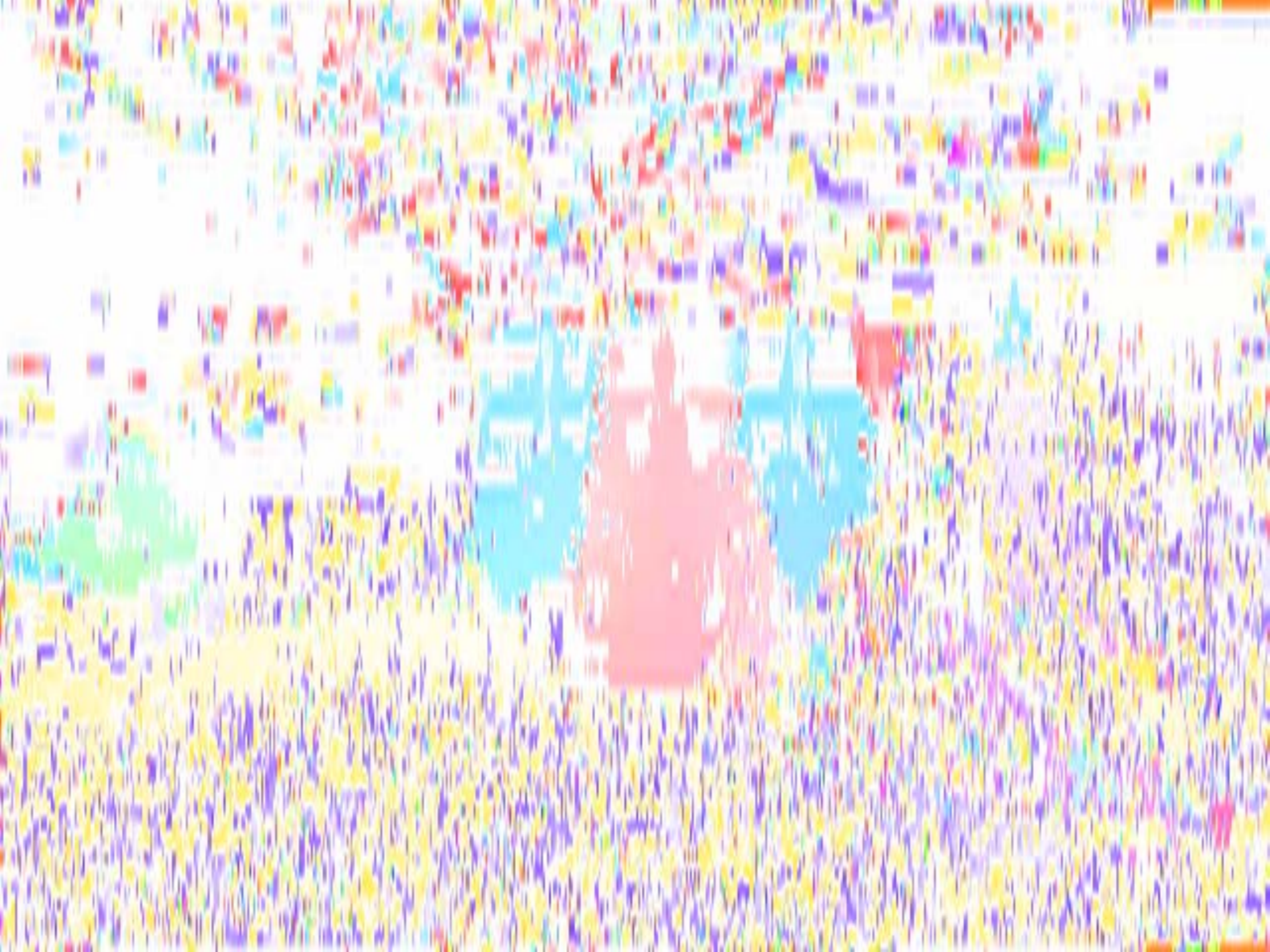}
\includegraphics[width=0.15\columnwidth]{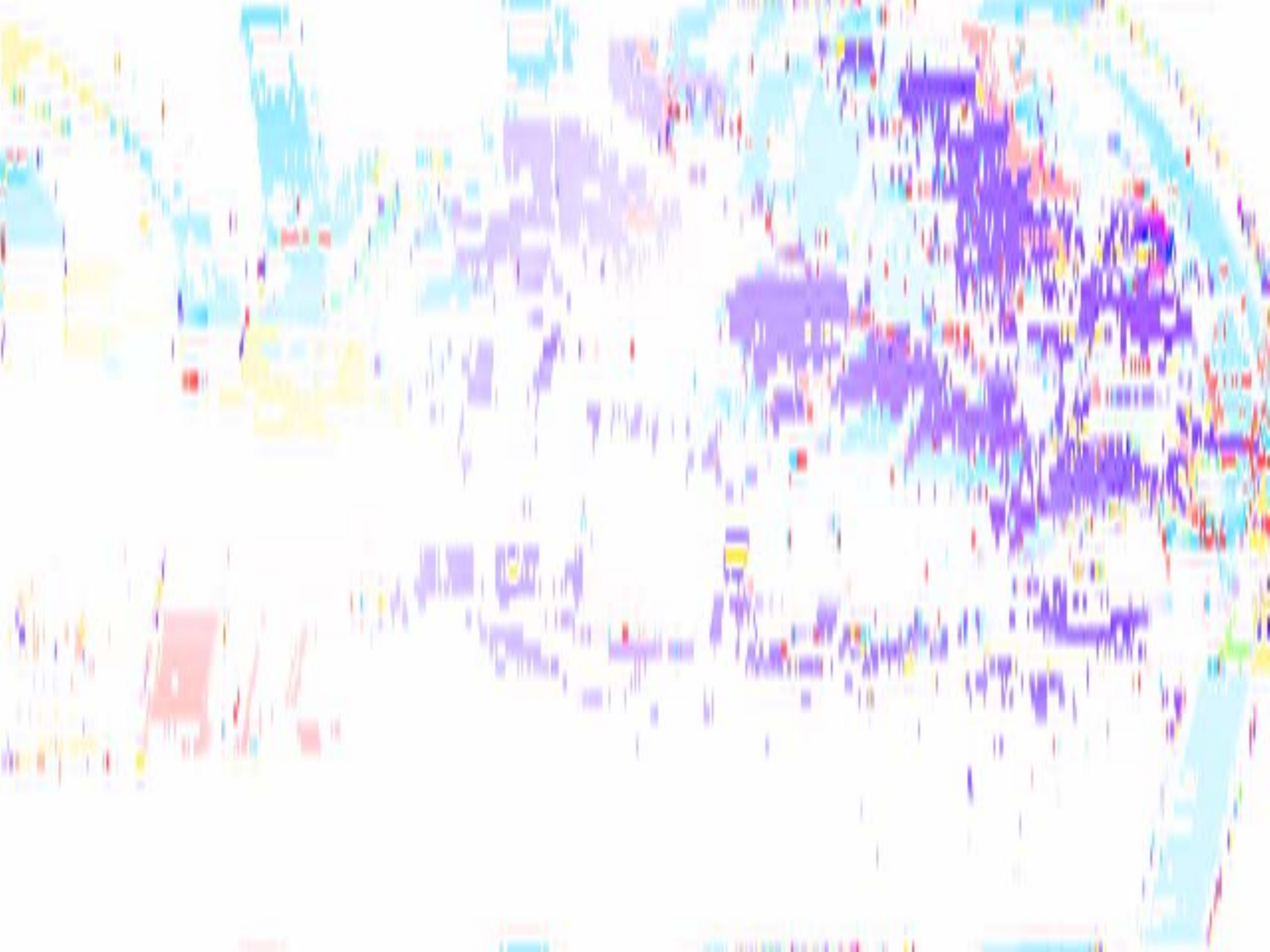}
\includegraphics[width=0.15\columnwidth]{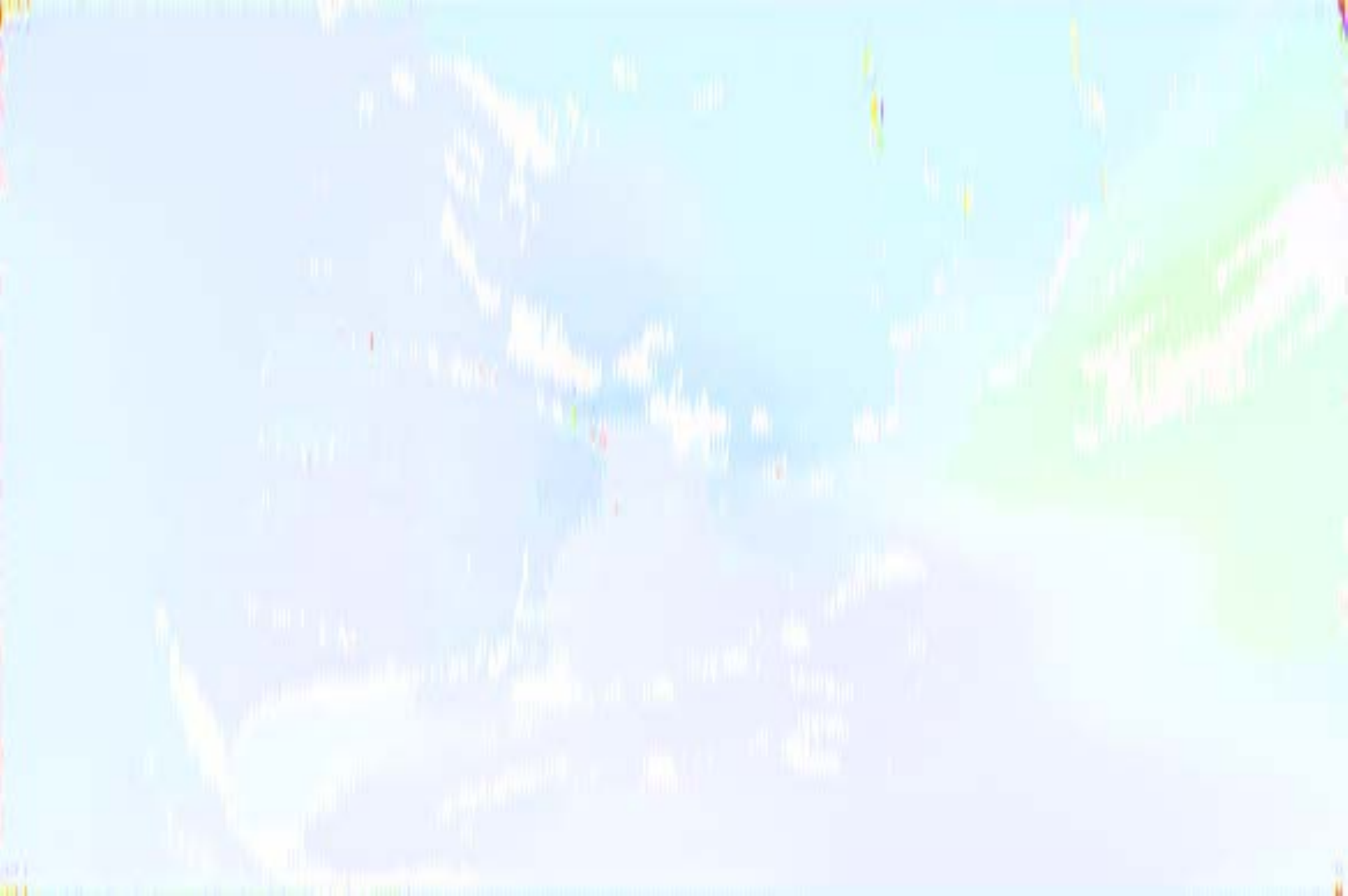}
\includegraphics[width=0.15\columnwidth]{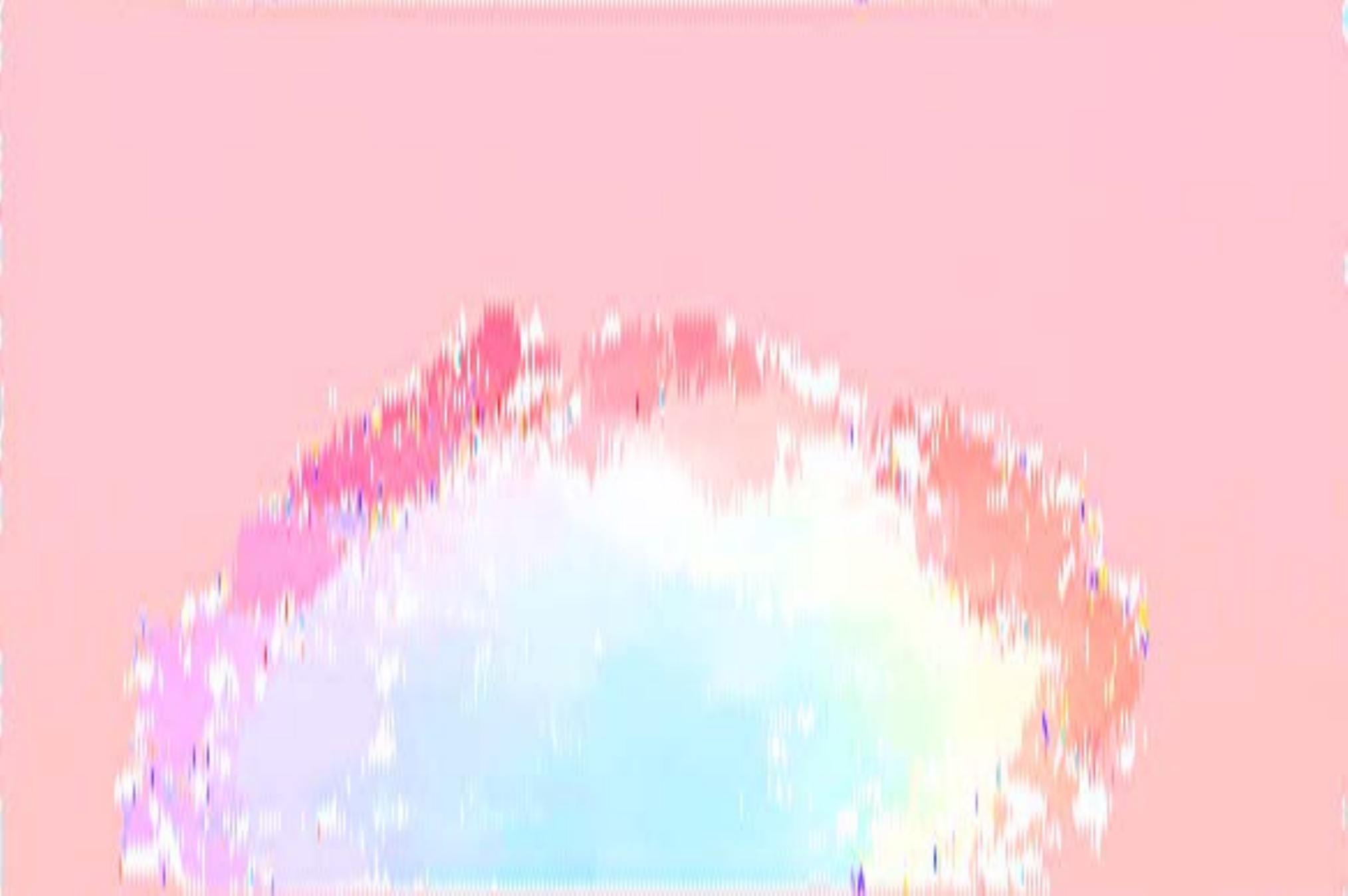}
\includegraphics[width=0.15\columnwidth]{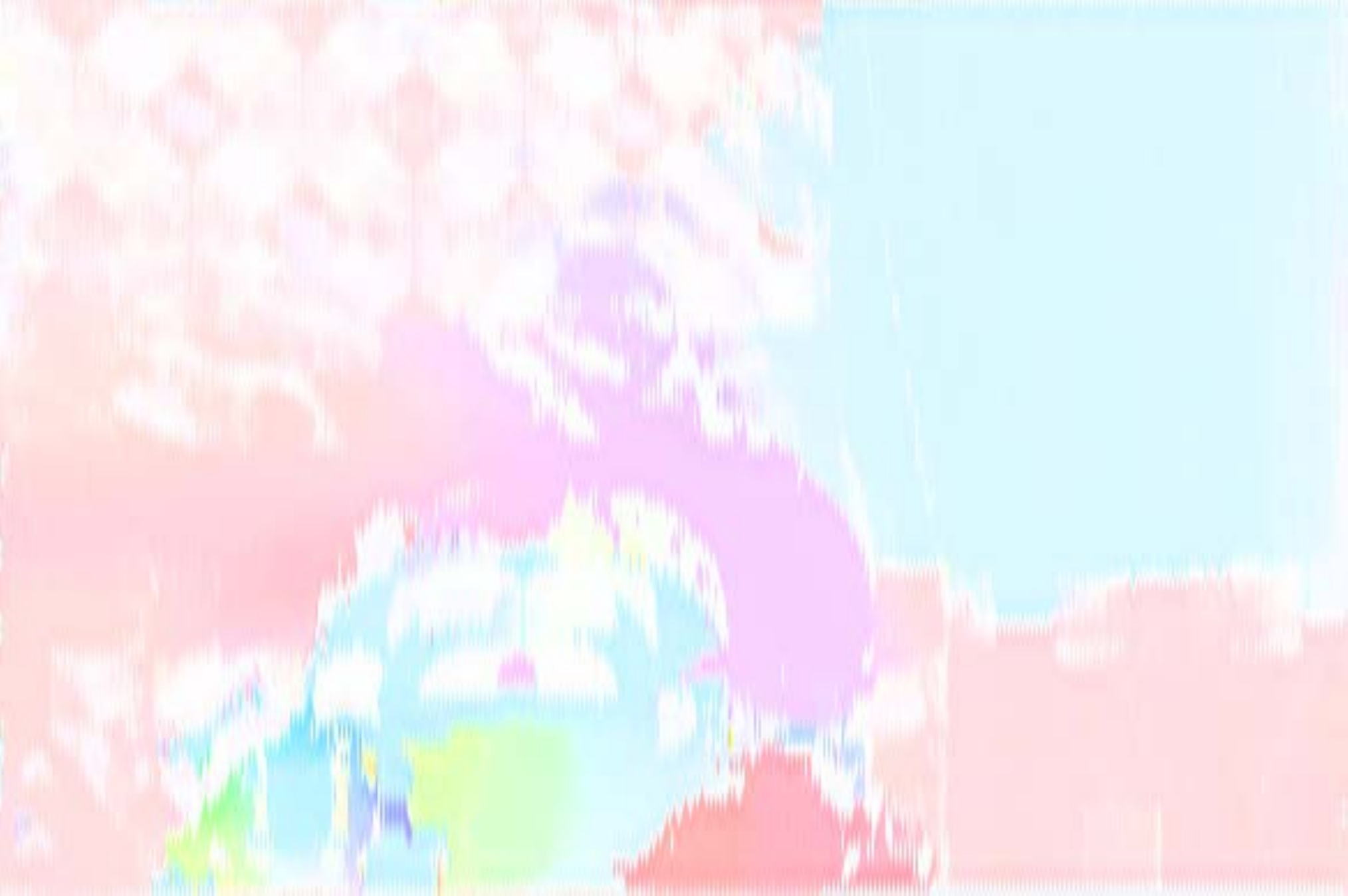}\\ \vspace{1 mm}
\includegraphics[width=0.15\columnwidth]{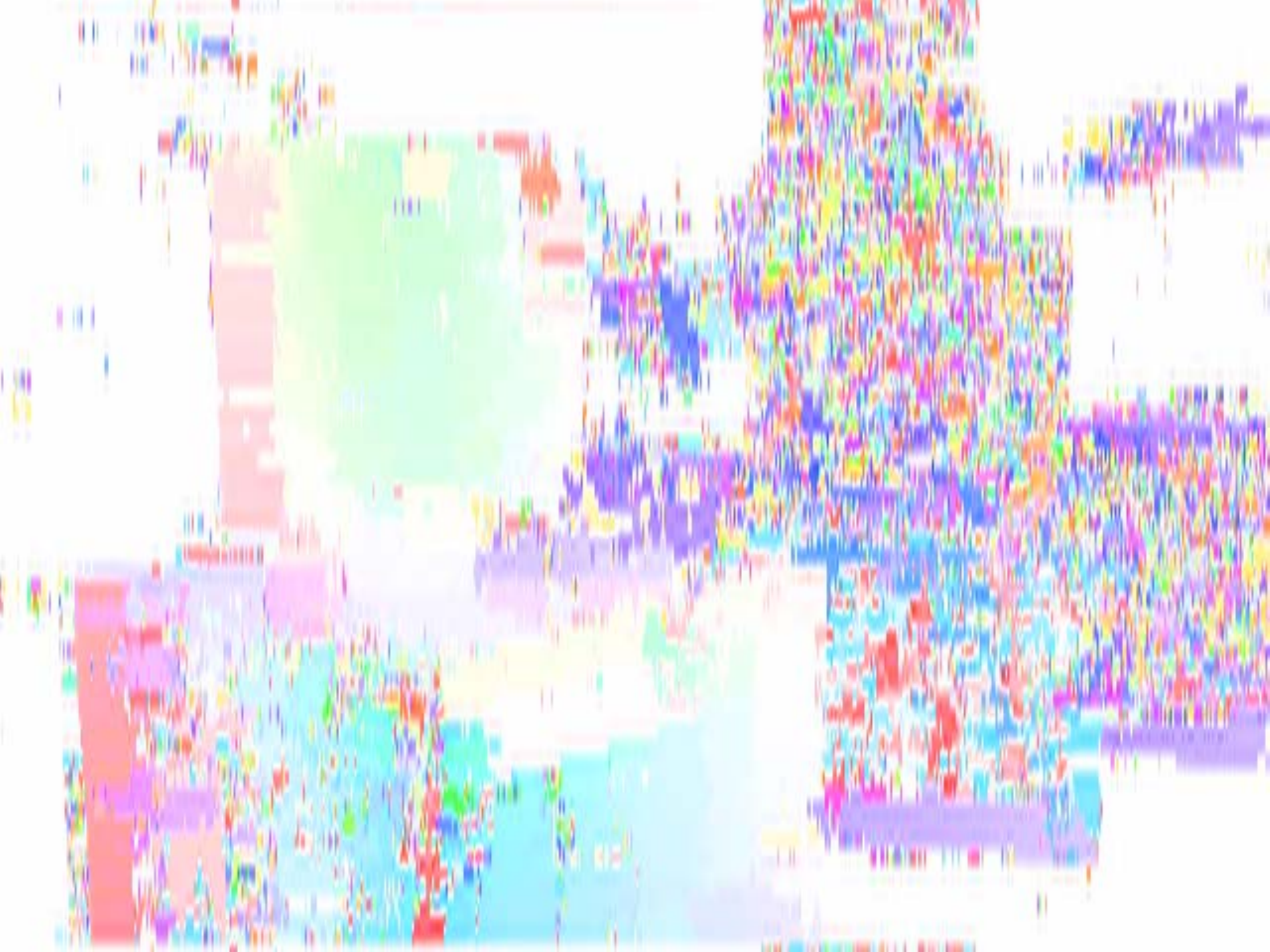}
\includegraphics[width=0.15\columnwidth]{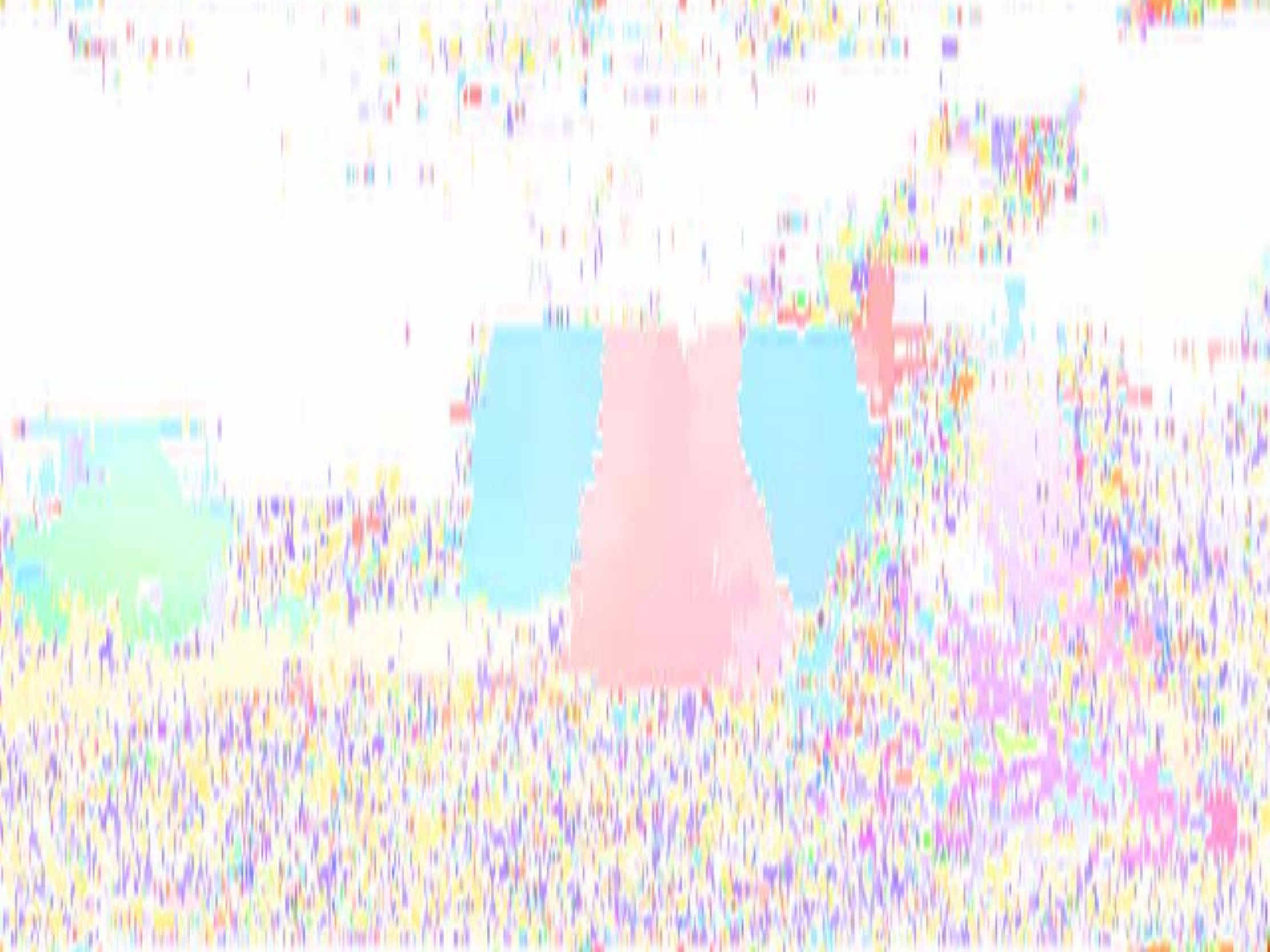}
\includegraphics[width=0.15\columnwidth]{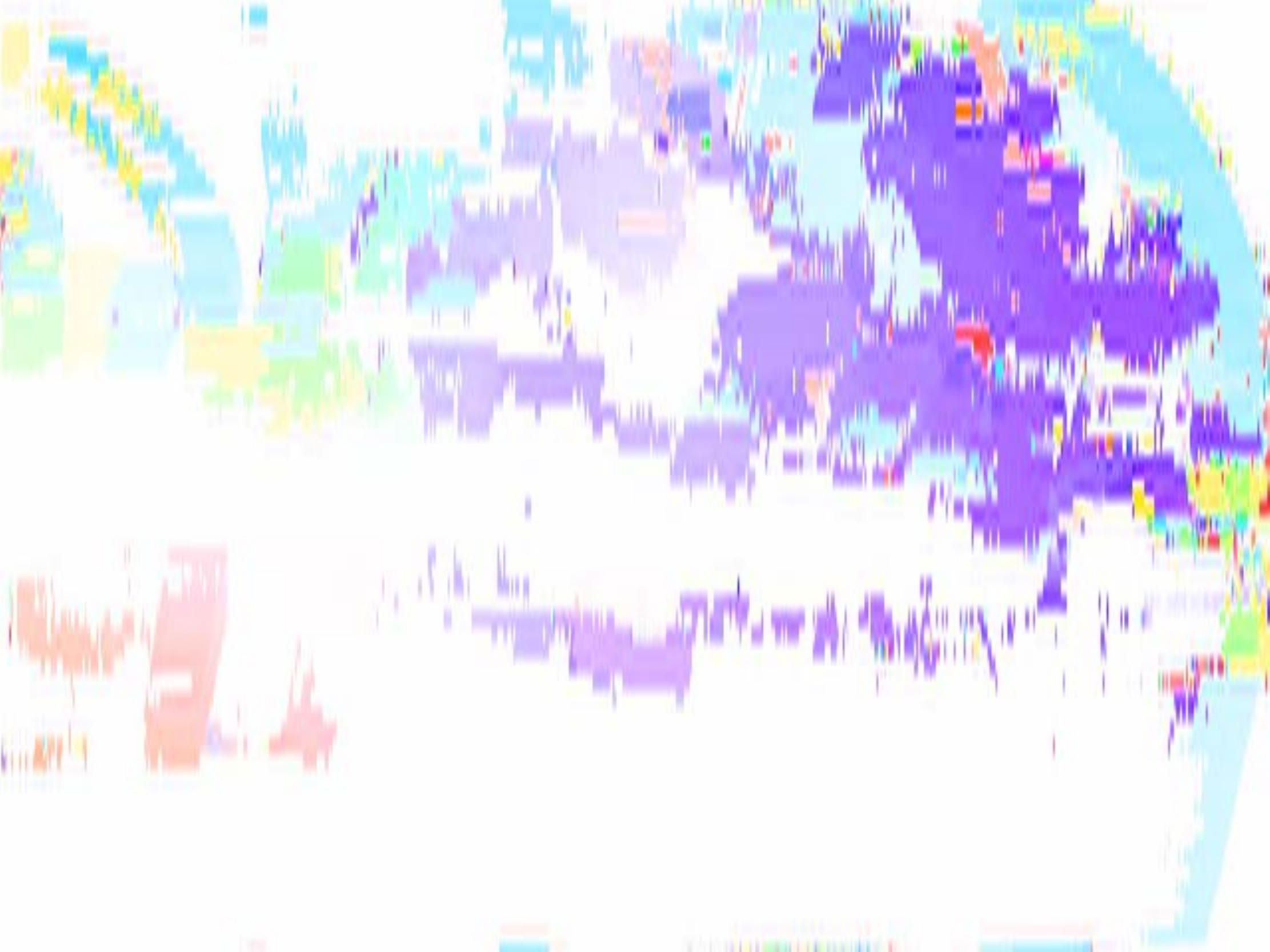}
\includegraphics[width=0.15\columnwidth]{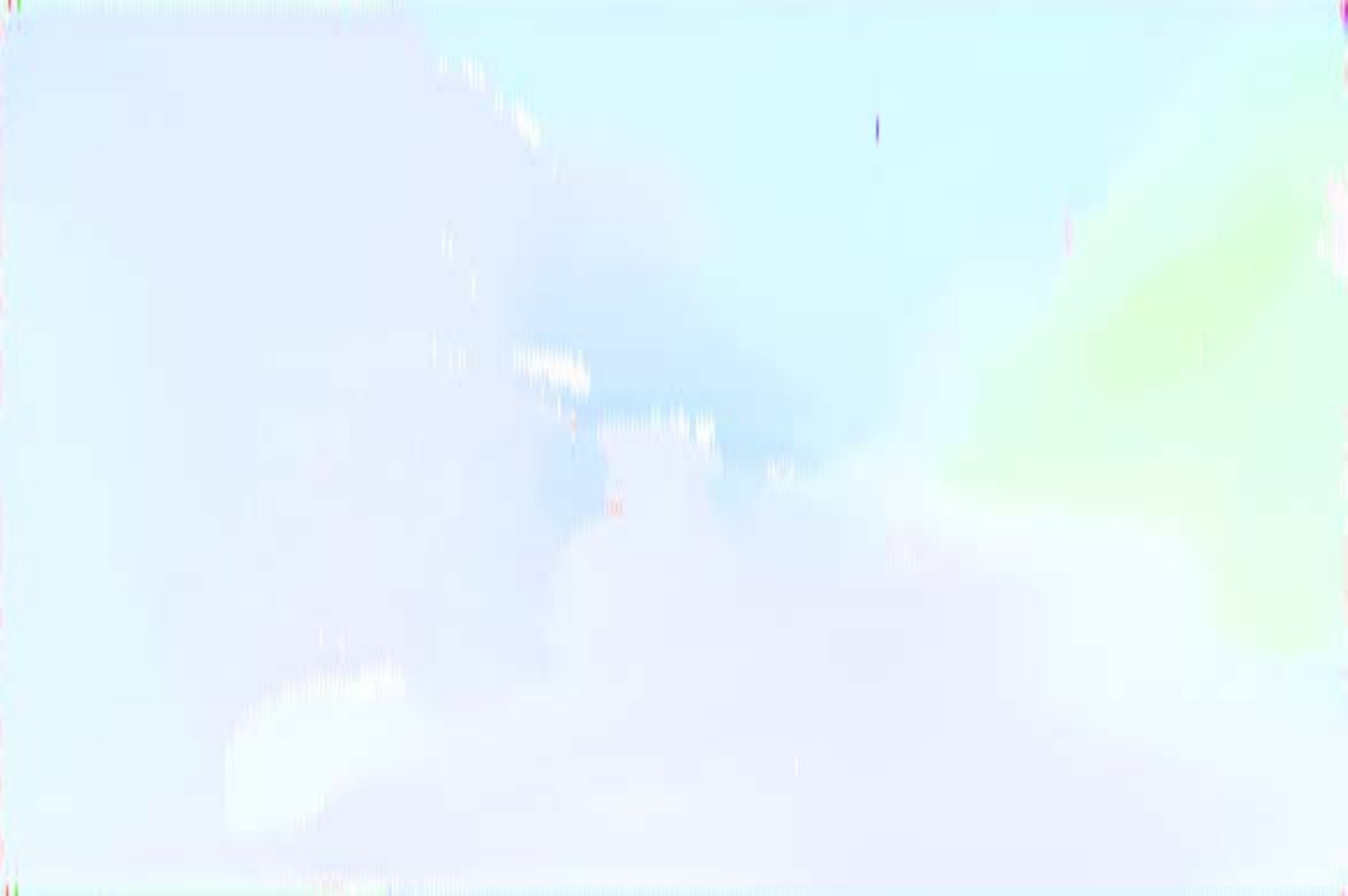}
\includegraphics[width=0.15\columnwidth]{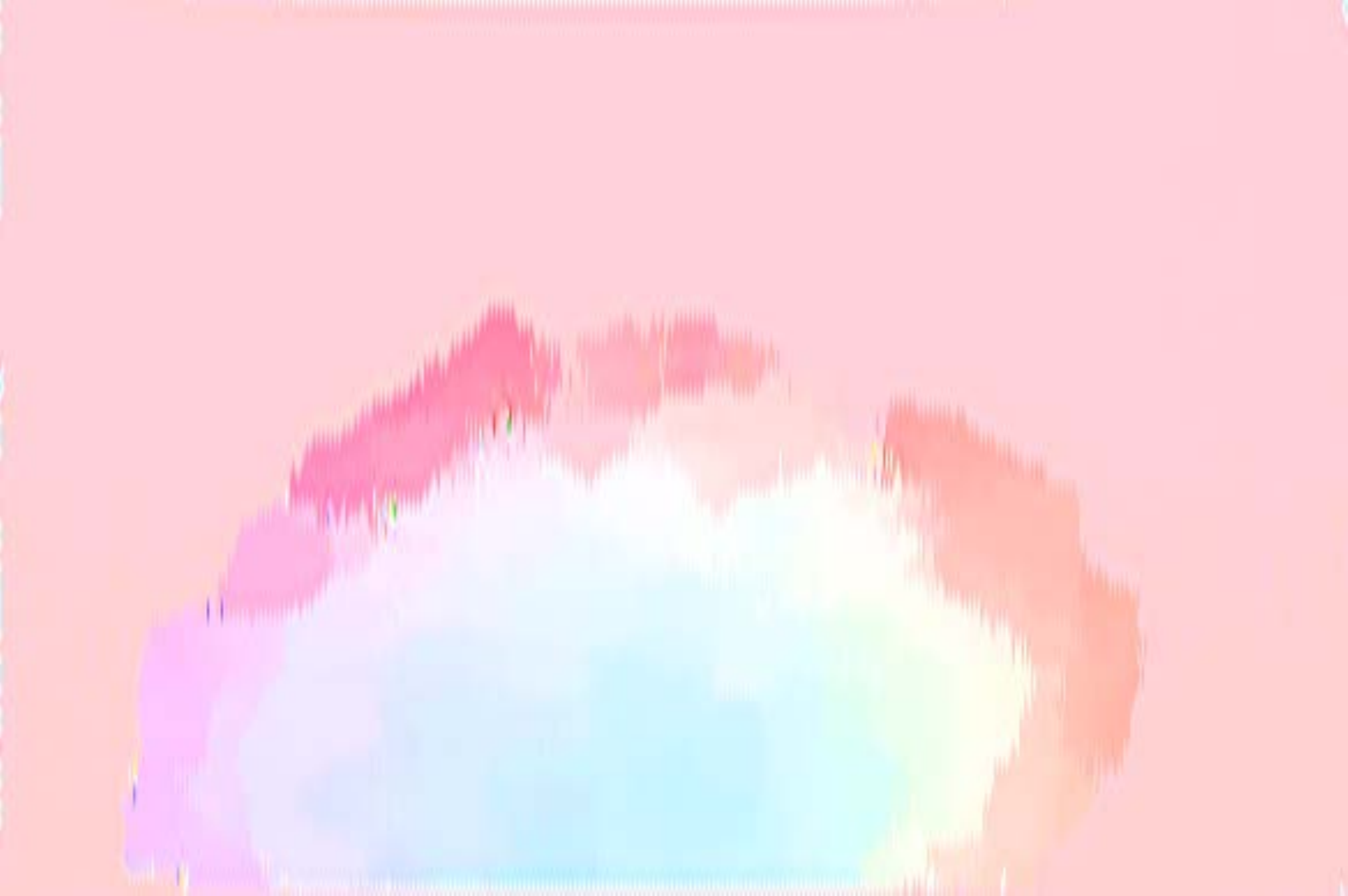}
\includegraphics[width=0.15\columnwidth]{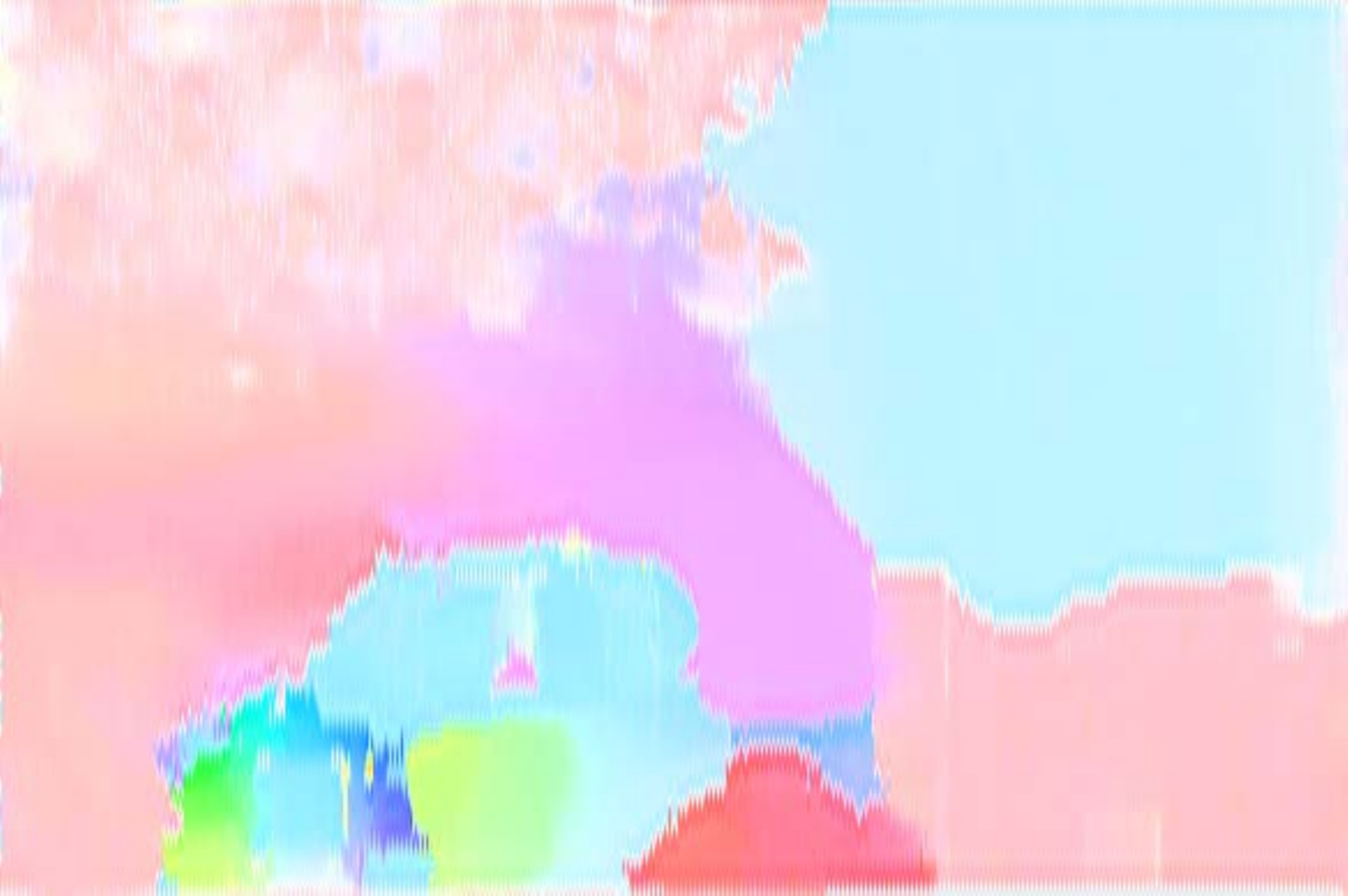}\\ \vspace{1 mm}
\includegraphics[width=0.15\columnwidth]{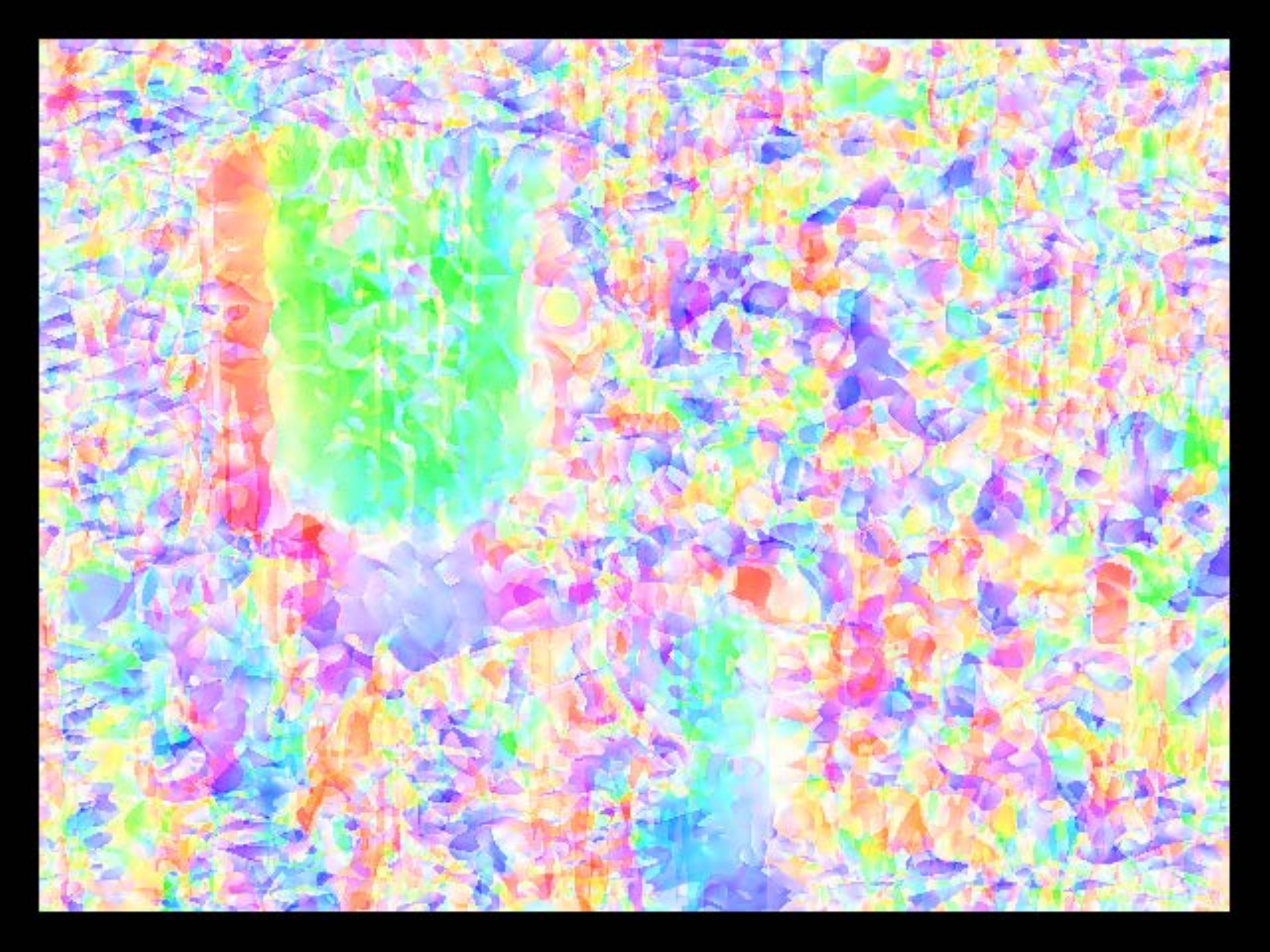}
\includegraphics[width=0.15\columnwidth]{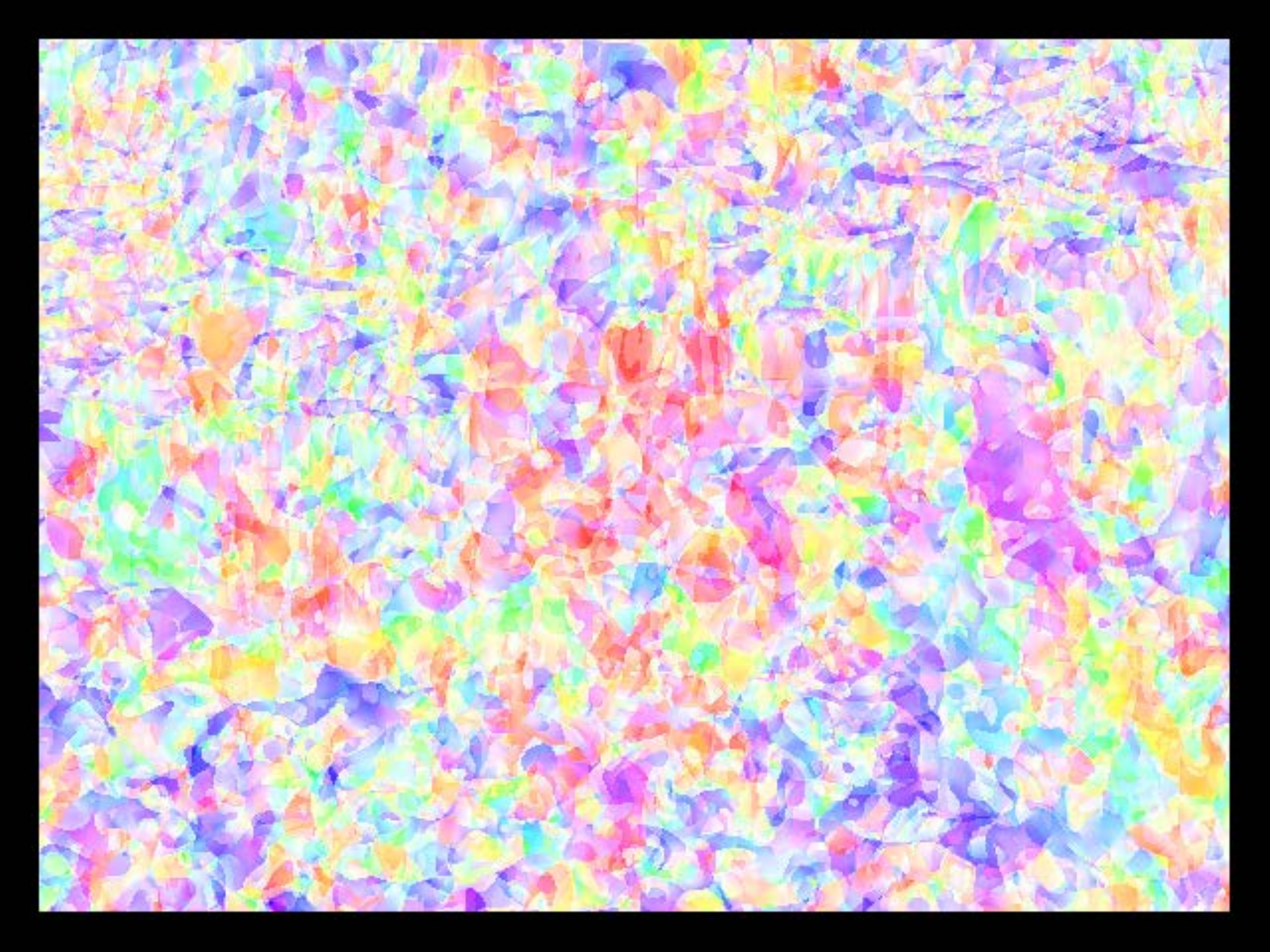}
\includegraphics[width=0.15\columnwidth]{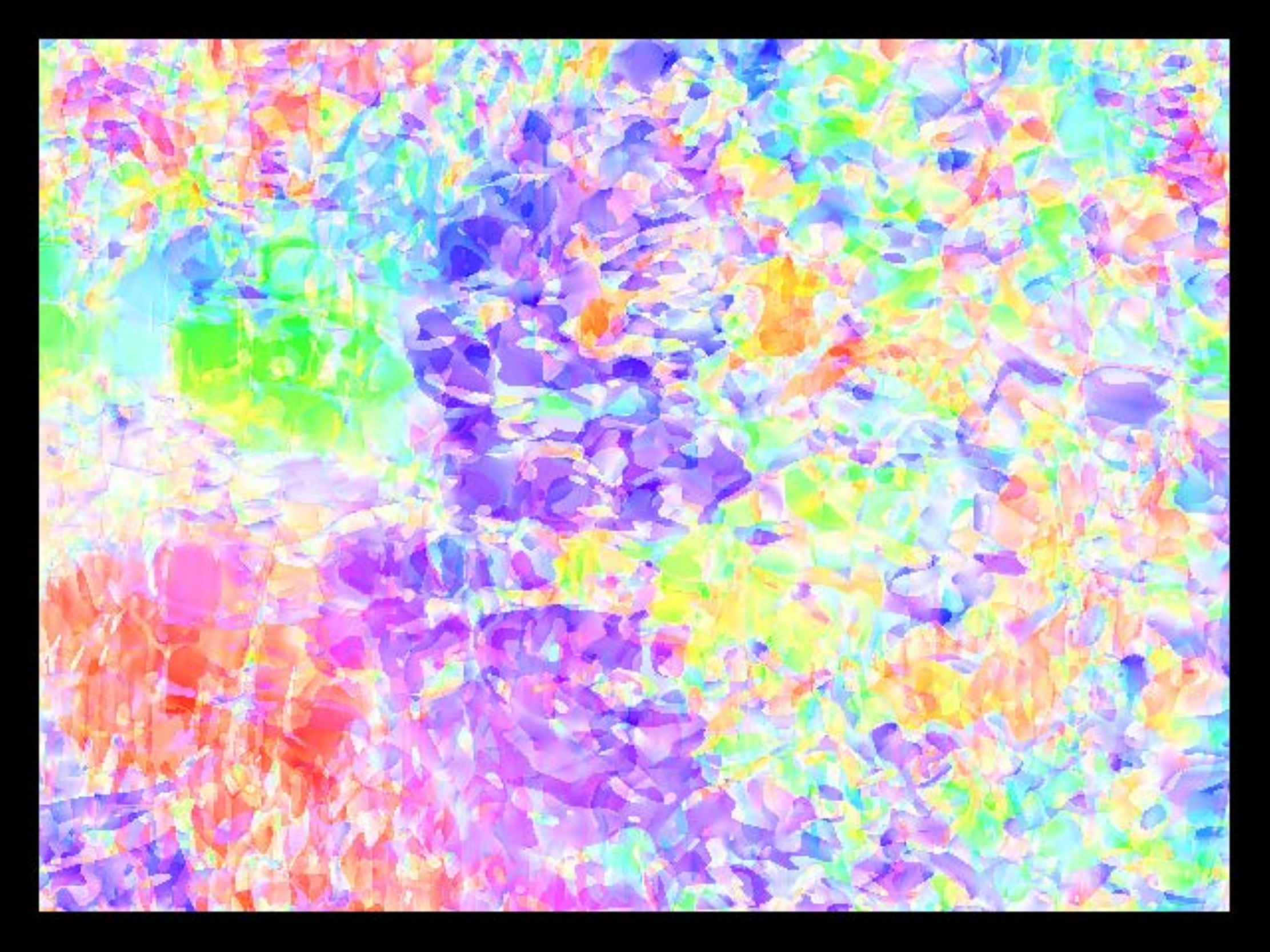}
\includegraphics[width=0.15\columnwidth]{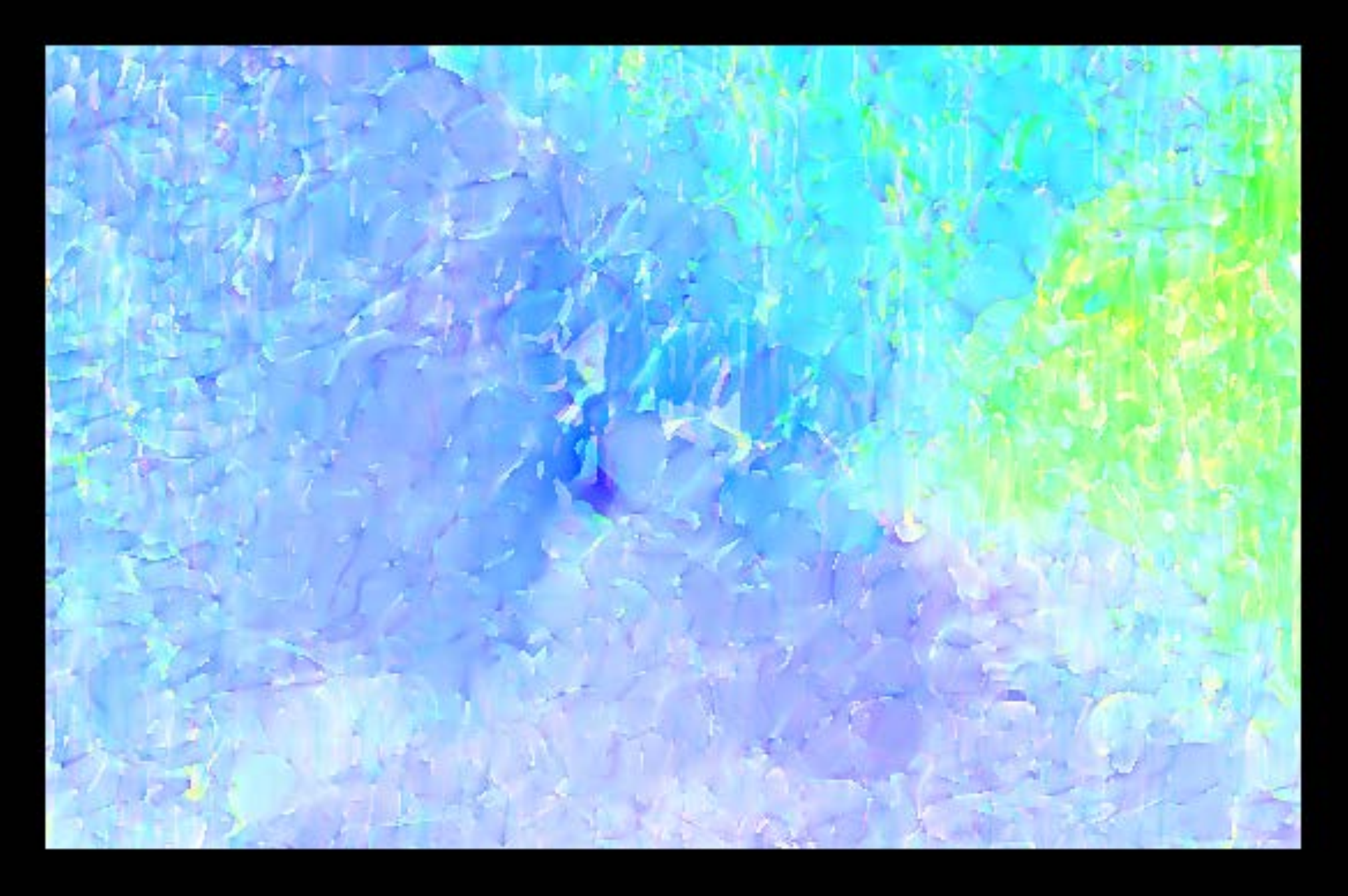}
\includegraphics[width=0.15\columnwidth]{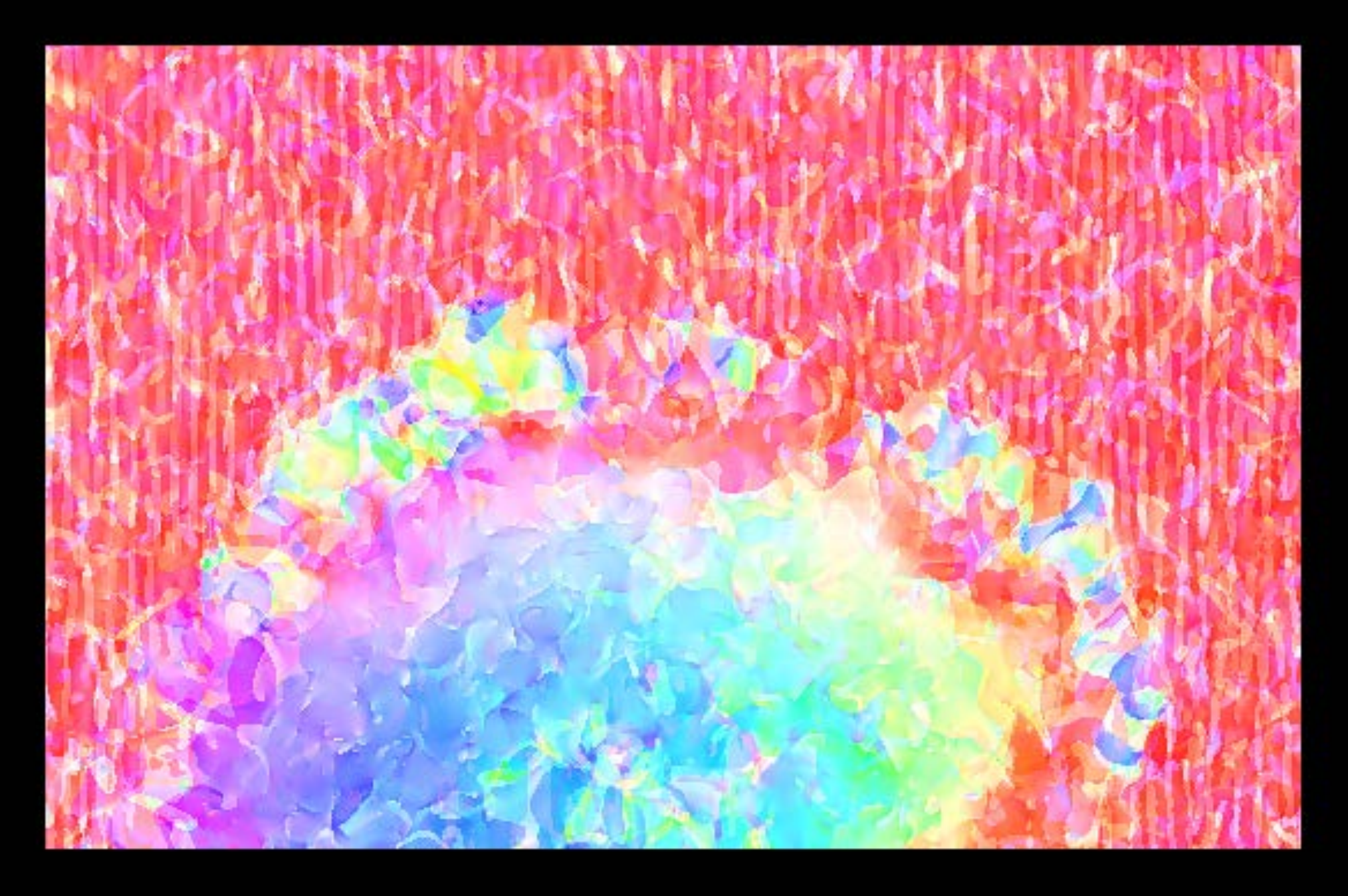}
\includegraphics[width=0.15\columnwidth]{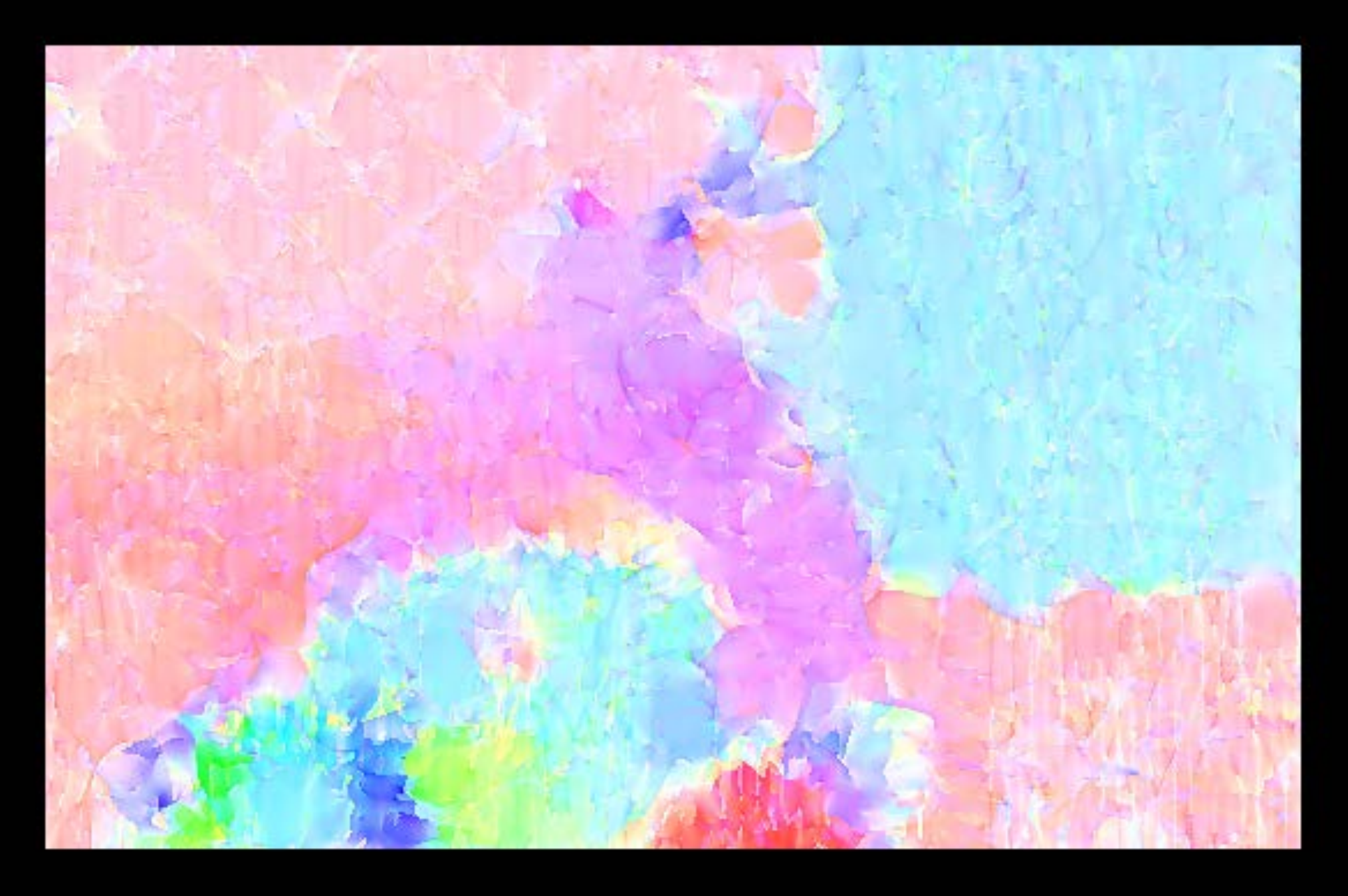}\\ \vspace{1 mm}
\includegraphics[width=0.15\columnwidth]{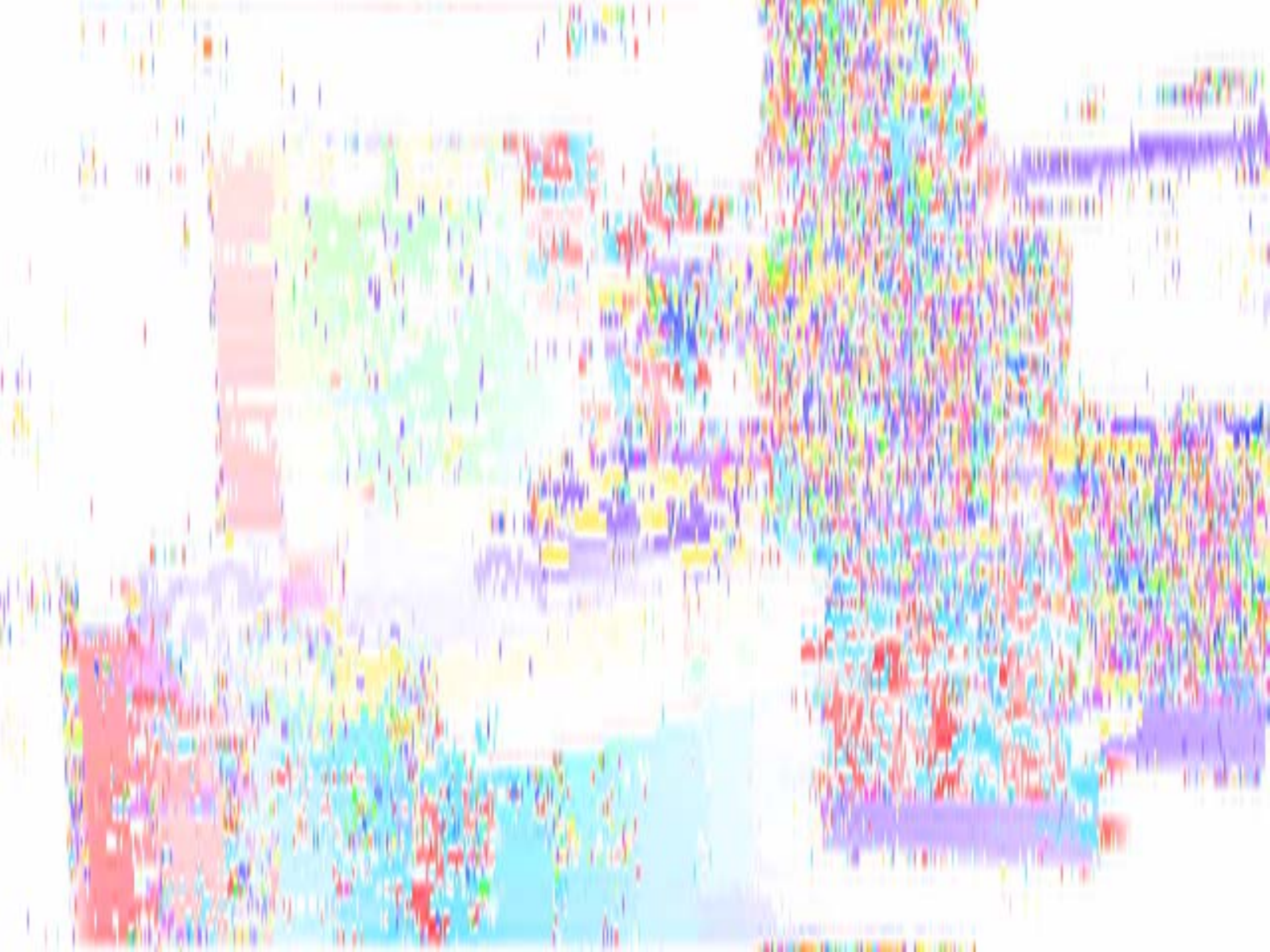}
\includegraphics[width=0.15\columnwidth]{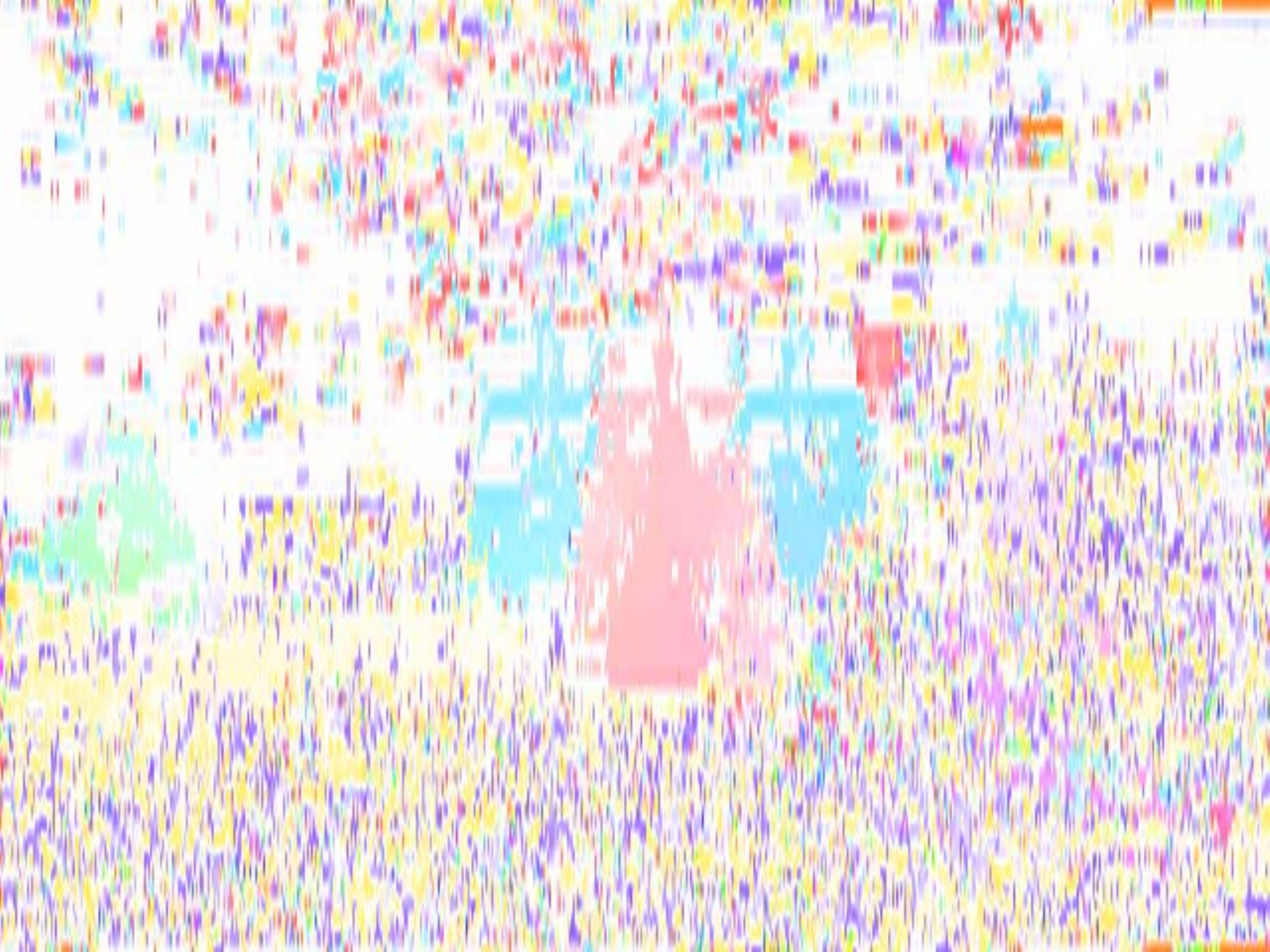}
\includegraphics[width=0.15\columnwidth]{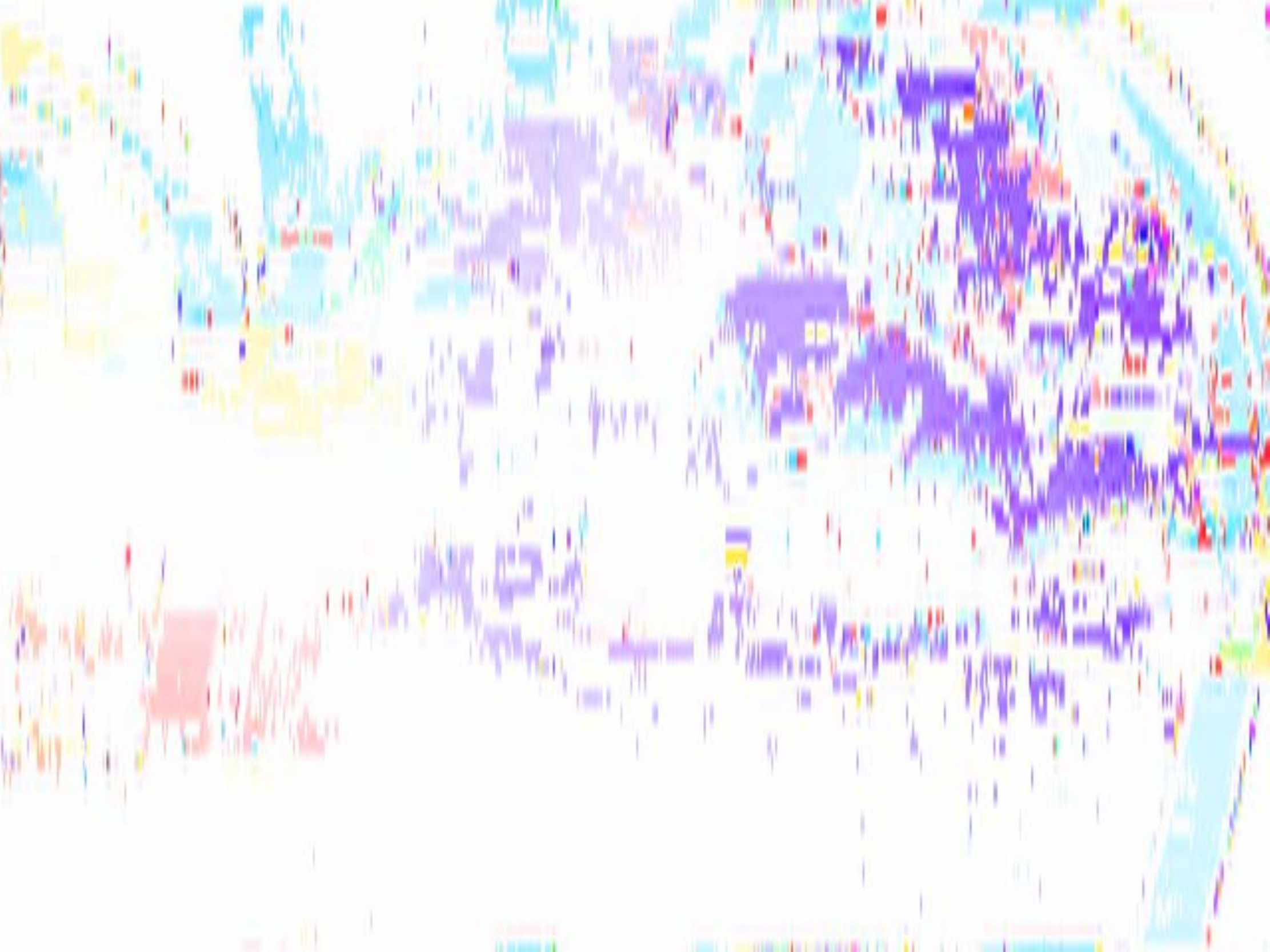}
\includegraphics[width=0.15\columnwidth]{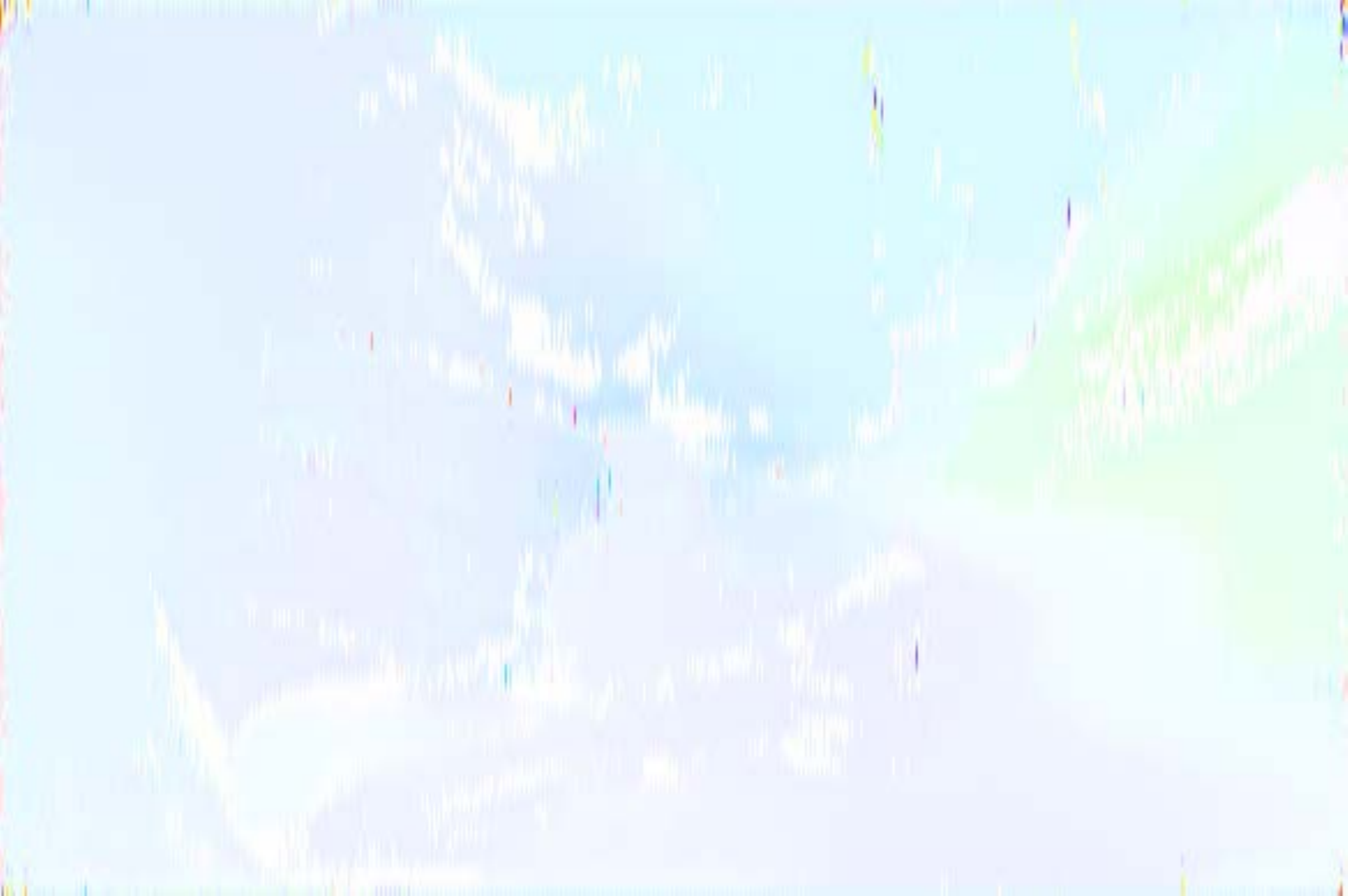}
\includegraphics[width=0.15\columnwidth]{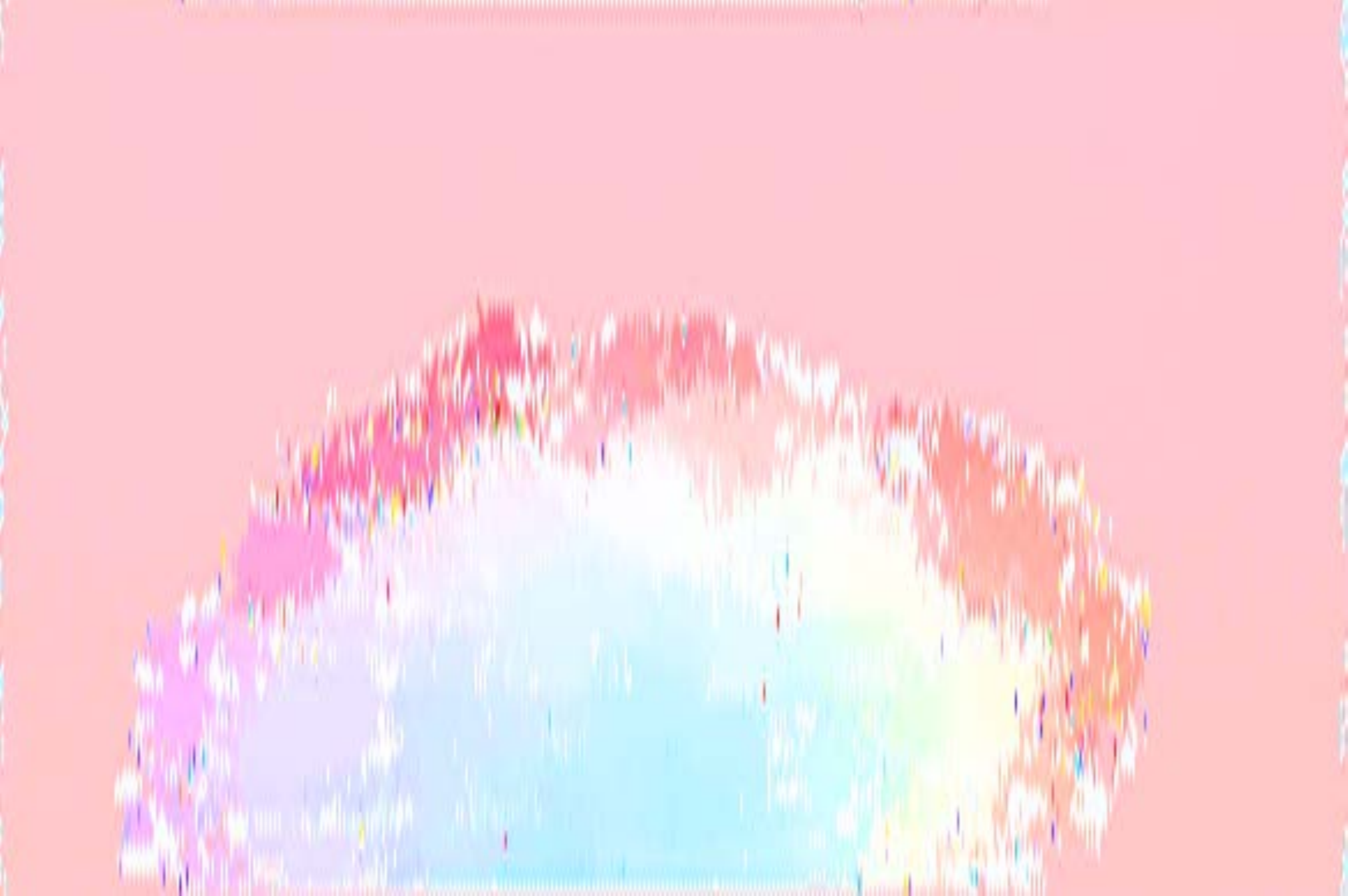}
\includegraphics[width=0.15\columnwidth]{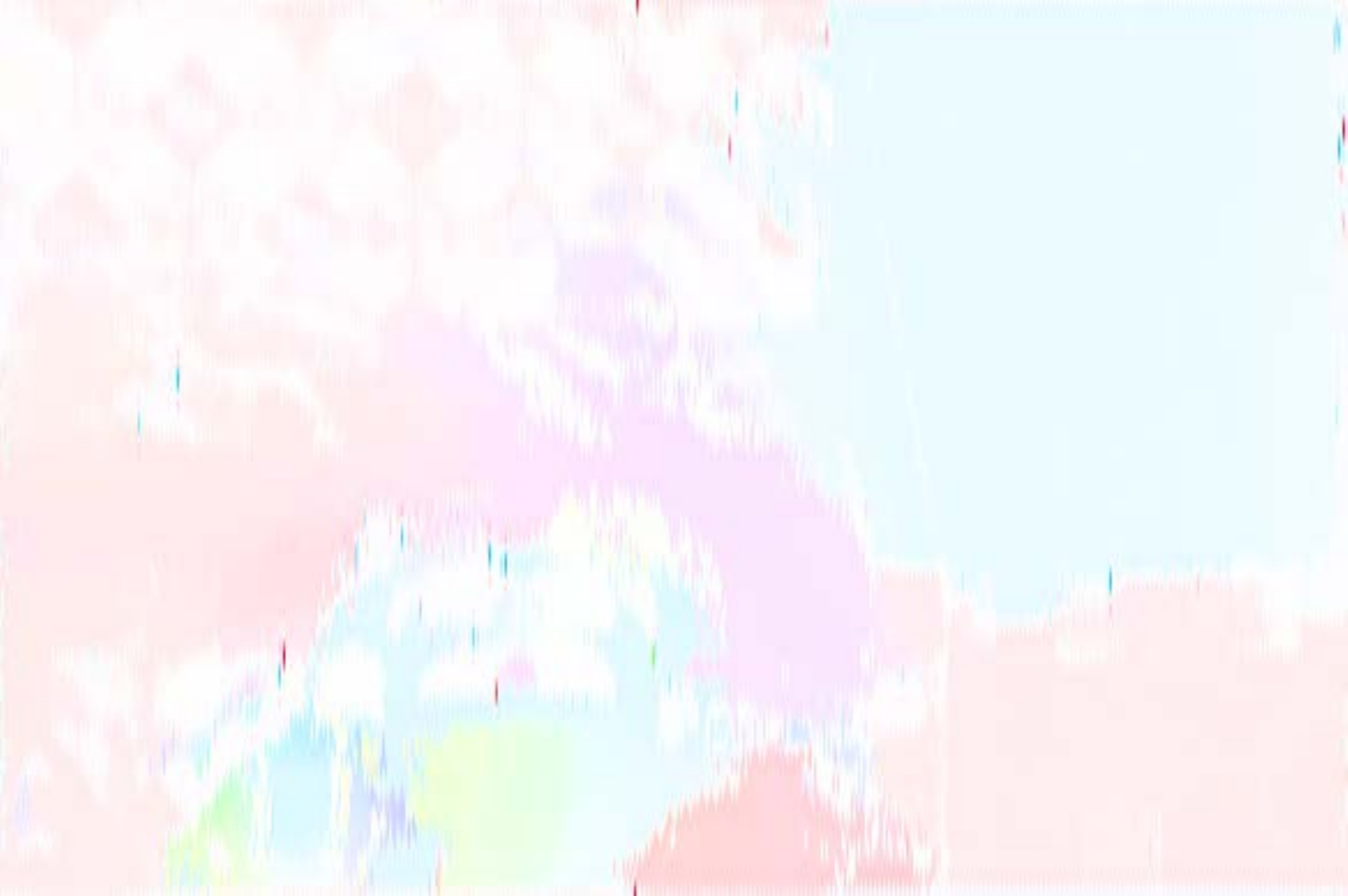}\\ \vspace{1 mm}
\includegraphics[width=0.15\columnwidth]{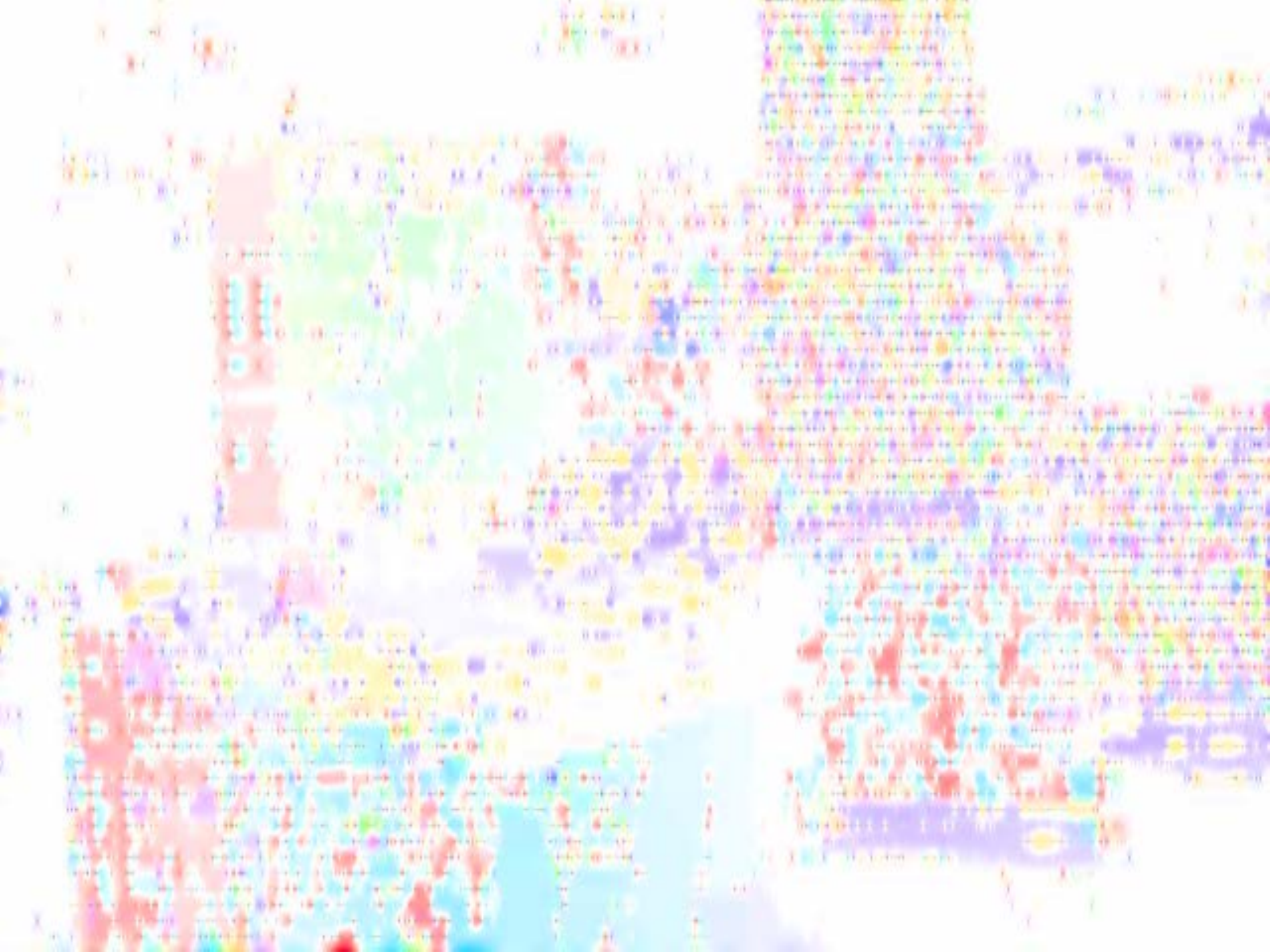}
\includegraphics[width=0.15\columnwidth]{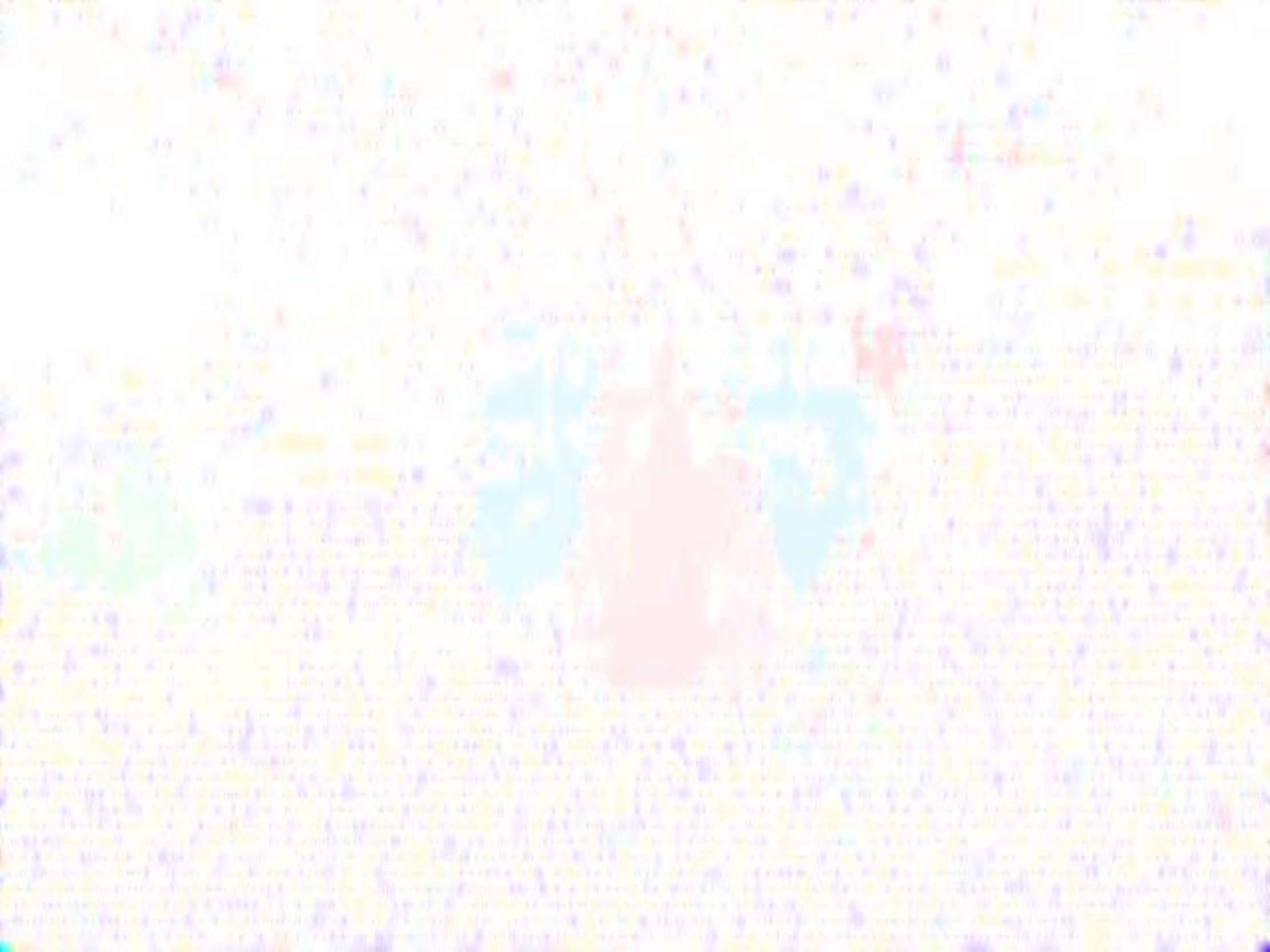}
\includegraphics[width=0.15\columnwidth]{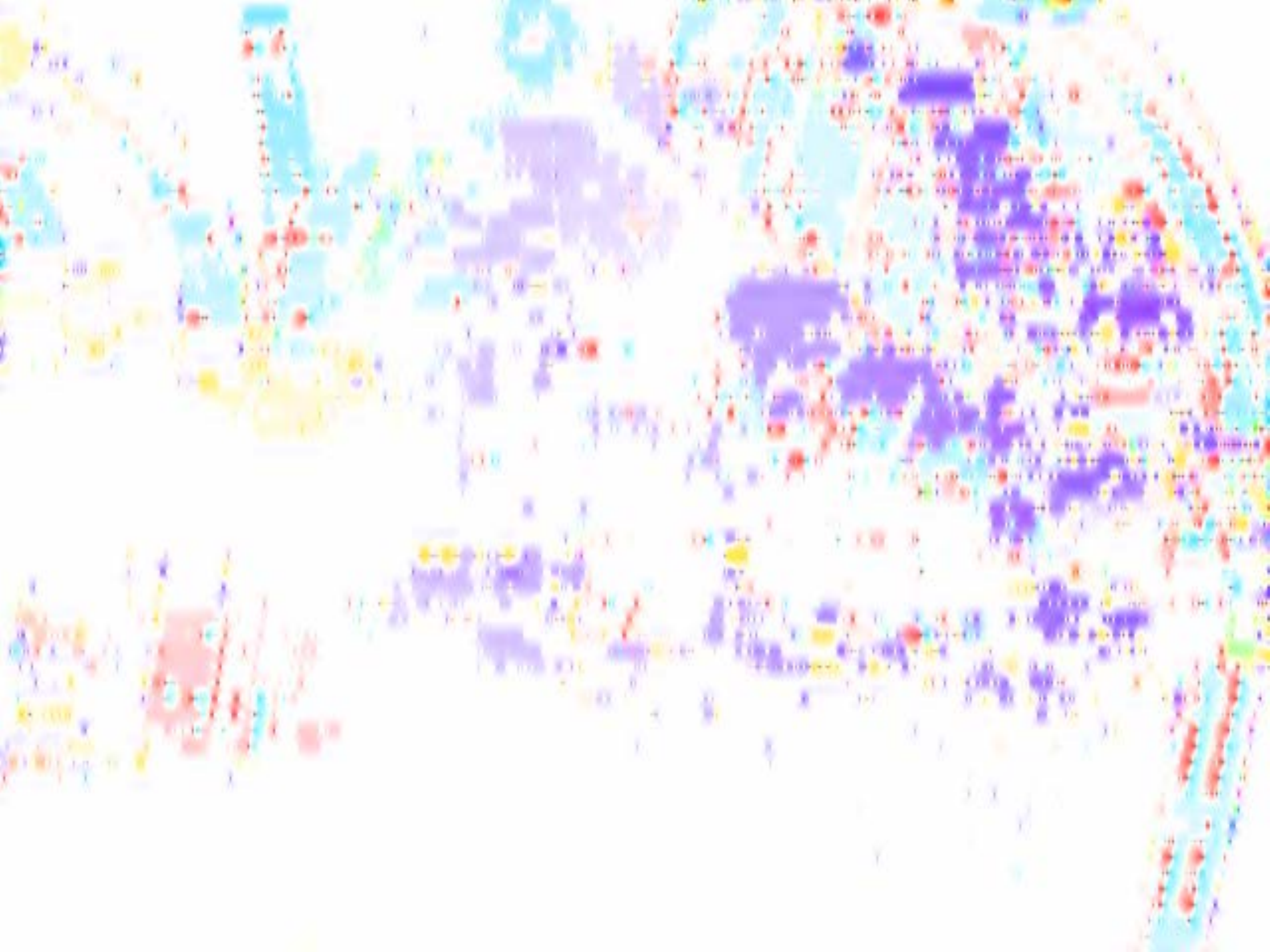}
\includegraphics[width=0.15\columnwidth]{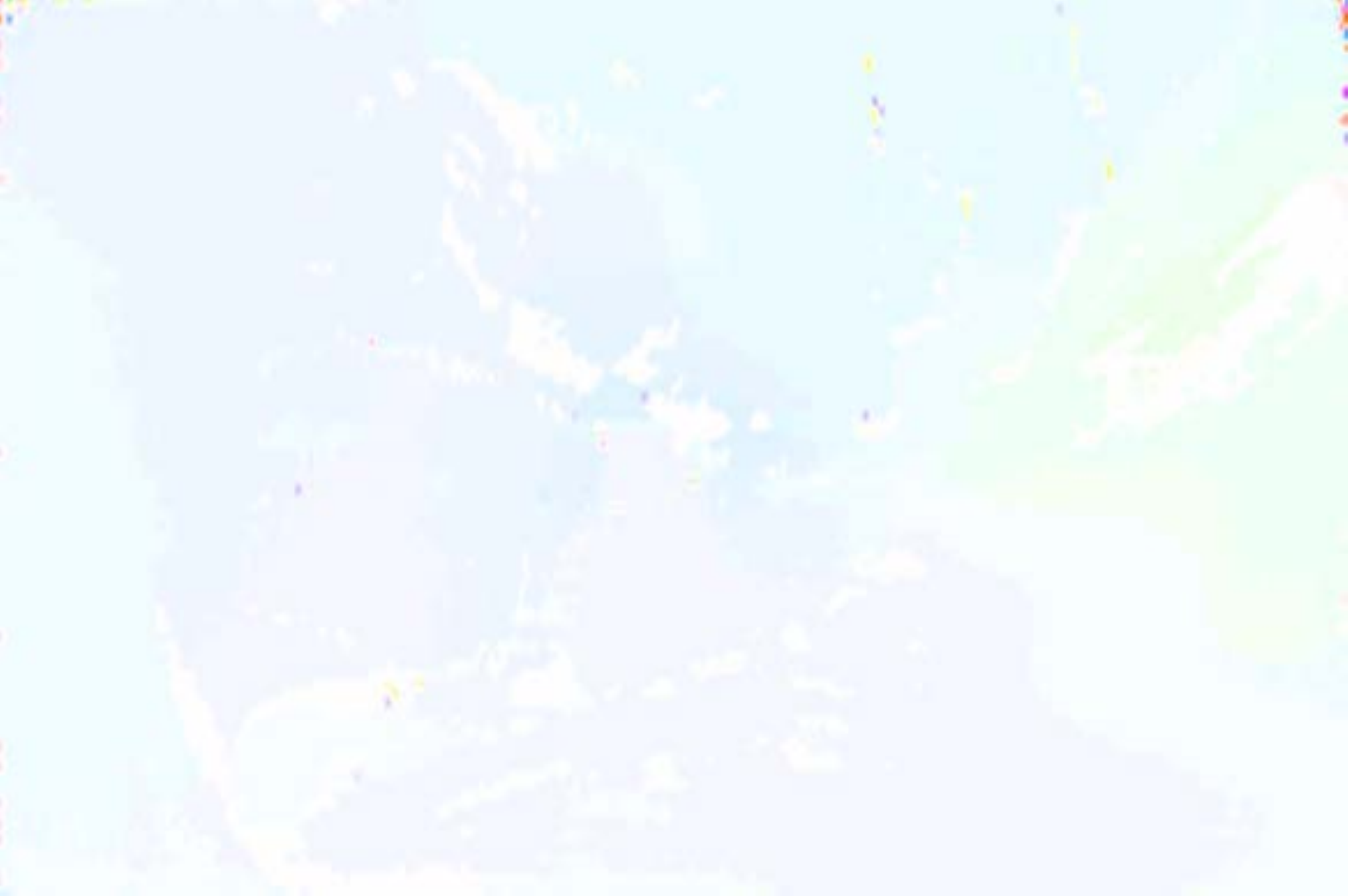}
\includegraphics[width=0.15\columnwidth]{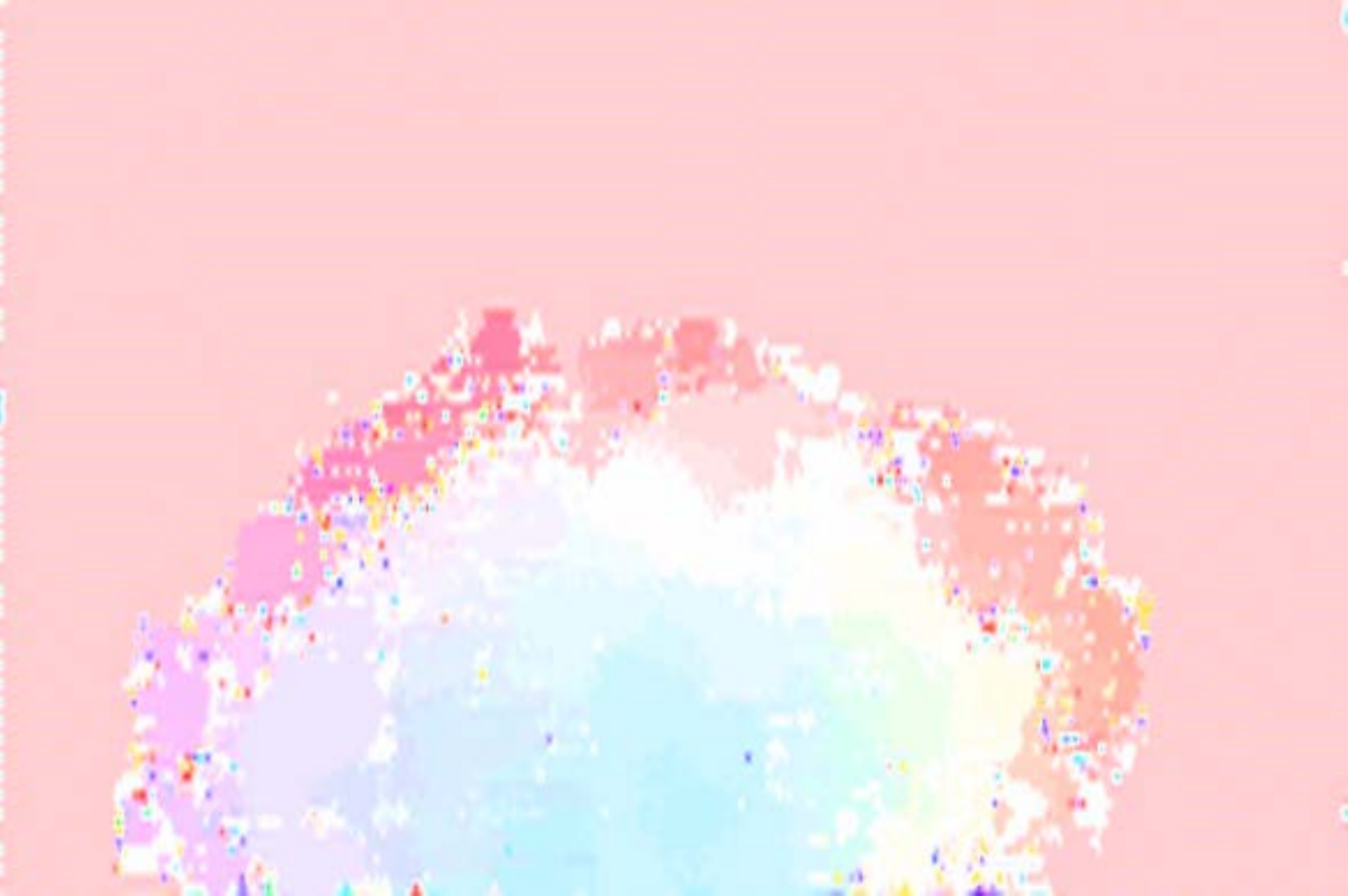}
\includegraphics[width=0.15\columnwidth]{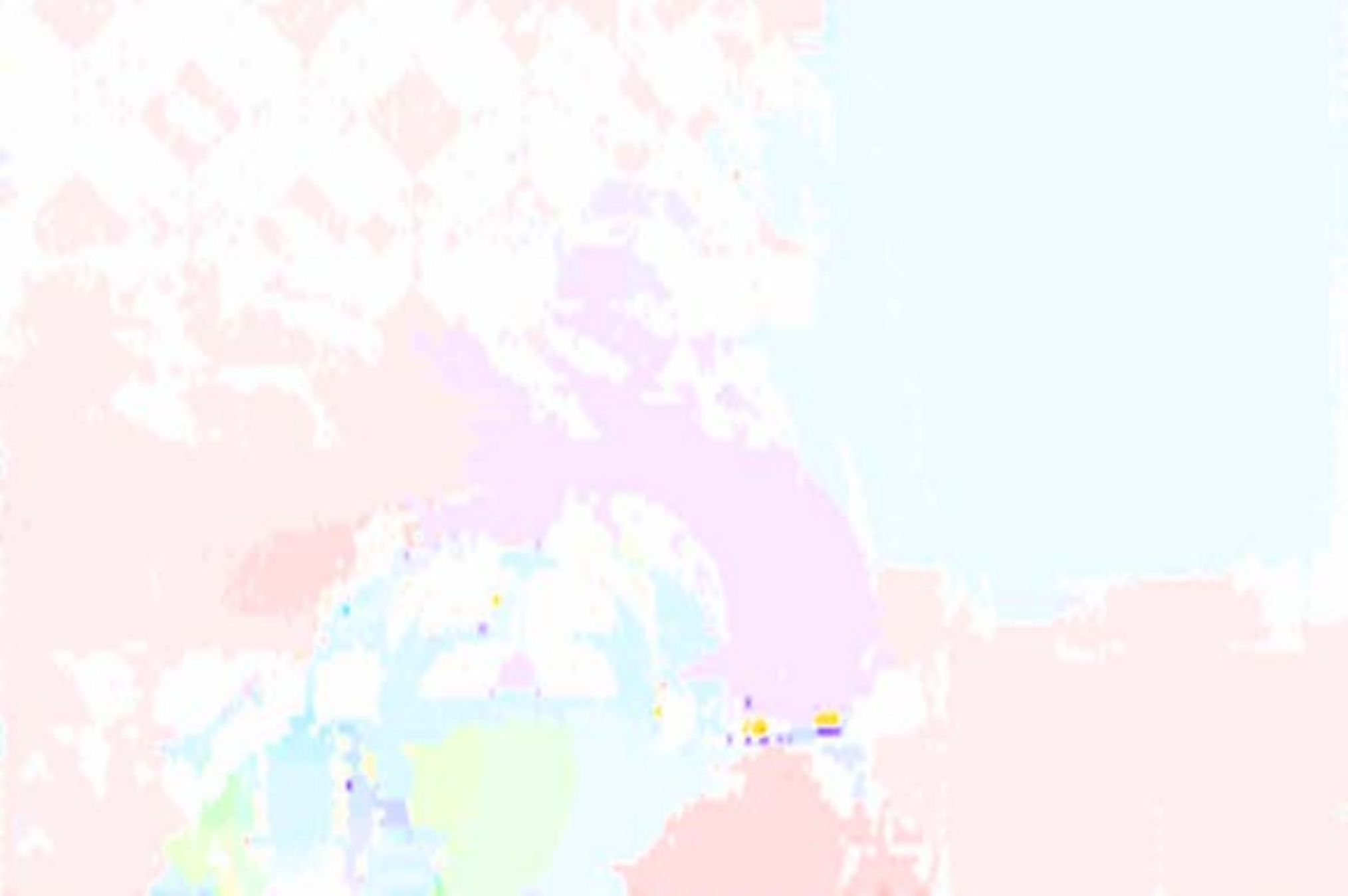}\\ \vspace{1 mm}
\includegraphics[width=0.15\columnwidth]{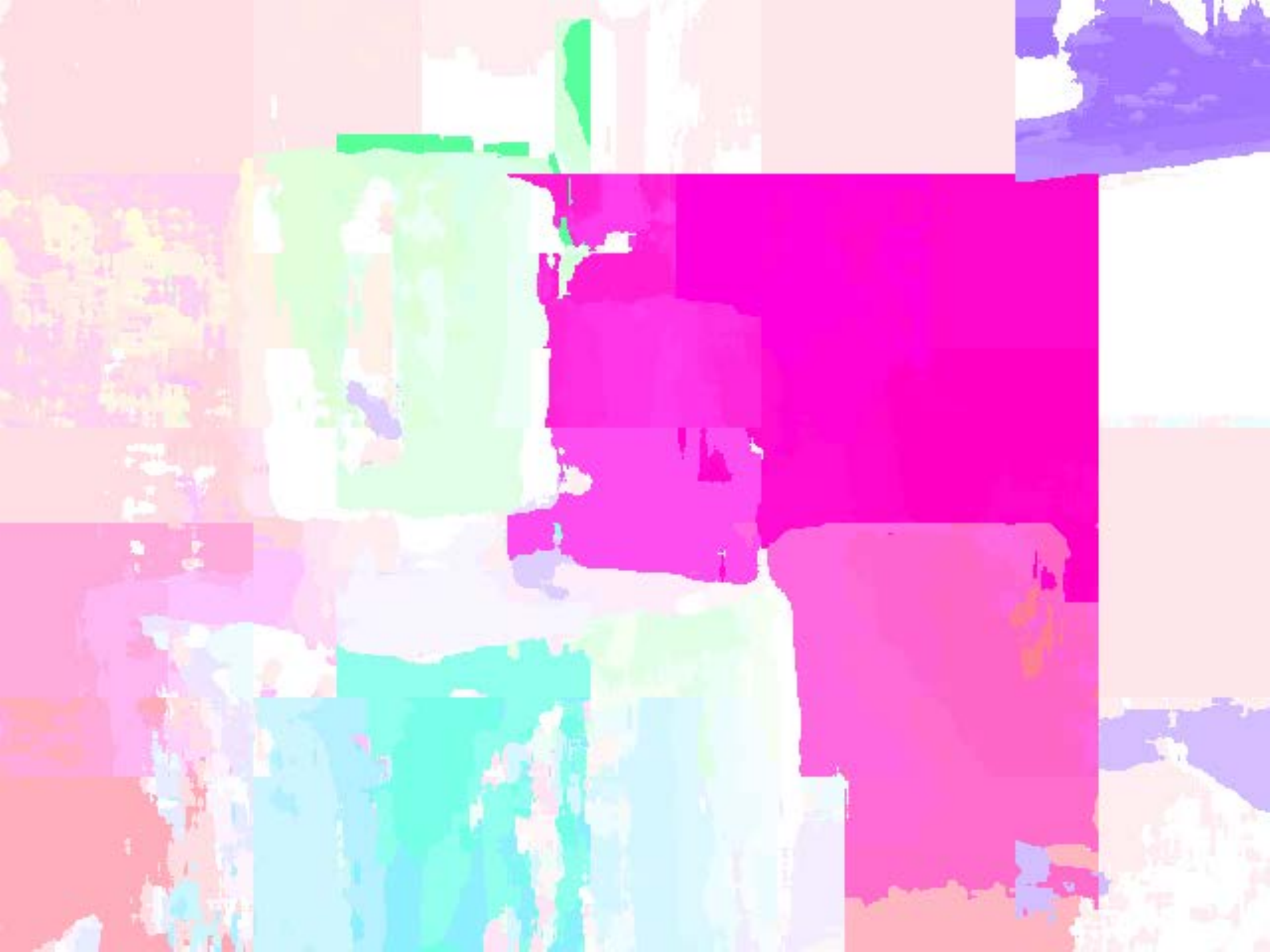}
\includegraphics[width=0.15\columnwidth]{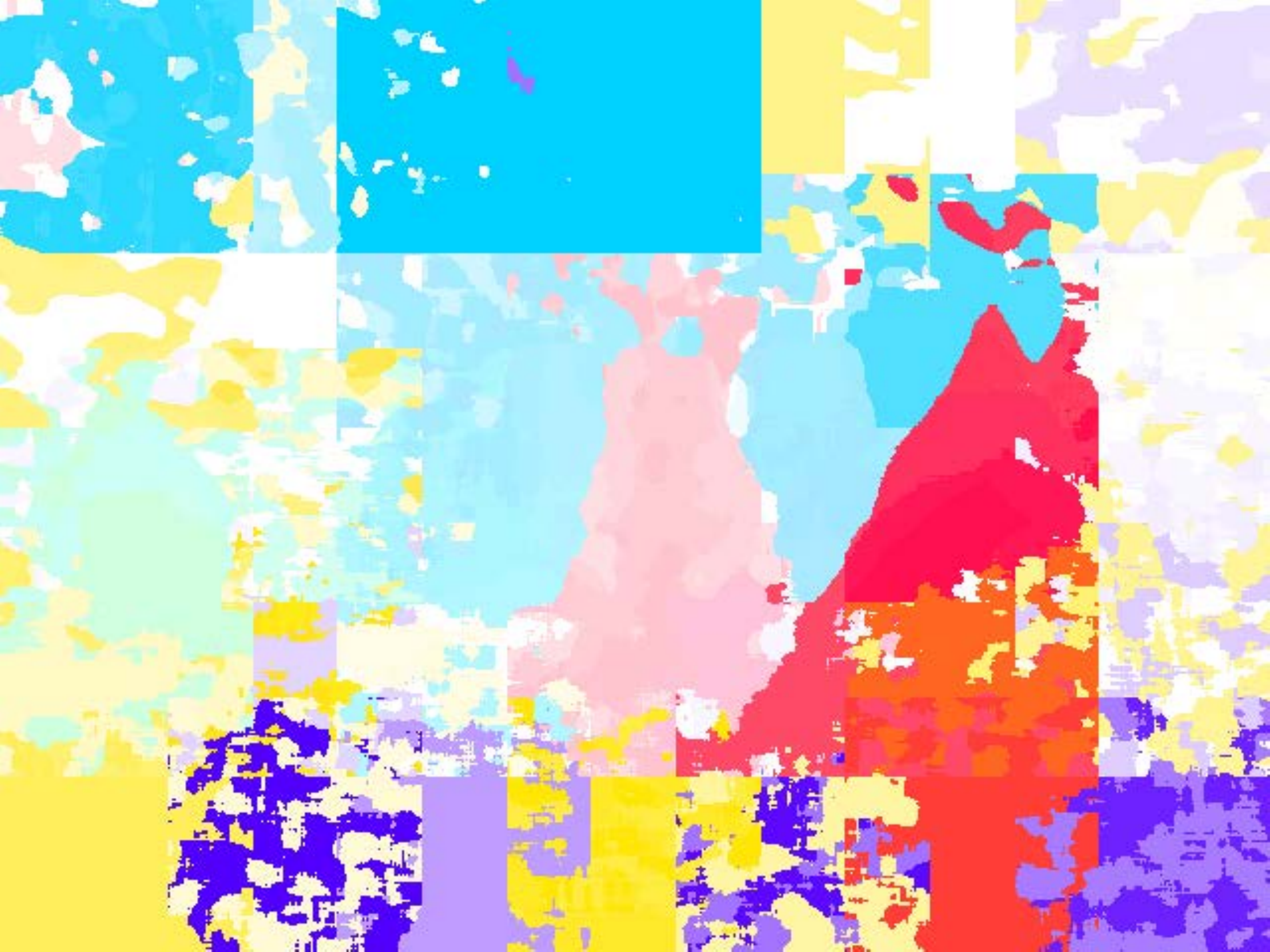}
\includegraphics[width=0.15\columnwidth]{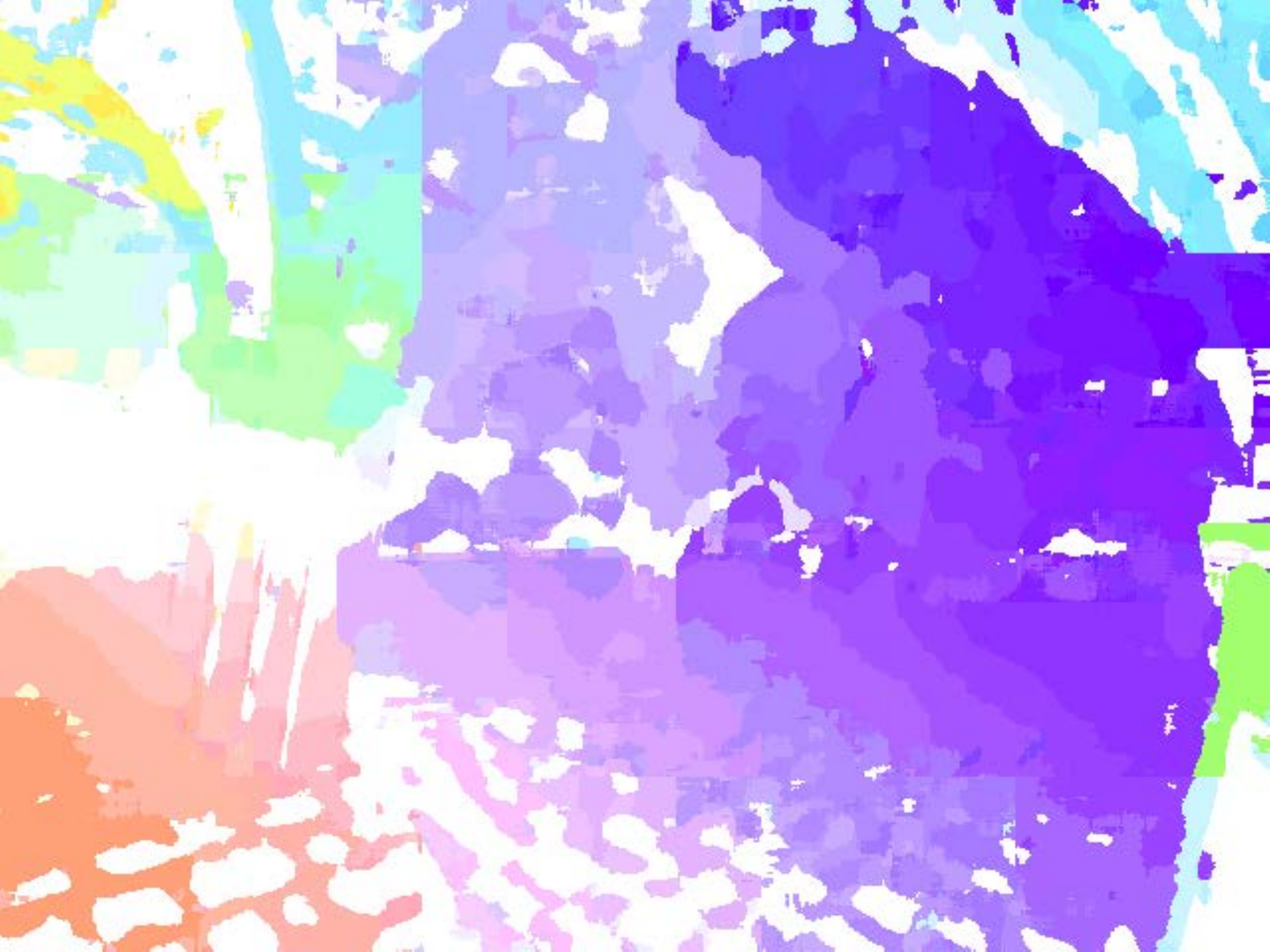}
\includegraphics[width=0.15\columnwidth]{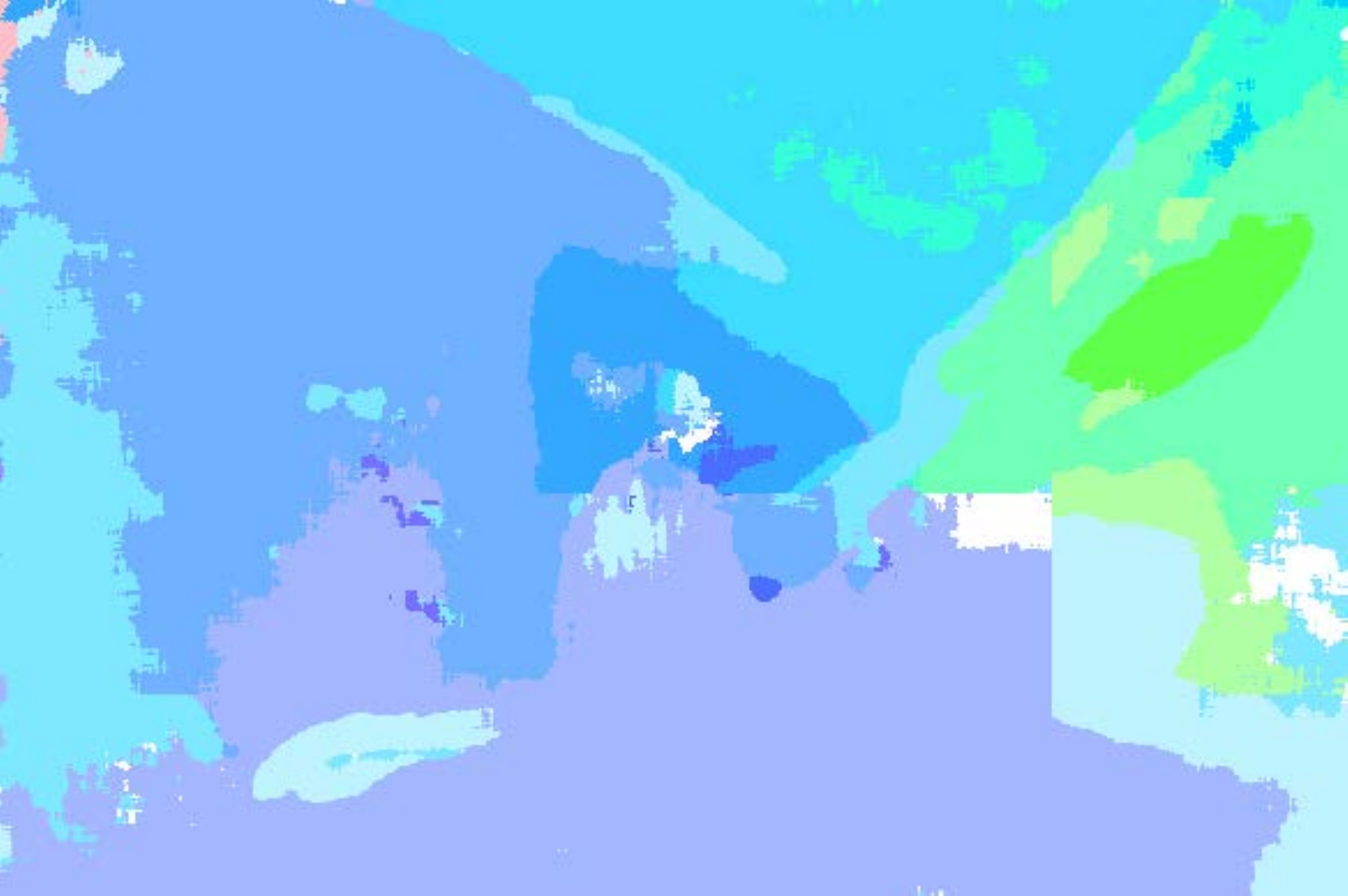}
\includegraphics[width=0.15\columnwidth]{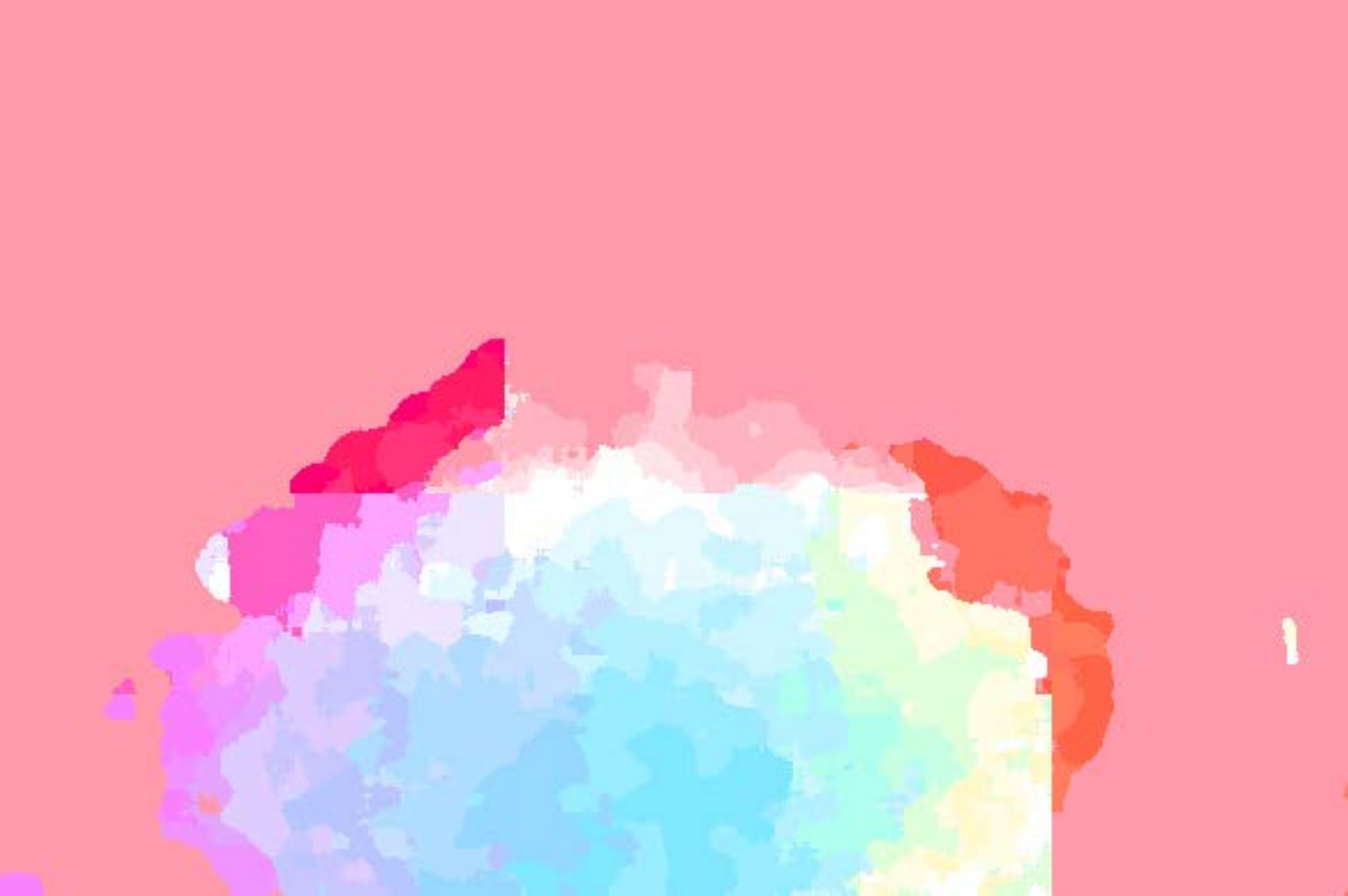}
\includegraphics[width=0.15\columnwidth]{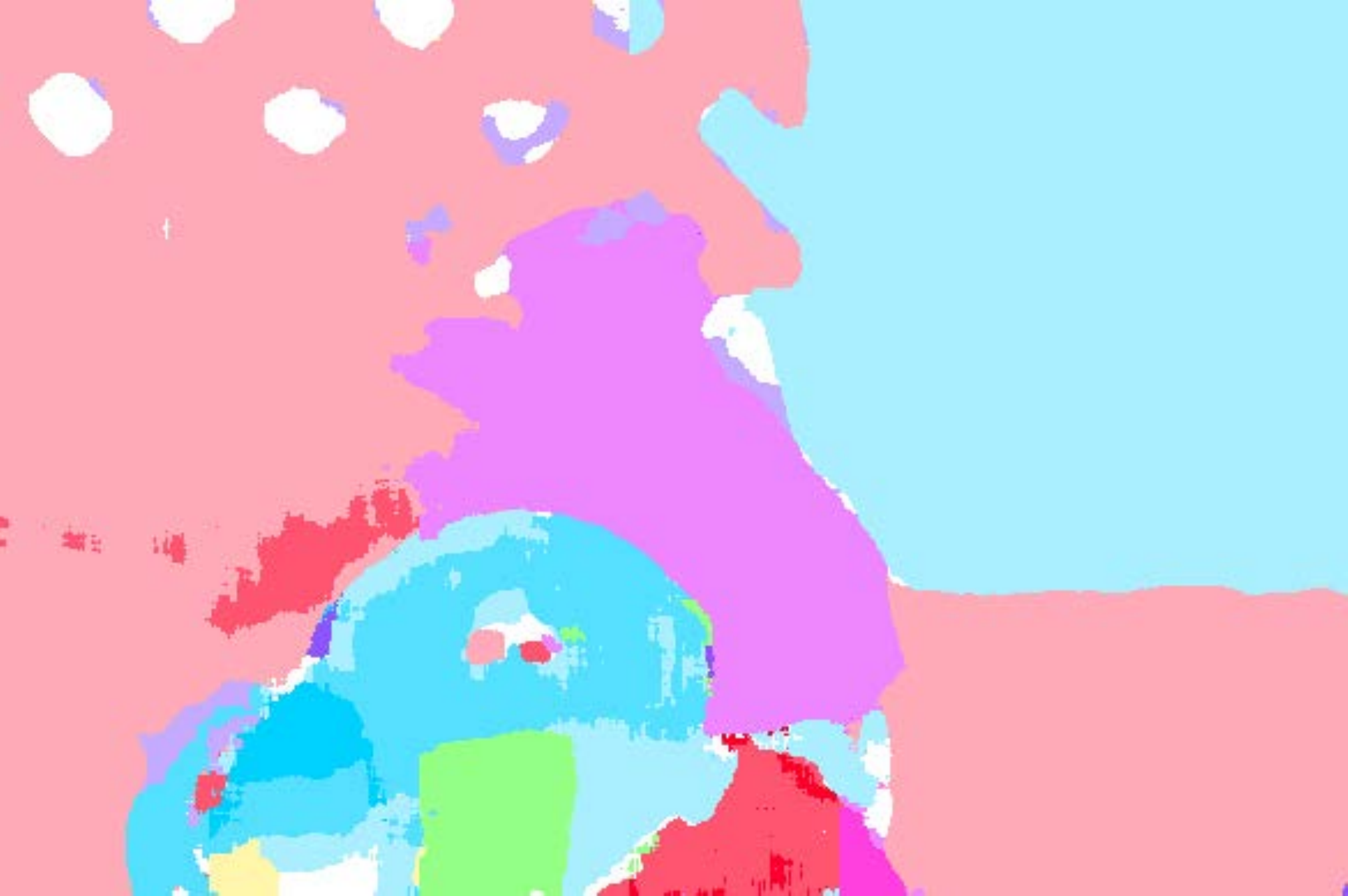}\\ \vspace{1 mm}
\includegraphics[width=0.15\columnwidth]{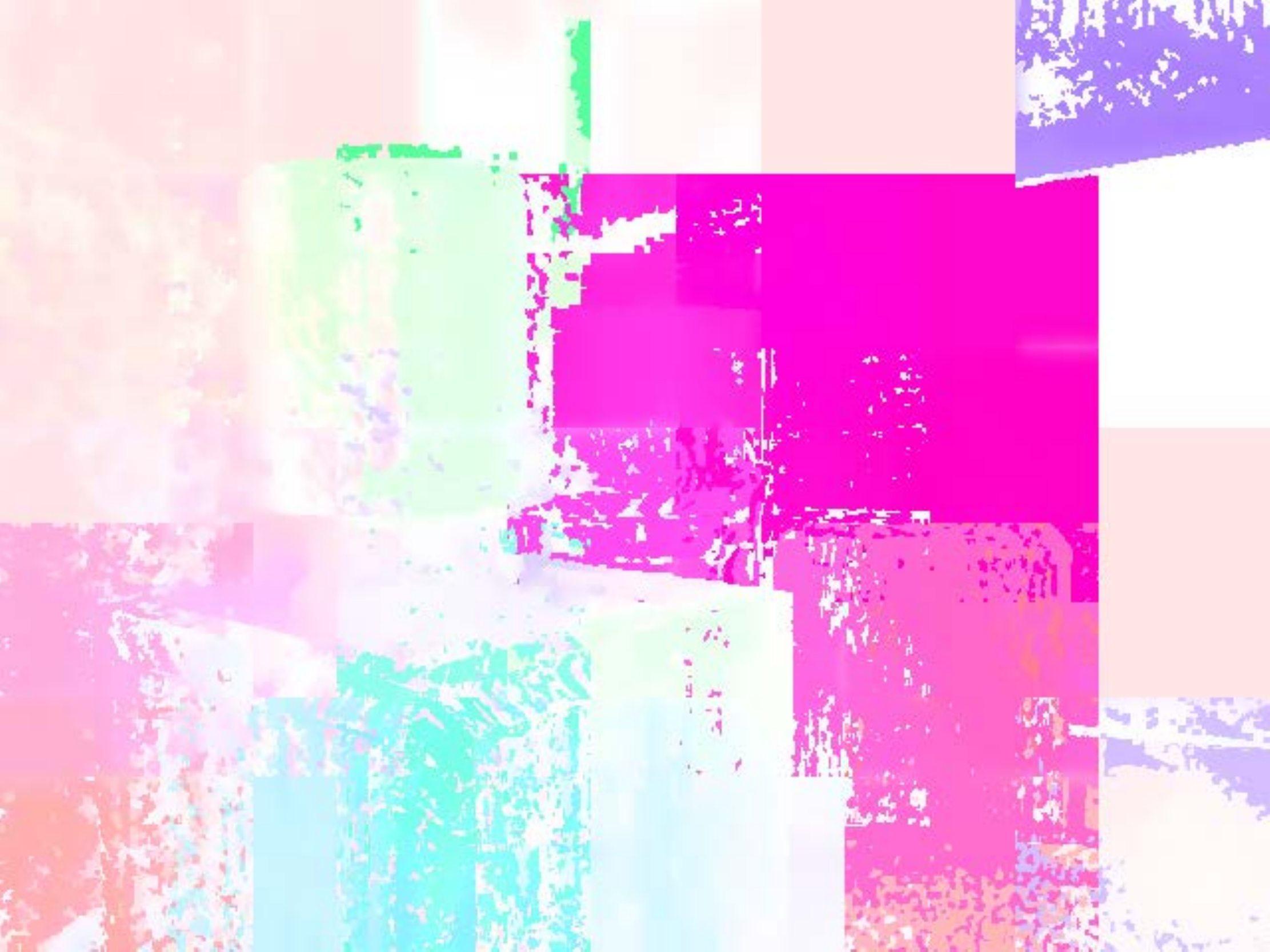}
\includegraphics[width=0.15\columnwidth]{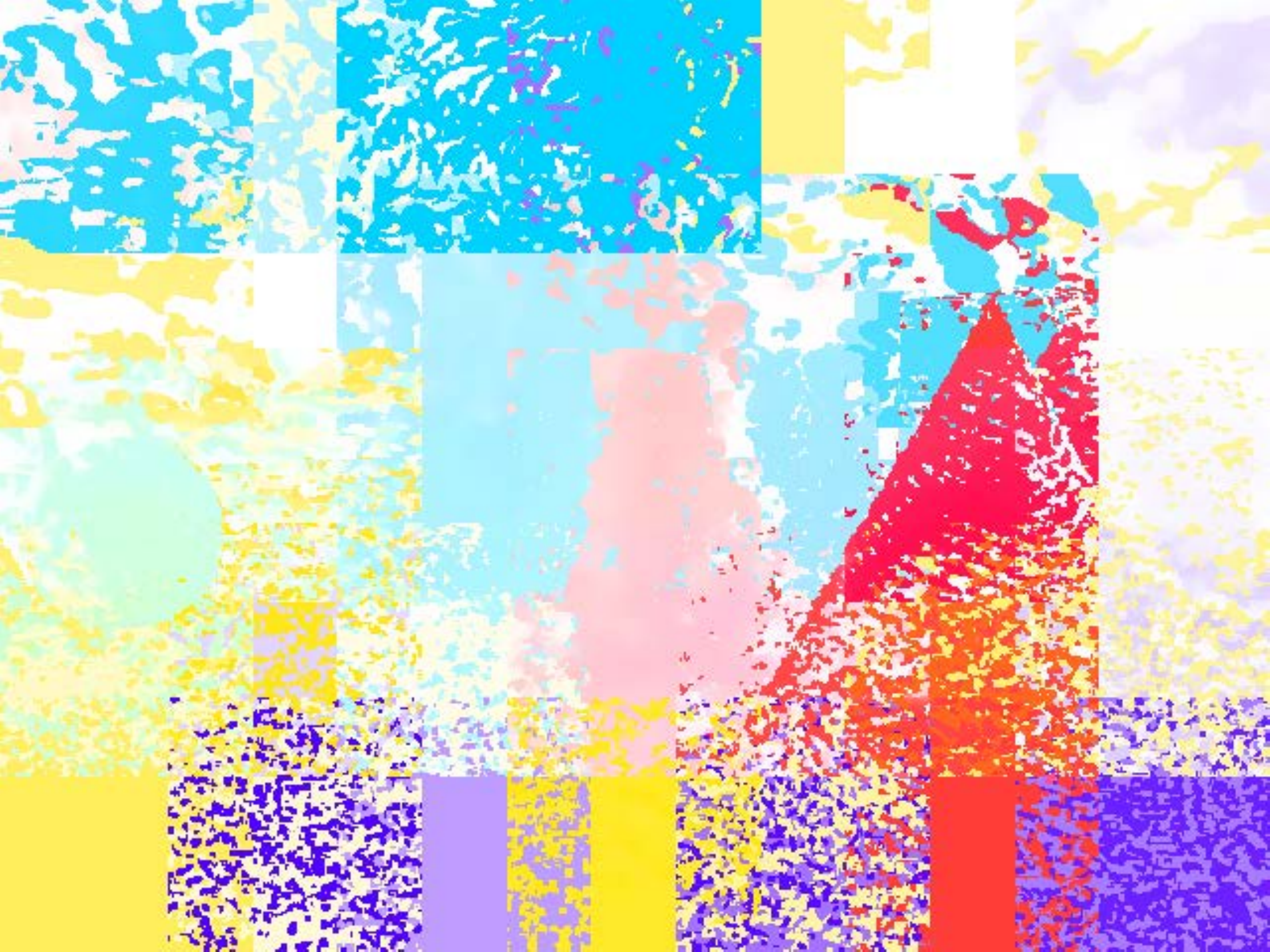}
\includegraphics[width=0.15\columnwidth]{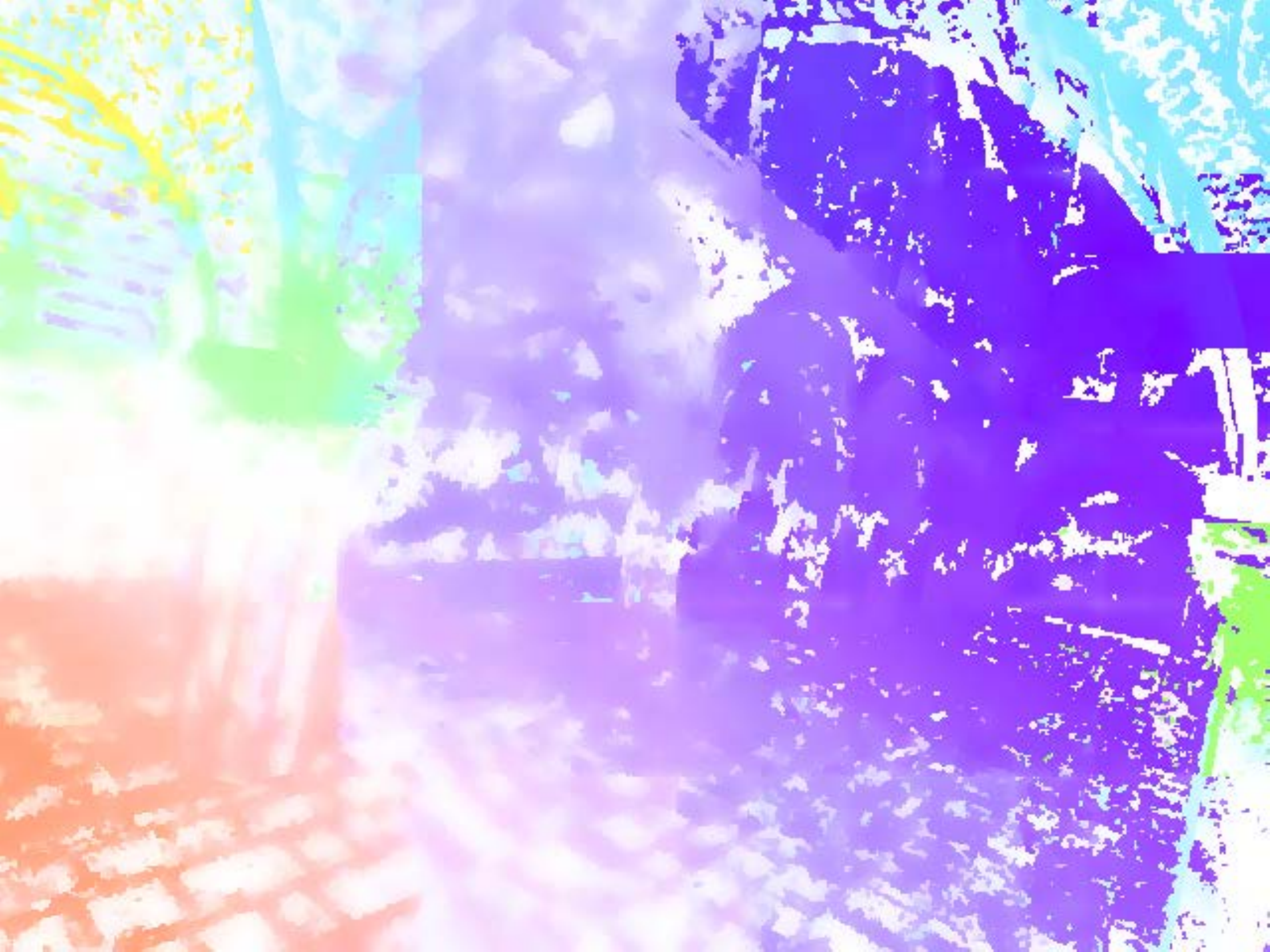}
\includegraphics[width=0.15\columnwidth]{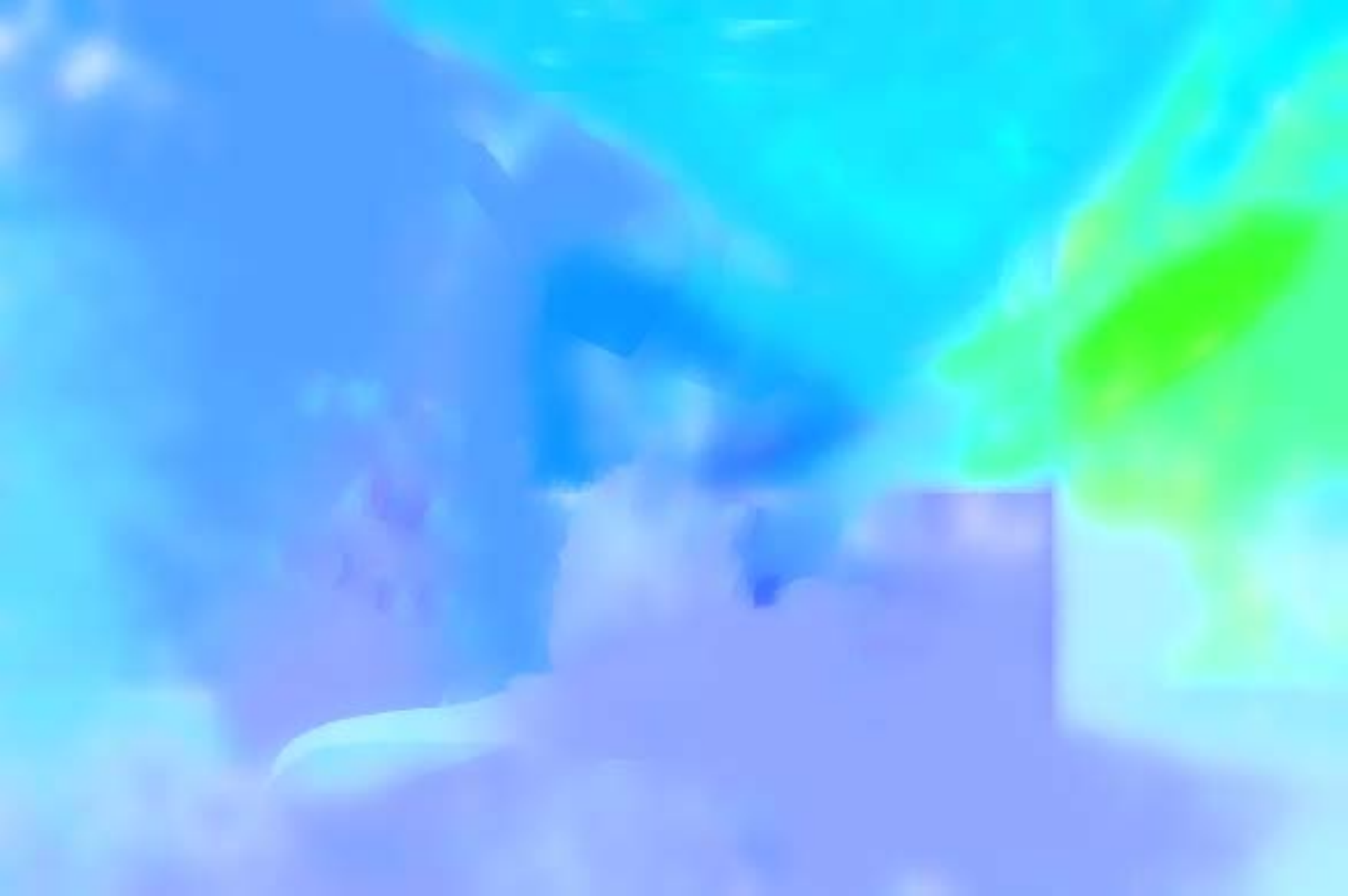}
\includegraphics[width=0.15\columnwidth]{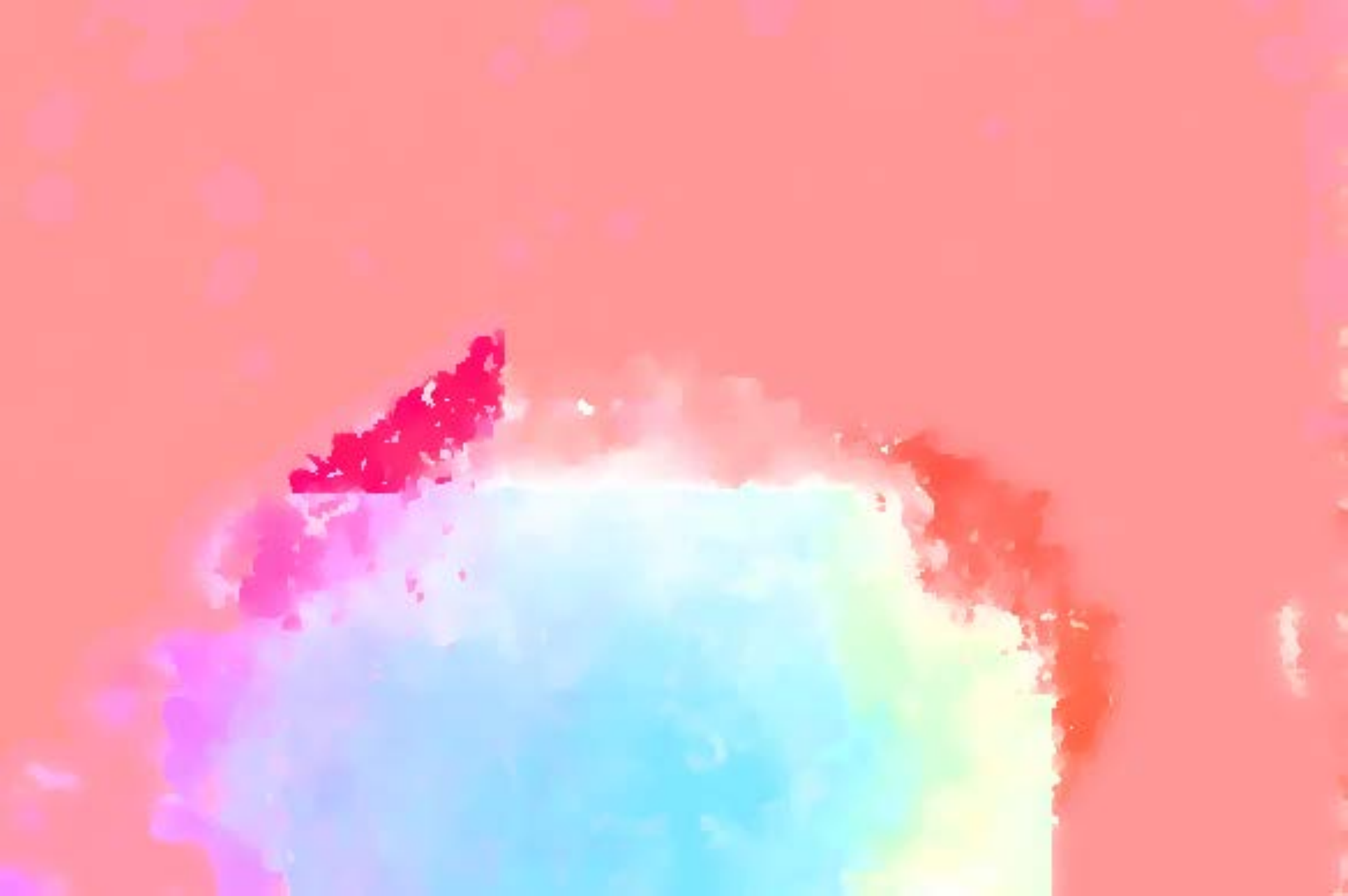}
\includegraphics[width=0.15\columnwidth]{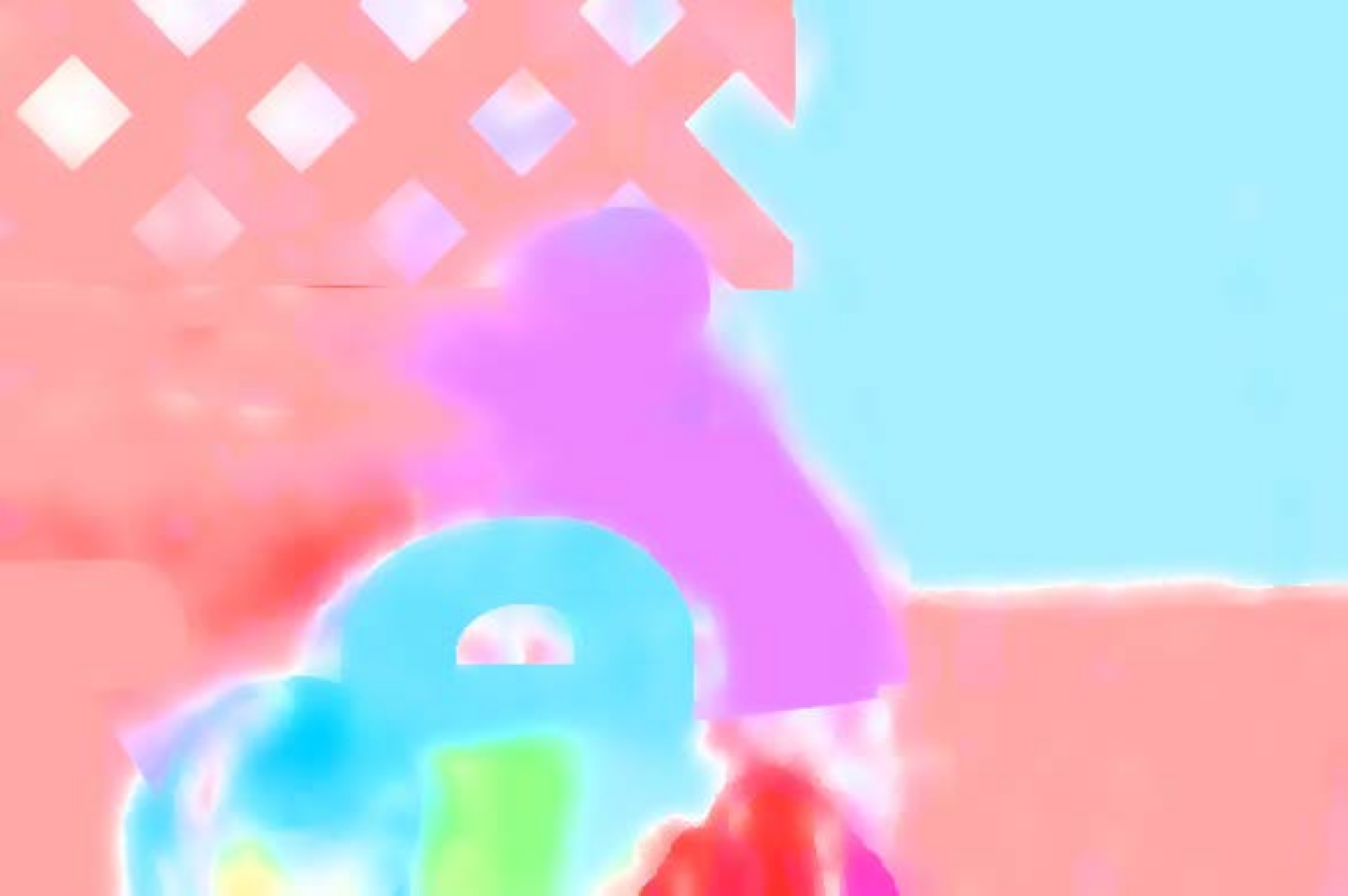}\\ \vspace{1 mm}
\includegraphics[width=0.15\columnwidth]{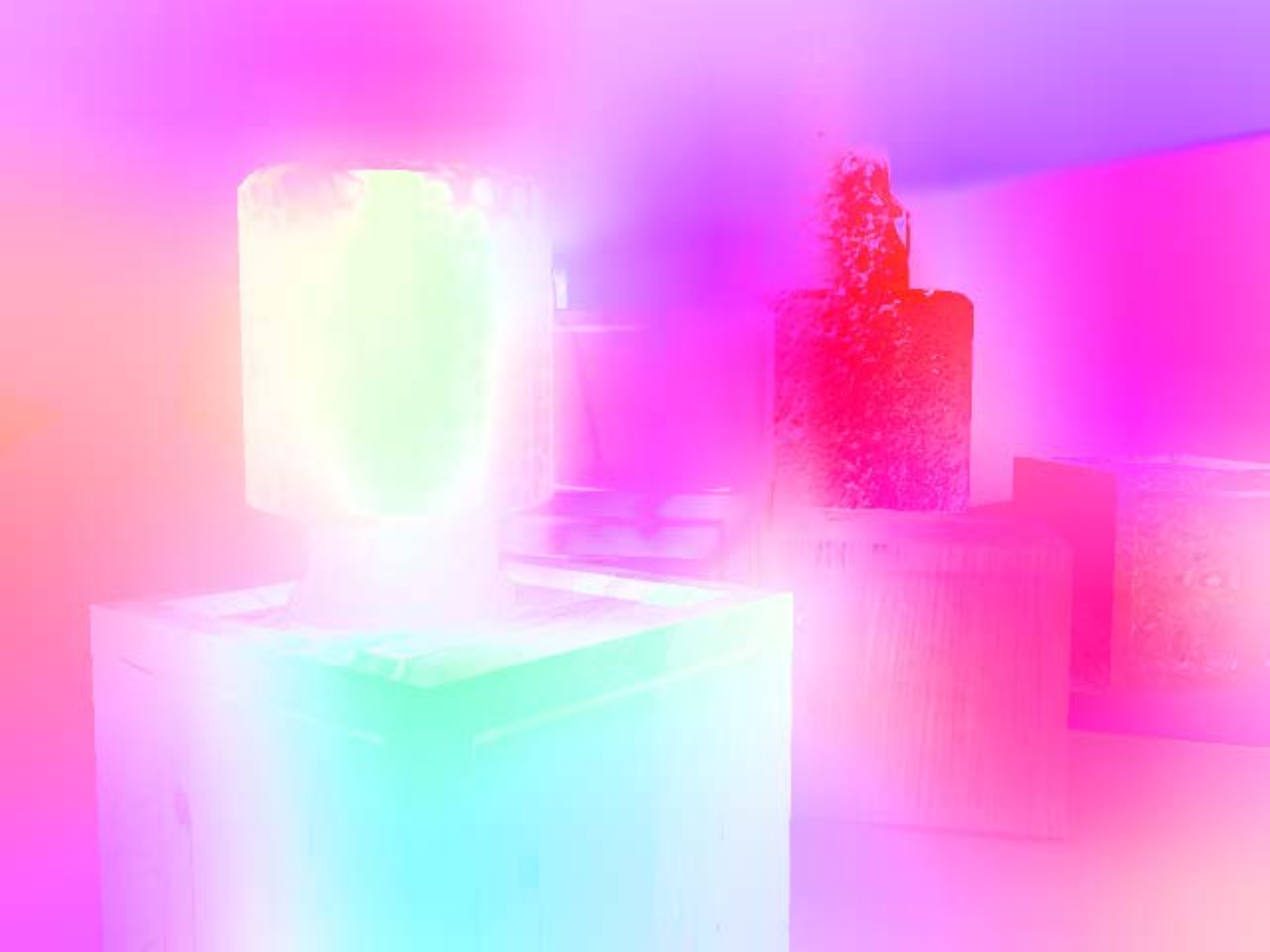}
\includegraphics[width=0.15\columnwidth]{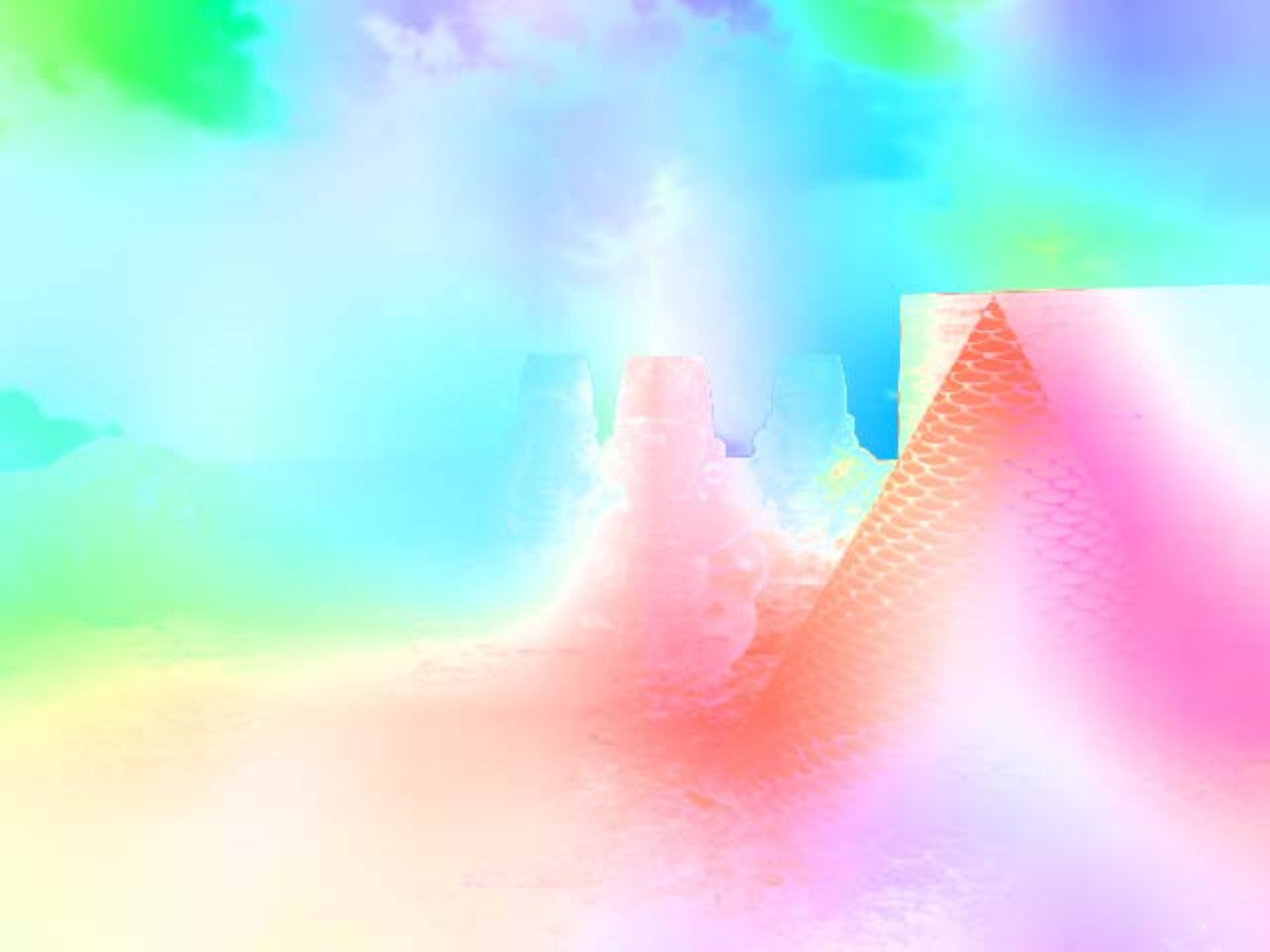}
\includegraphics[width=0.15\columnwidth]{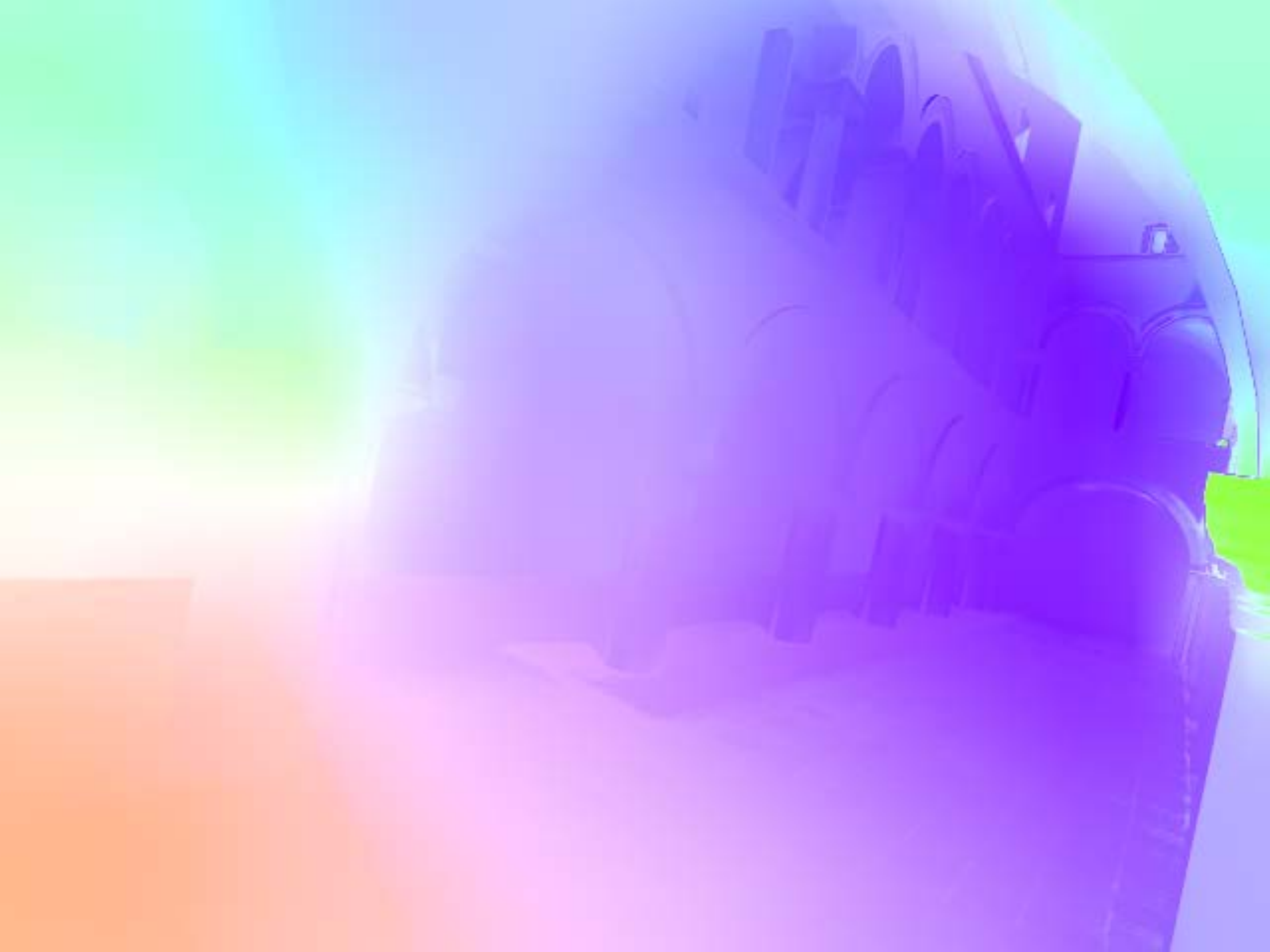}
\includegraphics[width=0.15\columnwidth]{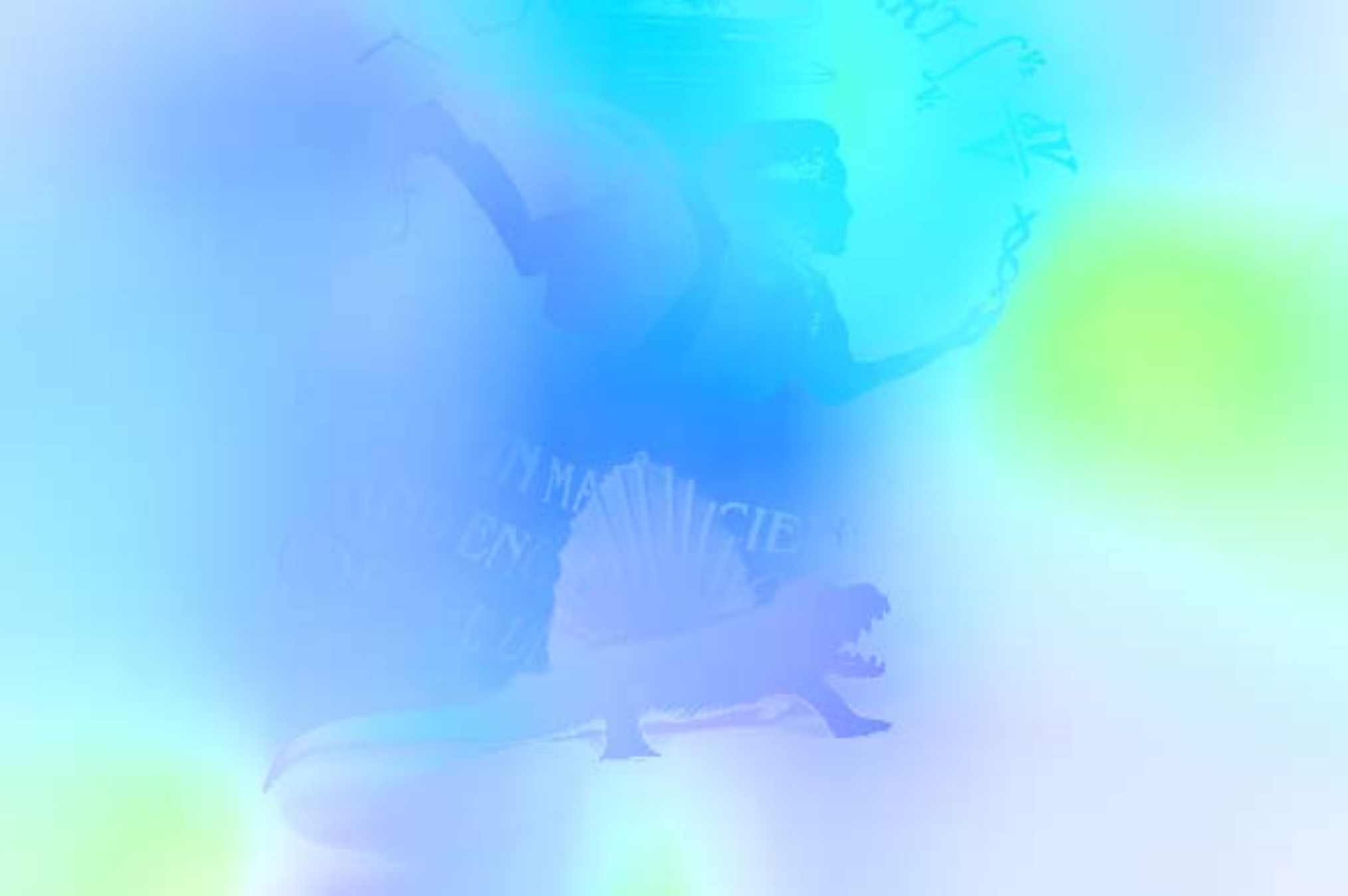}
\includegraphics[width=0.15\columnwidth]{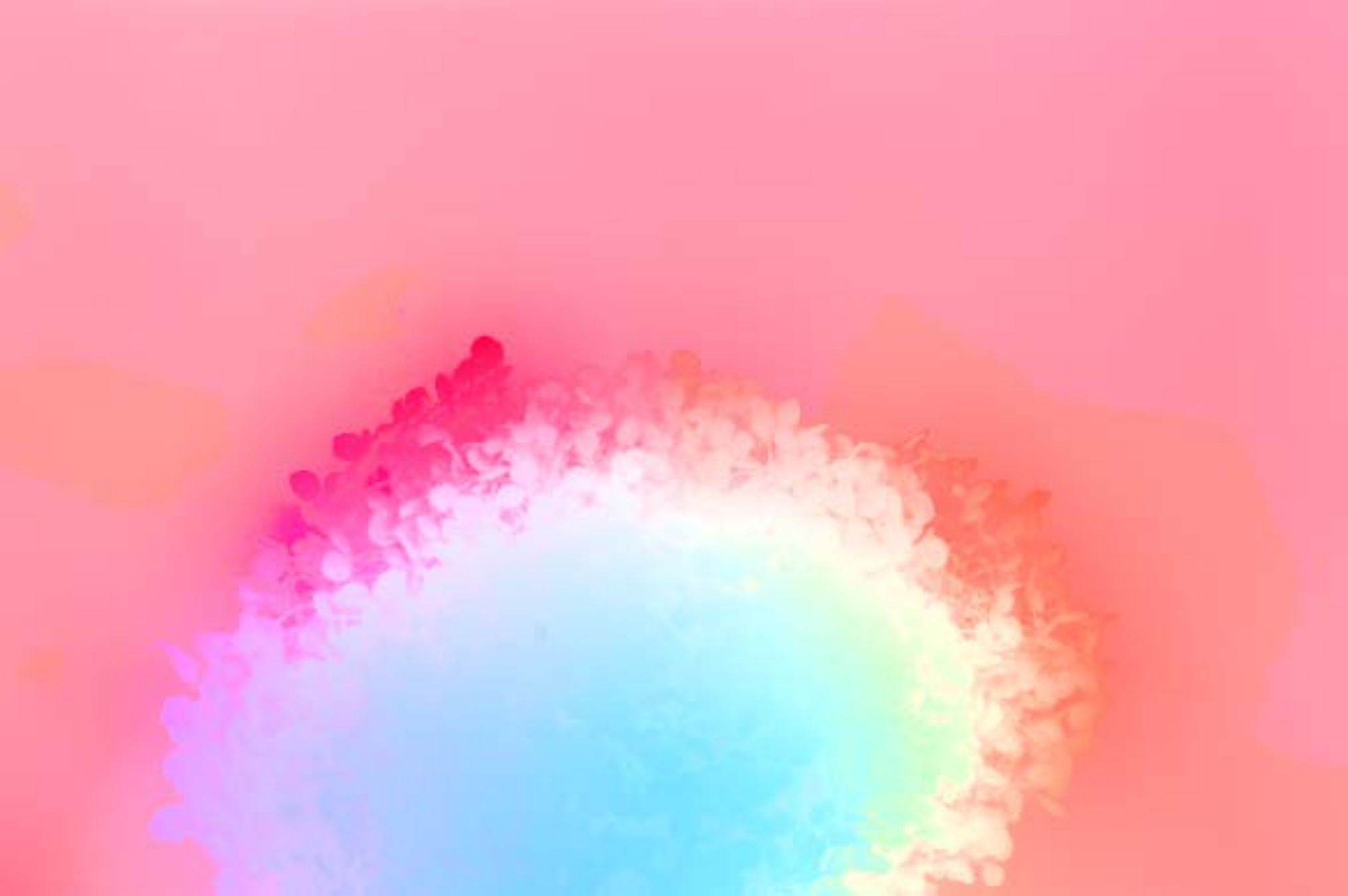}
\includegraphics[width=0.15\columnwidth]{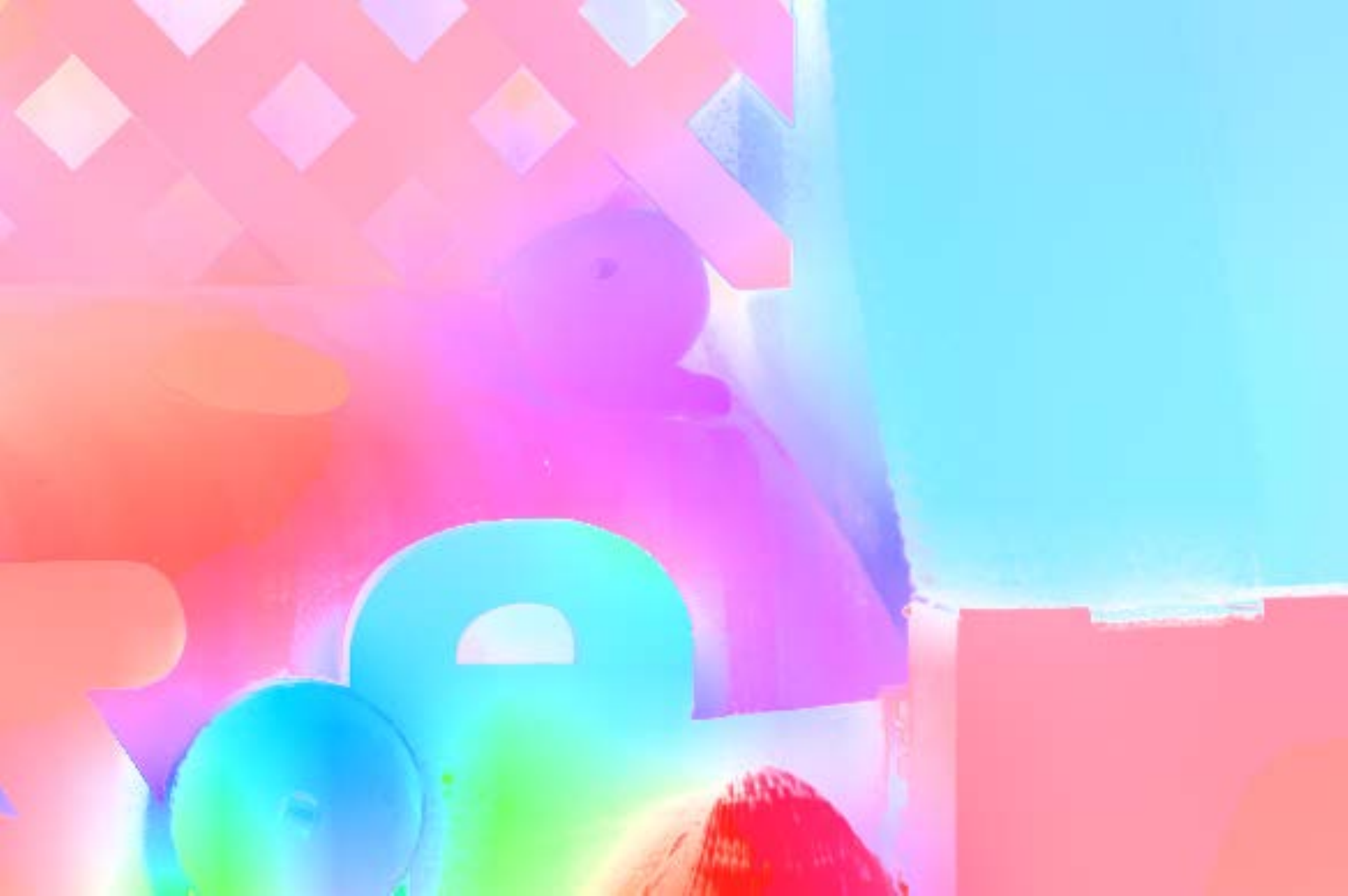}\\ \vspace{1 mm}
\includegraphics[width=0.15\columnwidth]{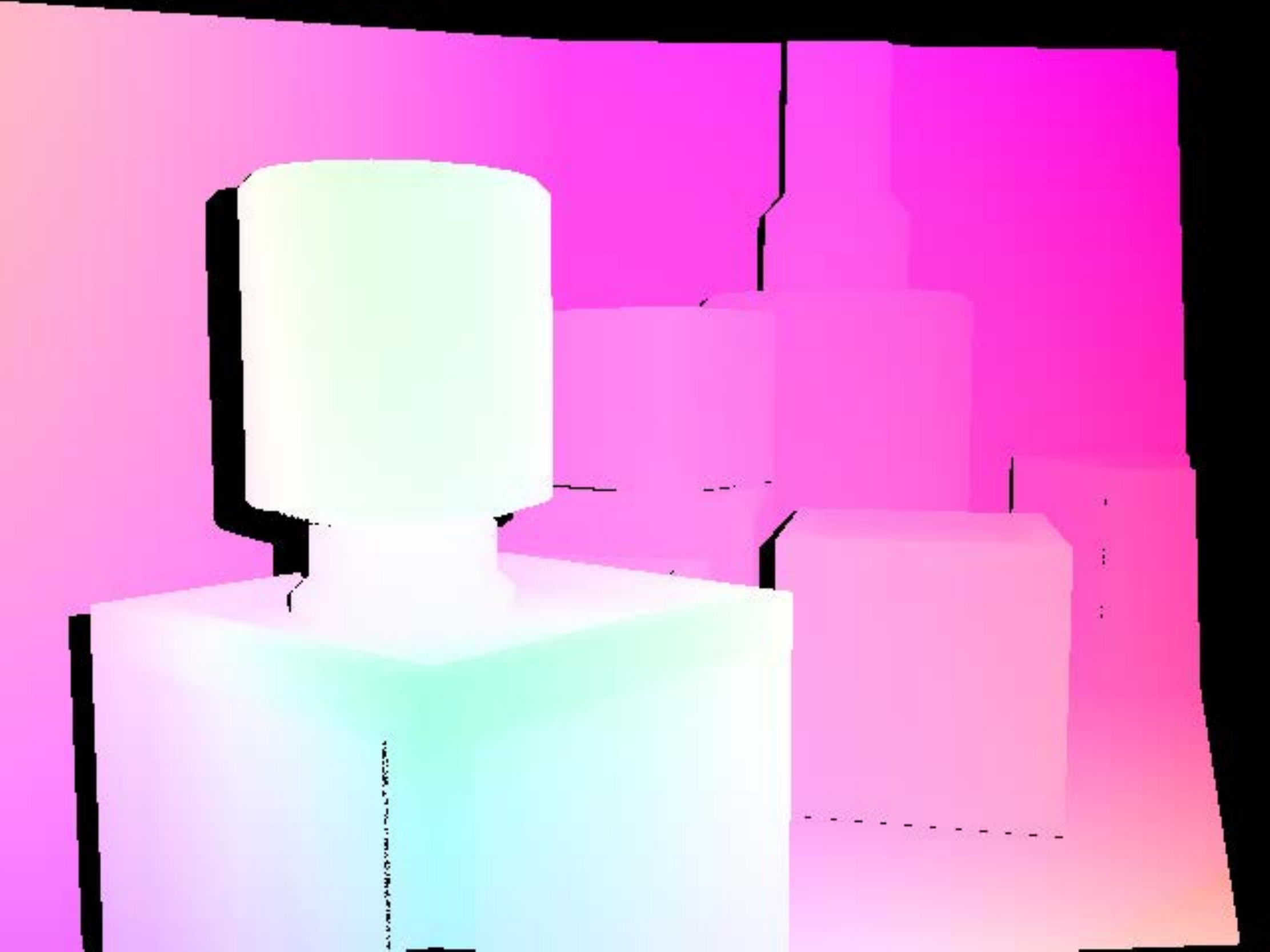}
\includegraphics[width=0.15\columnwidth]{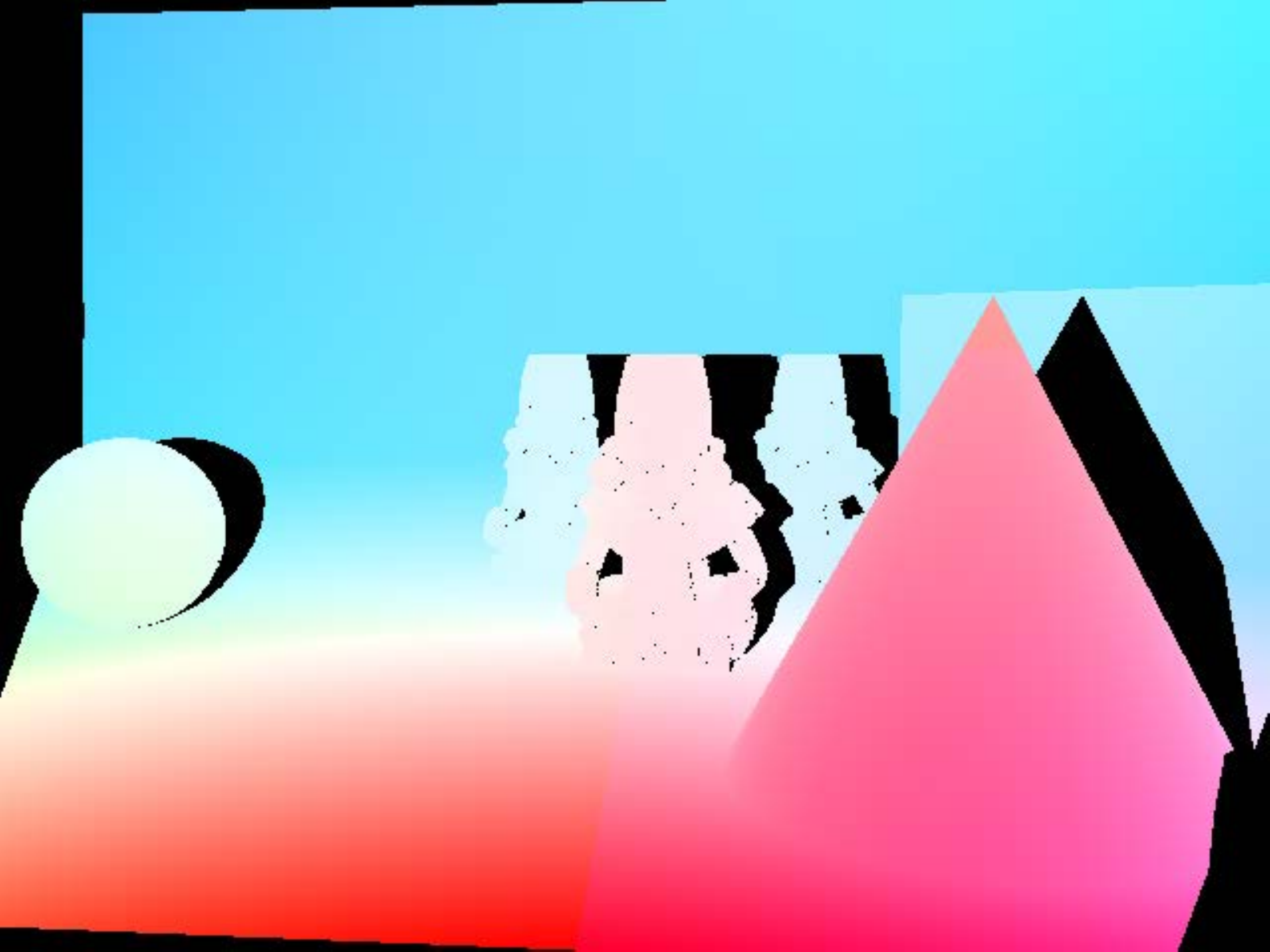}
\includegraphics[width=0.15\columnwidth]{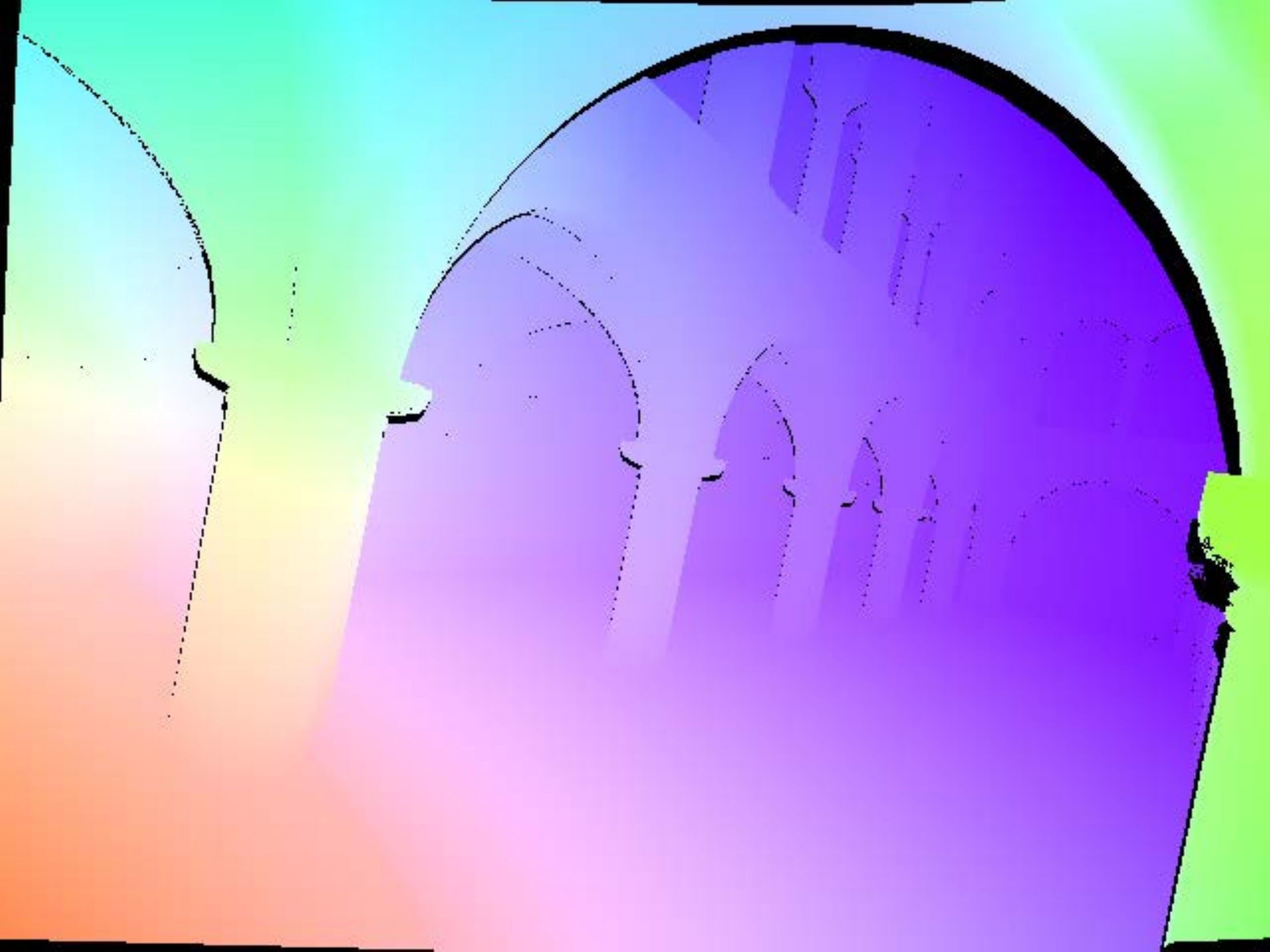}
\includegraphics[width=0.15\columnwidth]{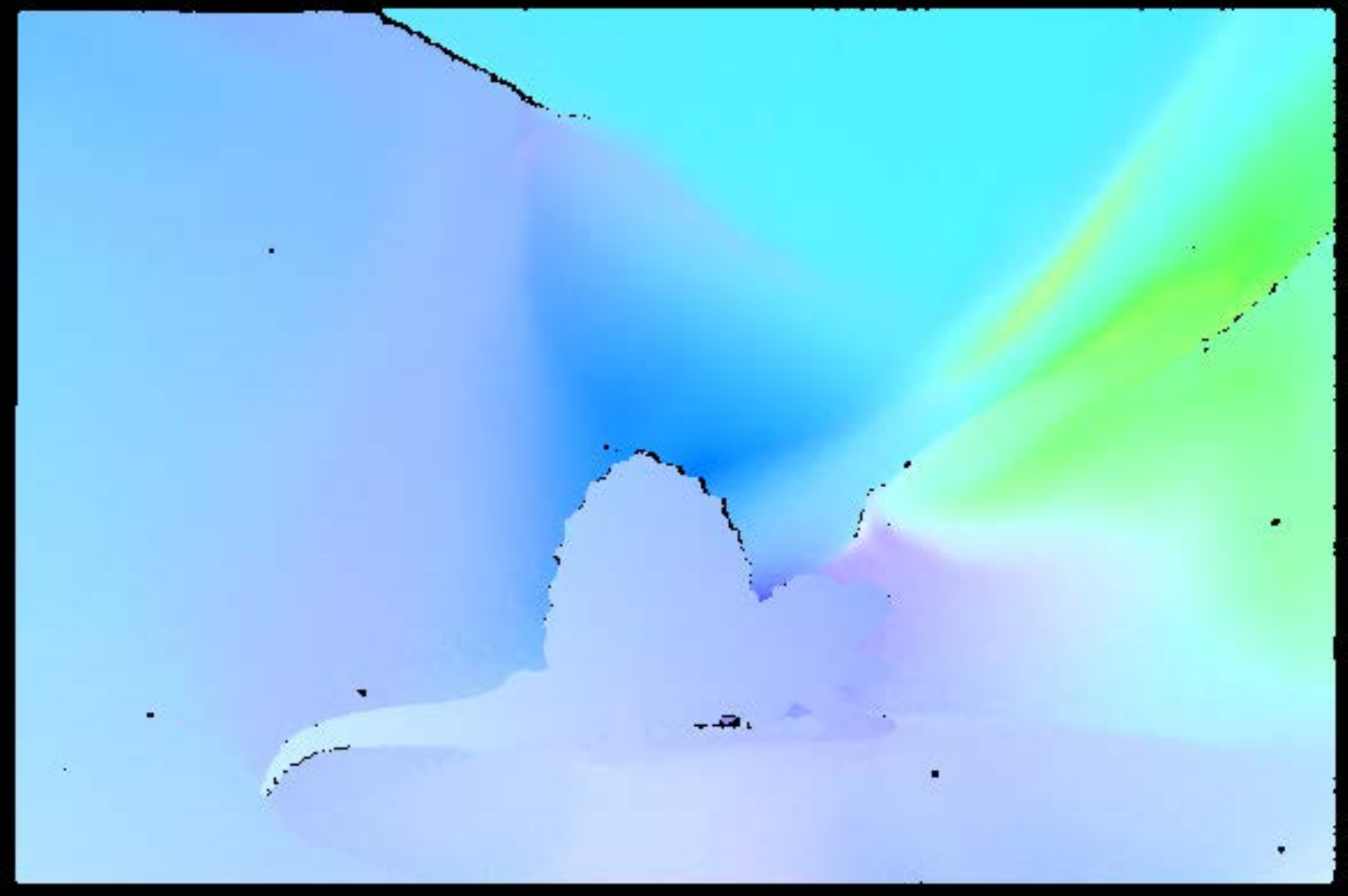}
\includegraphics[width=0.15\columnwidth]{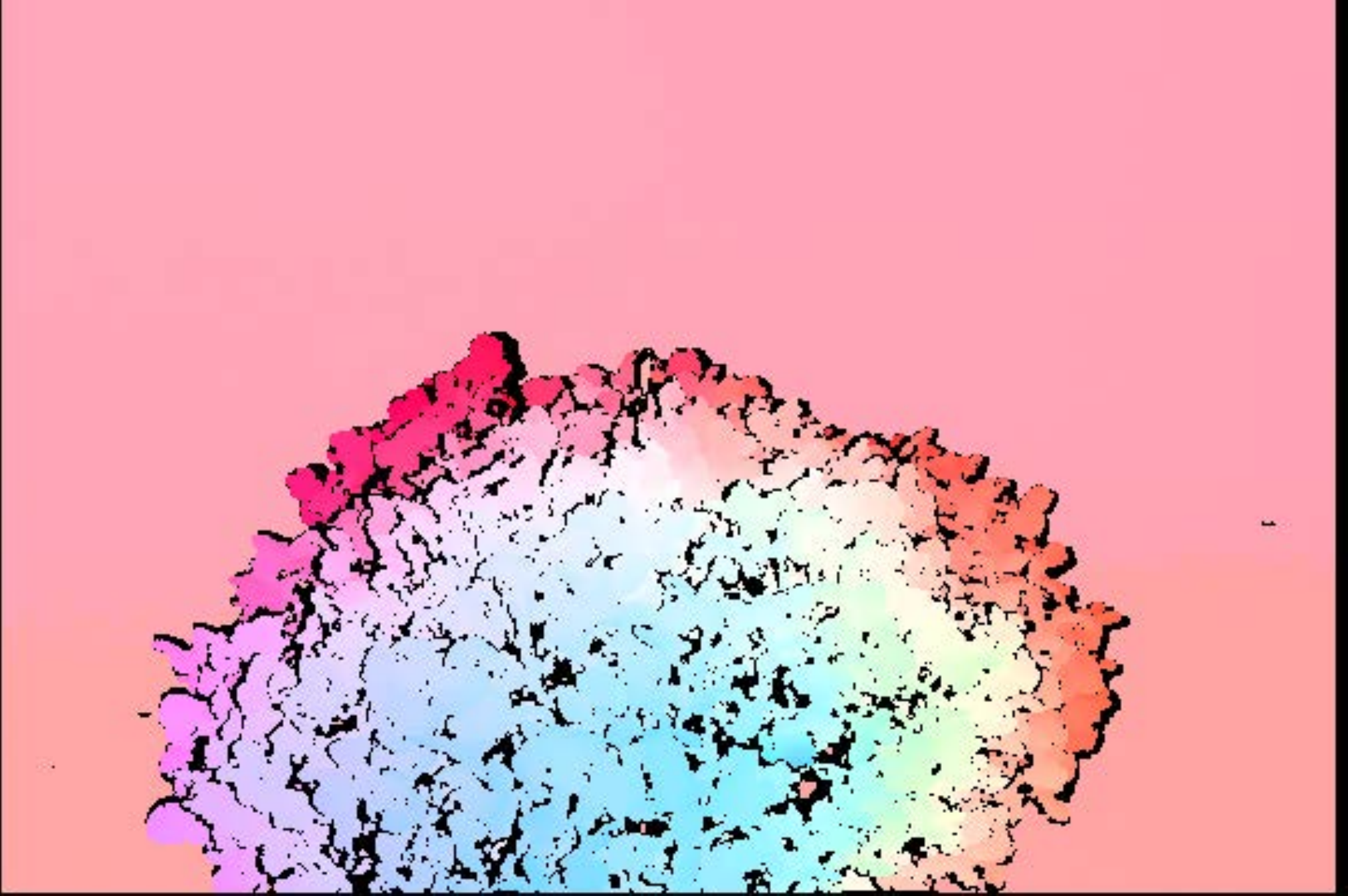}
\includegraphics[width=0.15\columnwidth]{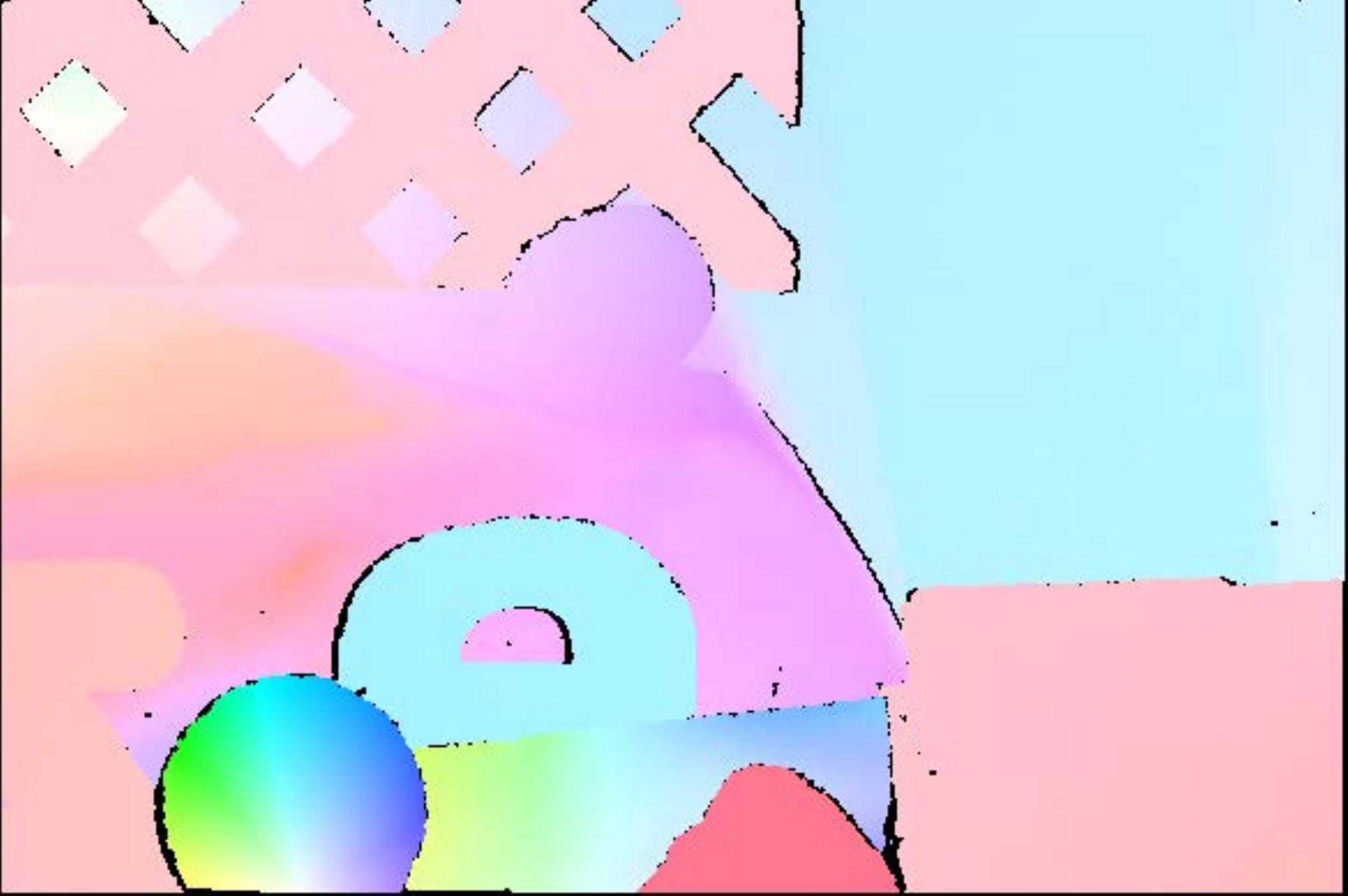}\\ \vspace{1 mm}
\caption{Examples of obtained optical flows using the MPI-SINTEL dataset with ground truth from Butler \cite{Butler2012}, for several sequences (six columns) and methods (from top to bottom \cite{Thomas1987,Argyriou2009,Gautama2002,Yan2008,Ho2008,Reyes2013,Reyes2014}, the proposed BLPC and the GT).}\label{fig:BPCReal01b}
\end{figure}

At the final stage, we unwrap the next smallest image $I^{m}_{i}(x,y)=I^{s}_{i}(x-dx(x,y),y-dy(x,y)$ (or $I^{o}_{i}(x,y)=I^{m}_{i}(x-dx(x,y),y-dy(x,y)$), where $dx$ and $dy$ are the scaled motion estimates from the previous layer. Motion compensation is applied to estimate the $\hat{I}^{m}_{i+1}$ (or $\hat{I}^{o}_{i+1}$) image and the difference $I^{m}_{d}=|I^{m}_{i}-\hat{I}^{m}_{i+1}|$ (or $I^{o}_{d}=|I^{o}_{i}-\hat{I}^{o}_{i+1}|$) is calculated. Finally, the process repeats the same, moving back to the first stage obtaining the new key point locations, until we reach the last layer that corresponds to the original size image.

In the case of an application related to face analysis or tracking the proposed optical flow framework can by adapted to estimate the motion information only in the area that a face is located. In more details, if a face is detected in a frame, then from the point locations in $\textbf{p}^{'}_{d}$, we retain only the ones that overlap with the face window. This approach can reduce significantly the computational complexity in certain applications that only for selected parts of a frame motion information is required.

%

\section{Results}
A comparative study was performed with state-of-the-art optical flow techniques operating both in the frequency and pixel domain. Video sequences with and without the ground truth were used for evaluating the performance. Also experiments with artificial data were performed to further demonstrate the concept of the proposed BLPC technique. In this study six datasets were utilised, and in more details the video sequences with ground truth provided by Baker in \cite{Baker2007}, the MPI-SINTEL dataset from Butler in \cite{Butler2012} and the UCLgtOFv1.1 in \cite{Aodha2010}. Furthermore, we used the CT dataset from Liang in \cite{Liang2014,Liang2015}, samples of videos with faces from Dhall's dataset in \cite{Dhall2012} and also 4K Test Sequences Reviving the Classics dataset \cite{Elemental2014} without ground truth. Selected frames and videos from these datasets are shown in figure \ref{fig:BPCVideoImages}. In our evaluation, several performance measures were utilised based on the availability of ground truth or not, \cite{Baker2007}. For all video sequences the motion compensated prediction error is used and is defined as the mean square error (MSE) between the ground-truth image (i.e. frame we try to predict) and the estimated compensated one
\begin{align}
MSE = \frac{1}{N}\sum_{(x,y)}(I_{c}(x,y)-I_{gt}(x,y))^{2}
\label{equ:BPC10}
\end{align}
where $N$ is the number of pixels, $I_{c}$ is the motion compensated image and $I_{gt}$ is the ground-truth frame. For color images, we take the $L2$ norm of the vector of RGB color differences. Also based on MSE we can obtain the PSNR defined as
\begin{align}
PSNR = 10*\log_{10}(\max(I_{c})^2/MSE)
\label{equ:BPC11}
\end{align}
Another measure based on the motion compensated prediction error is the gradient normalised root-mean-square difference between the ground-truth image and the compensated one
\begin{align}
NRMS = \left[\frac{1}{N}\sum_{(x,y)}\frac{(I_{c}(x,y)-I_{gt}(x,y))^{2}}{\|\nabla I_{gt}(x,y)\|^{2}+\epsilon}\right]^{\frac{1}{2}}
\label{equ:BPC12}
\end{align}
In our experiments the arbitrary scaling constant is set to be $\epsilon =1.0$ since it is the most common value used. About the motion compensation algorithm we used the one suggested in \cite{Chan2010} with subpixel accuracy.

Regarding the sequences with ground-truth flow available, the angular error (AE) between a flow vector $(u,v)$ and the ground-truth $(u_{gt},v_{gt})$ was used as a measure of performance for optical flow. The AE is defined as angle in 3D space between $(u,v,1.0)$ and $(u_{gt},v_{gt},1.0)$ and it can be computed using the equation below
\begin{align}
AE = \cos^{-1}\left(\frac{1.0+u\cdot u_{gt} + v\cdot v_{gt}}{\sqrt{1.0+u^{2}+v^{2}}\sqrt{1.0+u_{gt}^{2}+v_{gt}^{2}}}\right)
\label{equ:BPC13}
\end{align}
With this measure, errors in large flows are penalized less than errors in small flows. Also, we compute an absolute error in flow (AEF) deﬁned by:
\begin{align}
AEF = \sqrt{(u-u_{gt})^{2}+(v-v_{gt})^{2}}
\label{equ:BPC14}
\end{align}
Regarding the AEF measure is probably more appropriate for most applications, since regions of non-zero motion are not penalized less than regions of zero motion.

\textbf{Proof of concept using artificial data:} In the first part of this analysis we used artificial examples to demonstrate the issue of multiple motions present in an area for Phase Correlation. Therefore, nine pairs of images with two or more moving objects were used in our experiments (see a subset in figure \ref{fig:BPCArtificialImages}). In these examples the ground-truth is available for each moving object, and in our experiments we compared the proposed BLPC method with Phase Correlation. Corresponding blocks of size $32 \times 32$ centered around each pixel are used to apply PC and BLPC generating a dense vector field. About the boundary pixels wrap padding was used and the obtained motion vector was applied on the corresponding pixel location. Examples of the obtained optical flow estimates for each case are shown in figure \ref{fig:BPCArtificialImages}. Also we plotted the ratio of the highest over the second highest peak for each correlation surface over each pixel in all the artificial examples in figure \ref{fig:BPCArtificial02}. These figures demonstrate that the proposed approach operates with high accuracy at the presence of multiple motions in comparison with the traditional techniques. Furthermore, in table \ref{table:BPCArtificial01} all the obtained results are summarised and we can see that BLPC provides more accurate estimates both in terms of AE and PSNR.

\textbf{Comparative study using real video sequences:} Regarding the real sequences about 80 videos were used in our evaluation. We had a big variety in terms of frame resolution, starting from $584 \times 388$, nHD, HD, FHD and moving up to 4K UHD $3840\times 2160$. In this comparative study, the performance of the proposed $BLPC$ approach is compared 16 methods in total, with seven of them to be well-known state-of-the-art $PC$ based methods \cite{Thomas1987,Argyriou2009,Gautama2002,Yan2008,Ho2008,Reyes2013,Reyes2014}. In our experiments the overall obtained accuracy of the proposed $BLPC$ approach is superior in comparison to other frequency domain optical flow methods, resulting sharp, and precise motion estimates with good accuracy over motion boundaries and small details present in the image scene. Also the obtained results are close to state of the art methods operating in the pixel domain. In figure \ref{fig:BPCReal01b}, we can see the obtained optical flows for several sequences and methods including also the ground truth. In tables \ref{table:BPCReal01}-\ref{table:BPCReal05} all the obtained results are summarised and we can see that BLPC provides more accurate estimates both in terms of AE and PSNR in comparison with the other state-of-the-art frequency domain methods, while the complexity remains low. Overall the suggested optical flow method provides significant accuracy, specially at motion boundaries, and outperforms current well-known methods operating in the frequency domain.

\section{Conclusion}
In this paper, a new framework for optical flow estimation in the frequency domain was introduced. This approach is based on Phase Correlation and a novel Bilateral-Phase Correlation technique is introduced, incorporating the concept and principles of Bilateral Filters.  One of the most attractive features of the proposed scheme is that it retains the motion boundaries by taking into account the difference in value of the neighbouring pixels to preserve motion edges in combination with a weighted average of nearby intensity values, in a manner very similar to Gaussian convolution. BLPC yields very accurate motion estimates for a variety of test material and motion scenarios, and outperforms optical flow techniques, which are the current registration methods of choice in the frequency domain.

\section*{Acknowledgements}
\noindent This work is co-funded by the NATO within the WITNESS project under grant agreement number G5437. The Titan X Pascal used for this research was donated by NVIDIA.

{\small
\bibliographystyle{ieee}
\bibliography{BLPC_Ref}
}

\end{document}